\documentclass[square,numbers]{article}

\usepackage[preprint]{neurips_2024}

\usepackage[utf8]{inputenc} 
\usepackage[T1]{fontenc}    
\usepackage{hyperref}       
\usepackage{url}            
\usepackage{booktabs}       
\usepackage{amsfonts}       
\usepackage{nicefrac}       
\usepackage{microtype}      
\usepackage{xcolor}         
\usepackage{times}
\usepackage{soul}
\usepackage{graphicx}
\usepackage{amsmath}
\usepackage{amsthm}
\usepackage{amssymb}
\usepackage{algorithm}
\usepackage{algorithmic}
\usepackage{array}
\usepackage{longtable}
\usepackage{multirow}
\usepackage{wrapfig}
\usepackage{subcaption}     
\usepackage{colortbl}

\newtheorem{theorem}{Theorem}
\newtheorem{definition}{Definition}
\newtheorem{hypothesis}{Hypothesis}

\newtheorem{proposition}{Proposition}

\title{Unifying and Beyond the Theory of \\Pervasive Capability Degradation in LLM Post-training SFT from Feature Interference}

\author{%
  Hang Chen \\
  College of Computing and Data Science\\
 Nanyang Technological University\\
  \texttt{chen.hang@ntu.edu.sg} \\
  \And
   Jiaying Zhu\\
  School of Computer Science and Engineering\\
 The Chinese University of Hong Kong\\
  \texttt{jyzhu24@cse.cuhk.edu.hk} \\
  \And
  Wenya Wang\thanks{Corresponding author}\\
  College of Computing and Data Science\\
 Nanyang Technological University\\
  \texttt{wangwy@ntu.edu.sg} \\
}

\begin{document}

\maketitle

\begin{abstract}
Pervasive Capability Degradation (PCD) refers to the phenomenon in which supervised fine-tuning (SFT), while adapting a large language model (LLM) to a target task, often erodes the general capabilities acquired during pre-training. Prevailing explanations attribute PCD to macroscopic distribution shift and therefore lack microscopic manipulability, leaving existing mitigation methods without a complete theoretical foundation. In this paper, we first develop a bottom-up geometric theory of PCD grounded in feature interference under superposition: narrow-domain SFT data disrupt the almost-orthogonal feature structure formed during pre-training, and the resulting interference drives activations across decision hyperplanes in the output Laguerre–Voronoi geometry, flipping predictions on general tasks. This theory unifies existing methods into six paradigms and reveals both the source of their efficacy and their limitations. We then argue that cross-entropy under one-hot supervision is structurally limited in resolving PCD on its own, and accordingly propose the AEGIS loss, which directly constrains the drift of activation margins along all competing token directions, thereby decoupling interference suppression from task adaptation. Across 3 model families and 14 SFT tasks, AEGIS outperforms representative methods of every paradigm on target tasks and pervasive capabilities simultaneously, at only about 1.3× the computational cost of standard SFT. Its generality, scalability, and mechanistic interpretability offer a new perspective on mitigating PCD. Code is available at \url{https://github.com/Zodiark-ch/SFT-Conflict}.

\end{abstract}
\section{Introduction}
\label{sec:introduction}

Large-scale pre-training endows LLMs with broad general skills, which we refer to as \textit{pervasive task capabilities}~\citep{10.1145/3821566,rahman2025llmbaseddatascienceagents}. Following pre-training, supervised fine-tuning (SFT) serves as the primary approach for adapting general base models into specialized agents~\citep{11667189,harada-etal-2025-massive}. Simply put, post-training SFT aims to learn a downstream \textit{target task} while preserving the model's pre-trained pervasive capabilities~\citep{lai-etal-2025-survey,ICLR2026_5d6ae8ba}. However, SFT on a target task often degrades these pervasive capabilities—a phenomenon variously known as catastrophic forgetting~\citep{doi:10.1073/pnas.1611835114,pmlr-v80-serra18a}, alignment tax~\citep{fu-etal-2024-disperse,young2026alignmenttax}, or representation collapse~\citep{razdaibiedina-etal-2023-representation,pmlr-v202-yang23m}, which we unify under the term \textbf{Pervasive Capability Degradation (PCD)}. Rooted in the fundamental conflict between acquiring new skills and retaining existing ones, PCD remains a long-standing challenge in deep learning and a major bottleneck across transfer learning~\citep{5288526}, continual learning~\citep{pmlr-v70-zenke17a}, multi-task learning~\citep{11701460}, and model merging.

Currently, the prevailing explanation for PCD in the deep learning community is \textit{distribution conflict}~\citep{NEURIPS2025_6599417a,dong-etal-2024-abilities}: as a model adapts to a target task, its conditional output distribution $p(y \mid x)$ shifts toward the target distribution, thereby disrupting the original pervasive task distribution. Based on this view, numerous mitigation strategies have been proposed. For instance, data-centric approaches collect large sets of pervasive task samples and train on them alongside target task data, attempting to maintain the pervasive distribution while fitting the target distribution~\citep{ye2025limo,ICLR2025_be06e380}. Alternatively, algebraic approaches constrain parameter updates or loss gradients of the new task to be orthogonal to those of existing tasks, reducing interference with the pervasive distribution~\citep{8877742,9578000}. In our prior exploratory work~\citep{NEURIPS2025_dd37fdb2,ICLR2026_5576b86e,NEURIPS2026_chen,chen2026staticinterpretabilityanticipatingpostsft}, we also approached this problem from a mechanistic interpretability perspective, proposing the \textbf{Mechanistic Interpretability and Localization (MIL)} paradigm. Specifically, by using techniques such as circuit discovery to identify whether neurons in parameter space are relevant to the target task, we advocated updating only task-relevant neurons while freezing irrelevant ones to reduce the impact on the pervasive distribution.

However, the distribution conflict theory describes the overall shift of $p(y \mid x)$ only at a macroscopic statistical level. Although metrics such as KL divergence~\citep{karan2026finetuningsamplingsftlearns}, Wasserstein distance~\citep{kolluri-etal-2025-finetuning}, or the Fisher information matrix~\citep{muhamed2025towards} can model these shifts mathematically, they lack \textit{microscopic manipulability}. For example, macroscopic distributions cannot tell us how conflicts arise within internal activations, how to locally adapt distributions without collateral damage, or why standard cross-entropy gradients fail to reach an ideal distribution $p(y \mid x)$ that satisfies both tasks. Due to this lack of microscopic manipulability, existing methods rely on heuristic external constraints—such as above data mixing or gradient orthogonalization—to indirectly ease the conflict. While intuitive and supported by macroscopic theory, these constraints lack a complete microscopic control chain and thus only partially alleviate PCD. Taking our prior MIL work as an example, because the residual stream has a finite dimension, the model compresses a vast number of semantic features in \textit{superposition}~\citep{elhage2022toymodelssuperposition}, causing severe polysemantic entanglement between physical parameters and semantic features. Consequently, simply partitioning, pruning, or orthogonalizing weights in a static weight space cannot fully decouple target task features from irrelevant pervasive features at the parameter level~\citep{NEURIPS2023_3927bbdc}. This limitation shows that resolving PCD requires moving beyond the static weight space to directly manipulate dynamic hidden activation manifolds and output geometry.

Motivated by this insight, we develop a bottom-up, formal geometric theory to explain PCD at the microscopic level, unify existing PCD mitigation paradigms to identify their theoretical limits, and propose a loss function in activation geometry to directly address PCD.

Specifically, under the superposition~\citep{elhage2022toymodelssuperposition} and linear representation hypotheses~\citep{6300608}, hidden activations can be linearly decomposed into a combination of sparse semantic features. During pre-training, high data diversity ensures extreme feature sparsity, so any given group of features rarely co-activates, the 1-sparse loss ($L_1$) carries most of the weight of their interaction, and features are arranged into an almost-orthogonal configuration. 
During SFT, however, narrow domain data shifting loss weight from $L_1$ toward higher-order terms ($L_N$) and triggers a \textit{geometric phase transition}, which merges correlated features to significantly reduce feature orthogonality. This loss of orthogonality induces strong \textit{feature interference} when evaluating pervasive tasks. In the output Laguerre-Voronoi Diagram (LVD) space~\citep{ma2026laguerre}, this interference drives activations to drift toward competing token polyhedra coupled with these interfering features; crossing a decision hyperplane causes an incorrect competing token to win, resulting in PCD. Building on this theory, we categorize existing PCD mitigation methods into six paradigms and explain both their effectiveness and their theoretical limits: 
(1) \textbf{Reference Distribution Anchoring (RDA)}~\citep{schulman2017proximal,rafailov2023direct} constrains the KL divergence from the base model's output distribution to maintain the pervasive distribution, but suffers from a trade-off between target adaptation and reference preservation; 
(2) \textbf{Data Mixture Alignment (DMA)}~\citep{zhou2023lima,renduchintala2024smart} collects pervasive samples to approximate the pervasive distribution and increase feature sparsity against phase collapse, but faces an orders-of-magnitude sample deficit that leaves long-tail features unsupervised; 
(3) \textbf{Parameter-Space Orthogonalization (PSO)}~\citep{stoica2025model, gargiulo2025task} forces new task parameters to be orthogonal to existing ones to reduce interference cost, but suffers from severe attenuation as parameter orthogonality translates into feature orthogonality across non-linear activations; 
(4) \textbf{Delta-Parameter Pruning (DPP)}~\citep{pmlr-v235-yu24p,deng2025dare} and (5) \textbf{Mechanistic Interpretability and Localization (MIL)}~\citep{ICLR2026_5576b86e,chen2026staticinterpretabilityanticipatingpostsft} prune most SFT parameter updates or freeze task-irrelevant neurons to restrict anisotropic activation drift, yet neither can fully decouple the noise effect of feature interference; and 
(6) \textbf{Gradient Orthogonalization Optimization (GOO)}~\citep{lopez2017gradient,chaudhry2018efficient} constrains target task gradients to be orthogonal to other task gradients to limit interference, but is bottlenecked by the finite degrees of freedom of the gradient orthogonal basis.

Furthermore, we apply our theory to analyze the standard cross-entropy loss in SFT, showing that under one-hot supervision, the effect of feature interference is tightly coupled with task adaptation. Consequently, cross-entropy mainly counters feature merging by mimicking pre-training with massive pervasive samples. To overcome this limitation, we propose the \textbf{AEGIS} (\textbf{A}daptive \textbf{E}quipartition of \textbf{G}eometric \textbf{I}nterference in \textbf{S}uperposition) loss. By supervising activation drift across all competing token directions instead of relying solely on a discrete categorical probability distribution, AEGIS both decouples the feature interference effect from task adaptation and greatly reduces the number of samples needed for full feature coverage. This raises the cost of destructive geometric phase transitions and preserves the geometric structure of pervasive features. Empirically, we conduct comprehensive evaluations across the Mistral, Qwen, and Gemma model families on mainstream SFT tasks, including natural language understanding, arithmetic reasoning, code generation, and logical reasoning. Compared to existing paradigms, AEGIS consistently improves target task performance while preserving or even enhancing pervasive task capabilities, demonstrating the potential of our theoretical framework to resolve PCD.

In summary, our main contributions are twofold:
\begin{itemize}
    \item We bridge activation space, parameter space, and gradient feedback with macroscopic performance conflicts, offering a bottom-up account of how SFT induces PCD. Using this theory, we also unify existing PCD mitigation methodologies into six paradigms, explain why they are effective, and analyze their respective limitations.
    \item We argue through our theory that while standard cross-entropy is effective for task adaptation, it is structurally limited in resolving PCD under one-hot supervision. Building on this insight, we propose AEGIS, a loss function that leverages token LVD geometry to simultaneously constrain multiple token directions, and demonstrate its strong empirical performance across diverse post-training scenarios.
\end{itemize}

\section{Preliminaries}
To keep the theoretical framework concise and accessible across different domains, we summarize the primary mathematical symbols in Table~\ref{tab:notations}.

\begin{table*}
\centering
\caption{Summary of primary notations and symbols used throughout this paper.}
\label{tab:notations}
\resizebox{\textwidth}{!}{%
\begin{tabular}{lll}
\toprule
\textbf{Notation} & \textbf{Domain / Dimension} & \textbf{Description} \\
\midrule
\multicolumn{3}{l}{\textit{\textbf{1. Model Architecture \& Optimization}}} \\
$\theta$ & Parameter space & Model parameter set \\
$\mathcal{D}$ & Dataset & Dataset of training samples \\
$N$ & $\mathbb{N}^+$ & Total number of samples in a dataset or feature set \\
$\mathcal{V}, \, V$ & Set, $\mathbb{N}^+$ & Output vocabulary set and vocabulary size ($V = |\mathcal{V}|$) \\
$W, \, b$ & Tensor / Vector & General model weight matrices and bias vectors \\
$h$ & $\mathbb{R}^m$ & Residual stream activation vector \\
$m$ & $\mathbb{N}^+$ & Hidden dimension of the activation space ($m \ll k$) \\
$\mathcal{L}$ & $\mathbb{R}$ & Loss function \\
\midrule
\multicolumn{3}{l}{\textit{\textbf{2. Feature Superposition Theory}}} \\
$k$ & $\mathbb{N}^+$ & Total number of latent features in superposition ($k \gg m$) \\
$f, \, F$ & $\mathbb{R}^m, \, \mathbb{R}^{m \times k}$ & Unit feature direction vector and feature dictionary matrix $F = [f_1, \dots, f_k]$ \\
$a$ & $\mathbb{R}^k$ & Feature activation intensity vector $a = [a_1, \dots, a_k]^\top$ \\
\midrule
\multicolumn{3}{l}{\textit{\textbf{3. Planar Laguerre-Voronoi Diagram (LVD)}}} \\
$c_j, \, \nu_j$ & $\mathbb{R}^m, \, \mathbb{R}$ & Geometric center ($c_j \in \mathbb{R}^m$) and Laguerre scalar weight ($\nu_j \in \mathbb{R}$) of token $j$ \\
\bottomrule
\end{tabular}%
}
\end{table*}

\subsection{LLM Post-Training Supervised Fine-Tuning}
\label{subsec:prelim_sft}

Let $\mathcal{M}_{\theta^0}$ denote a foundation LLM parameterized by $\theta^0$ pre-trained on a general corpus $\mathcal{D}_G$, endowing $\mathcal{M}_{\theta^0}$ with pervasive capabilities. 
To adapt the model to a downstream target task $T$ with dataset $\mathcal{D}_{T} = \{(x_i, y_{t, i})\}_{i=1}^{N_T}$, supervised fine-tuning (SFT) updates parameters $\theta^0 \to \theta$ to master $T$ while preserving pervasive proficiency on $\mathcal{D}_G= \{(x_i, y_{ i})\}_{i=1}^{N_G}$. 
This adaptation ideally optimizes the joint objective:
\begin{equation}
    \min_{\theta} \mathbb{E}_{(x, y_t)\sim \mathcal{D}_{T}}[\mathcal{L}(y_t|x;\theta)] + \mathbb{E}_{(x, y)\sim \mathcal{D}_{G}}[\mathcal{L}(y|x;\theta)],
    \label{eq:sft_objective}
\end{equation}
where $\mathcal{L}(y|x;\theta) = -\sum_{\tau=1}^{|y|} \log P_\theta(y_\tau \mid x, y_{<\tau})$ is the standard autoregressive cross-entropy loss.

In practice, optimizing Eq.~\eqref{eq:sft_objective} directly is hindered by restricted access to $\mathcal{D}_G$ and optimization conflicts, inevitably triggering capability degradation:

\begin{definition}[Pervasive Capability Degradation, PCD]\label{defPCD}
Let $\mathcal{P}(\mathcal{M}_\theta, \mathcal{D})$ denote the evaluation performance of model $\mathcal{M}_\theta$ on distribution $\mathcal{D}$. SFT from $\theta^0$ to $\theta$ suffers from \textbf{Pervasive Capability Degradation (PCD)} if target adaptation incurs an intolerable performance drop on $\mathcal{D}_G$:
\begin{equation}
    \mathcal{P}(\mathcal{M}_{\theta}, \mathcal{D}_T) > \mathcal{P}(\mathcal{M}_{\theta^0}, \mathcal{D}_T) \quad \text{and} \quad \mathcal{P}(\mathcal{M}_{\theta^0}, \mathcal{D}_G) - \mathcal{P}(\mathcal{M}_{\theta}, \mathcal{D}_G) > \epsilon_{\text{PCD}},
\end{equation}
where $\epsilon_{\text{PCD}} > 0$ denotes the degradation tolerance threshold.
\end{definition}

\subsection{The Superposition Hypothesis and Feature Interference}
\label{subsec:prelim_superposition}

Because the number of latent semantic features $k$ learned by LLMs vastly exceeds the hidden dimension $m$ ($k \gg m$), models rely on the \textit{superposition hypothesis}~\citep{elhage2022toymodelssuperposition} to represent overcomplete feature sets within restricted residual streams.

\begin{definition}[Feature Superposition]\label{defSuperposition}
Under the linear representation hypothesis, an activation $h \in \mathbb{R}^m$ in the residual stream is expressed via a feature dictionary $F = [f_1, \dots, f_k] \in \mathbb{R}^{m \times k}$ with unit directions ($\|f_i\|_2 = 1$) and an activation intensity vector $a = [a_1, \dots, a_k]^\top \in \mathbb{R}^k$:
\begin{equation}
    h = F a = \sum_{i=1}^k a_i f_i, \quad \text{where } k \gg m.
    \label{eq:superposition}
\end{equation}
Superposition is enabled by \textbf{feature sparsity}: each feature activates with low marginal probability $P(a_i \neq 0) = 1 - S_i \ll 1$, where $S_i \to 1$.
\end{definition}

Since $k \gg m$, pairwise orthogonality across all features is geometrically impossible, guaranteeing non-zero overlap: $\exists \, i \neq j, \, f_i^\top f_j \neq 0$. Probing feature $i$ along its designated direction $f_i$ yields:
\begin{equation}
    f_i^\top h = f_i^\top \left( a_i f_i + \sum_{j \neq i} a_j f_j \right) = a_i + \sum_{j \neq i} a_j (f_i^\top f_j),
    \label{eq:interference}
\end{equation}
where $a_i$ is the target signal, and $\sum_{j \neq i} a_j (f_i^\top f_j)$ is the \textbf{interference} from non-orthogonal features. High sparsity ($S \to 1$) sustains superposition by suppressing co-activation, rendering concurrent interference quadratically rare: $P(a_i > 0, a_j > 0) = \mathcal{O}((1-S)^2) \ll 1 - S$.

In actual LLMs, physical parameter channels couple with latent features in a non-uniform and sparse manner, substantiated by two core observations. First, distinct channels (i.e., individual rows or columns of a weight matrix $W$) produce divergent output activations for the identical input state, demonstrating that individual channels allocate disparate projection weights across features. Second, each channel's sensitivity is heavily concentrated on a small subset of salient features while remaining negligible on the rest of features, a localized selectivity broadly corroborated by mechanistic interpretability literature~\citep{geva2021transformer,liu2025rethinking,gromov2025unreasonable}.

To enable mathematically tractable derivations that directly map parameter-space updates (such as norm variations, spatial rotations, and subspace orthogonalizations) to directional feature displacements in the residual stream,  we formalize this localized, sparse physical coupling into an idealized one-to-one virtual parameterization.

\begin{hypothesis}[Virtual Sparse Weight Hypothesis]\label{hypo:virtual_sparse_weight}
Under a first-order linear approximation, the composite transformation of an intermediate Transformer layer (such as an Attention, MLP block or both) is governed by an end-to-end operator $M = W_{\text{in}} W_{\text{out}} \in \mathbb{R}^{m \times m}$, where $W_{\text{in}} \in \mathbb{R}^{m \times d}$ and $W_{\text{out}} \in \mathbb{R}^{d \times m}$ with $d$ being the dimension of the Attention or MLP module. We posit the existence of an overcomplete virtual parameter space wherein $M$ factorizes losslessly into idealized feature-decoupled operators:
\begin{equation}
    M = W_{\text{in}}^{\text{virt}} W_{\text{out}}^{\text{virt}}, \quad \text{where } W_{\text{out}}^{\text{virt}} \in \mathbb{R}^{k \times m} \quad \text{and} \quad W_{\text{in}}^{\text{virt}} \in \mathbb{R}^{m \times k}.
    \label{eq:virtual_factorization}
\end{equation}\footnote{For notational conciseness in subsequent developments, our parameter-level analysis focuses on the read-in detection operator $W_{\text{out}}^{\text{virt}} \in \mathbb{R}^{k \times m}$, hereinafter denoted simply as $W \in \mathbb{R}^{k \times m}$. Unless explicitly stated otherwise, parameter matrices $W$ referenced in subsequent sections represent this canonical virtual weight.}
Each virtual parameter row $W_i \in \mathbb{R}^{1 \times m}$ corresponds bijectively to feature direction $f_i \in \mathbb{R}^m$. 
\end{hypothesis}
\begin{figure*}[htbp]
    \centering
    \includegraphics[width=0.9\textwidth]{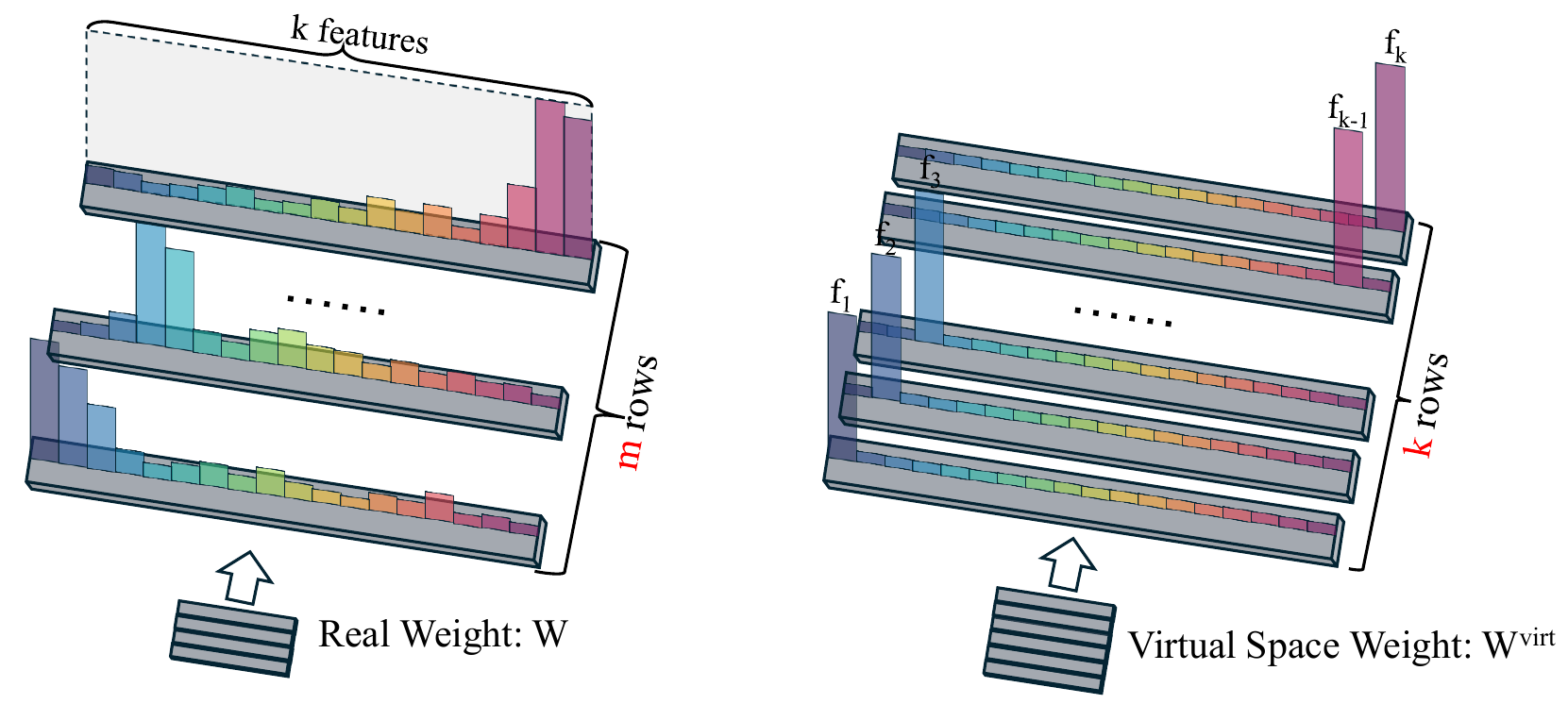}
    \caption {Illustration of real weight ($W$) and virtual space weight ($W^{virt}$). Each row of $W$ manages a subset of a few salient features, but each row of $W^{virt}$ manages one individual feature. }
    \label{fig:Kmapping}
\end{figure*}
Through Hypothesis~\ref{hypo:virtual_sparse_weight}, operations performed on individual rows of the parameter matrix translate equally to features in activation space, as visualized in Figure~\ref{fig:Kmapping}. This bridge yields two foundational algebraic properties (detailed proofs provided in Appendix~\ref{app:weight_feature_bridge_proof}):
\textbf{Orthogonal Equivalence}: $ W_a \cdot W_b =0 \Rightarrow f_a^\top f_b=0$ 
for any two feature channels $a, b \in \{1, \dots, k\}$, and \textbf{Directional Additivity}: $ f_a' \propto f_a + \lambda \left( \frac{\|W_b\|_2}{\|W_a\|_2} \right) f_b$ when $W_a' = W_a + \lambda W_b$ with $\lambda \in \mathbb{R}$.

\subsection{Laguerre Geometry of LLMs}
\label{subsec:prelim_lvd}

In autoregressive LLMs, the unembedding layer parameterized by $W_U = [W_{U, 1}, \dots, W_{U, V}] \in \mathbb{R}^{m \times V}$ and $b_U = [b_{U, 1}, \dots, b_{U, V}]^\top \in \mathbb{R}^V$ maps activation $h \in \mathbb{R}^m$ to logits $z_j(h) = W_{U, j}^\top h + b_{U, j}$, predicting token $\hat{y} = \arg\max_{j \in \mathcal{V}} z_j(h)$. The decision rule $z_j(h) \ge z_k(h)$ defines an intersection of affine half-spaces:
\begin{equation}
    (W_{U, j} - W_{U, k})^\top h + (b_{U, j} - b_{U, k}) \ge 0, \quad \forall k \in \mathcal{V} \setminus \{j\},
    \label{eq:hyperplane_boundary}
\end{equation}
tessellating $\mathbb{R}^m$ into convex polyhedral cells equivalent to a Laguerre-Voronoi Diagram (LVD).

\begin{definition}[Laguerre-Voronoi Diagram, LVD]\label{defLVD}
An LVD (or power diagram) partitions $\mathbb{R}^m$ using weighted sites $\{(c_j, \nu_j)\}_{j=1}^V$ with centers $c_j \in \mathbb{R}^m$ and scalar weights $\nu_j \in \mathbb{R}$. The power distance from activation $h \in \mathbb{R}^m$ to site $j$ is defined as:
\begin{equation}
    D_j(h) = \|h - c_j\|_2^2 - \nu_j.
    \label{eq:laguerre_dist}
\end{equation}
The convex polyhedral cell for token $j$ is given by:
\begin{equation}
    \text{Cell}_j^{\text{LVD}} = \left\{ h \in \mathbb{R}^m \;\middle|\; D_j(h) \le D_k(h), \; \forall k \in \mathcal{V} \right\}.
    \label{eq:lvd_cell}
\end{equation}
\end{definition}

Locating the minimal power distance cell strictly equates to selecting the maximum logit ($\arg\min_{j \in \mathcal{V}} D_j(h) \equiv \arg\max_{j \in \mathcal{V}} z_j(h)$) via the lossless parameter mapping~\citep{ma2026laguerre}:
\begin{equation}
    c_j = \frac{1}{2} W_{U, j} \in \mathbb{R}^m, \quad \nu_j = b_{U, j} + \frac{1}{4}\|W_{U, j}\|_2^2 \in \mathbb{R},
    \label{eq:lvd_mapping}
\end{equation}
which yields $D_j(h) = \|h\|_2^2 - z_j(h)$, where $\|h\|_2^2$ is invariant across candidate tokens.

Moreover, recent geometric studies~\citep{budd2025splinterpimprovingunderstandingtraining} establish that  each token's cell $\text{Cell}_j^{\text{LVD}}$ is predominantly governed by a sparse subset of latent features $\mathcal{S}_j \subset \{f_1, f_2, \dots\}$ ($|\mathcal{S}_j| \ll k$), such that its unembedding direction can be effectively decomposed as $W_{U,j} \approx \sum_{i \in \mathcal{S}_j} \omega_{j,i} f_i$ with projection weights $\omega_{j,i} > 0$. Consequently, each token's decision region is steered by a localized, distinct cluster of features.

\section{Mechanistic Explanation of PCD}
\label{sec:mechanistic_theory}

In this section, we formulate a mechanistic theory explaining how SFT induces Pervasive Capability Degradation (PCD). We show that domain narrowing during SFT shifts the weight of the reconstruction loss from lower-order toward higher-order terms, triggering geometric phase transitions that reduce the effective number of feature directions. While this reduction suppresses compounding interference on the target task, it induces directional interference surges on general data, eroding LVD cell margins and causing discrete prediction errors.

\subsection{Shift of Loss Weight from Lower-Order to Higher-Order Terms}
\label{subsec:loss_shift}

Under Definition~\ref{defSuperposition}, the model encodes $k$ latent semantic features within an $m$-dimensional residual stream ($k \gg m$). The overall expected feature reconstruction loss $\mathcal{L}$ expands as a binomial mixture across feature co-activation cardinalities~\citep{elhage2022toymodelssuperposition}:
\begin{equation}
    \mathcal{L} = \sum_{p=0}^k \binom{k}{p} (1-S)^p S^{k-p} L_p = S^k L_0 + k(1-S)S^{k-1} L_1 + \binom{k}{2}(1-S)^2 S^{k-2} L_2 + \dots,
    \label{eq:binomial_loss_expansion}
\end{equation}
where $S \in [0, 1)$ denotes feature sparsity, and $L_p$ represents the conditional reconstruction loss when exactly $p$ latent features concurrently activate (a $p$-sparse activation state). 
In the pre-training corpus $\mathcal{D}_G$, broad semantic diversity yields high feature sparsity ($S \to 1$), making marginal activation probabilities vanishingly small ($1 - S \ll 1$). Because $(1-S) \gg (1-S)^2 \gg \dots \gg (1-S)^k$, higher-order co-activations are negligible. As $L_0$ is a feature-independent bias penalty, the interaction among such features on $\mathcal{D}_G$ is strictly governed by the lower-order loss $L_1$ (the 1-sparse conditional loss)~\citep{elhage2022toymodelssuperposition}:
\begin{equation}
    L_1 \approx \underbrace{\sum_{i=1}^k I_i \left( 1 - \mathrm{ReLU}(\|f_i\|_2^2) \right)^2}_{\text{Feature Benefit}} + \underbrace{\sum_{i \neq j} I_j \mathrm{ReLU}(f_j^\top f_i)^2}_{\text{Interference Cost}},
    \label{eq:l1_breakdown}
\end{equation}
where $I_i$ is the importance of feature $i$. When $L_1$ carries most of the weight, a feature mostly activates without its potential interferers. Balancing reconstruction accuracy against pairwise cross-talk, the model organizes the $k$ feature vectors into an almost-orthogonal polyhedral packing in $\mathbb{R}^m$ where mutual overlaps satisfy $|f_i^\top f_j| \le \epsilon \ll 1$.

Conversely, the SFT dataset $\mathcal{D}_T$ is domain-specific, substantially increasing the activation frequency of task-relevant features ($1 - S_T \gg 1 - S_G$). Furthermore, a subset of $N$ task-relevant features $\mathcal{S} = \{f_1, \dots, f_N\}$ ($N \ge 2$) becomes strongly correlated: $P_{\mathcal{D}_T}(a_1 > 0, \dots, a_N > 0) \to 1$. 
When these correlated features co-activate under the higher-order loss $L_N$, assuming unit intensities $a_n = 1$ and equal importances $I_n = I$ for $n \in \mathcal{S}$, the conditional loss $L_N$ evaluates to:
\begin{equation}
    L_{N, \mathcal{S}} = \sum_{n=1}^N I \left( 1 - f_n^\top h \right)^2 = \sum_{n=1}^N I \left( 1 - \|f_n\|_2^2 - \sum_{j \neq n} f_n^\top f_j \right)^2.
\end{equation}
Assuming normalized vectors ($\|f_n\|_2 \approx 1$), expanding this expression isolates the compounding interference cost:
\begin{equation}
    \mathcal{C}_{\text{inter}}^{(N)} = I \sum_{n=1}^N \left( \sum_{j \neq n} f_n^\top f_j \right)^2 = I \sum_{n=1}^N \left[ \sum_{j \neq n} (f_n^\top f_j)^2 + \sum_{j \neq n} \sum_{l \neq n, j} (f_n^\top f_j)(f_n^\top f_l) \right].
    \label{eq:compounding_inter_general}
\end{equation}
Equation~\eqref{eq:compounding_inter_general} shows that compounding interference scales quadratically as $\mathcal{O}(N^2)$ with the number of bound features. To minimize this cost, the representation undergoes a sharp geometric phase transition~\citep{elhage2022toymodelssuperposition}: the model abandons symmetric polytopes that maintain distinct directions, snapping into configurations that represent fewer total feature directions.

\paragraph{Example: From $\mathcal{L}_1$ to $\mathcal{L}_2$.}
To ground this transition, consider two features $f_a$ and $f_b$ that shift from independent
activations under $\mathcal{L}_1$ to deterministic co-occurrence under $\mathcal{L}_2$
($a_a = 1$, $a_b = 1$), embedded among the remaining features $f_c$ ($c \notin \{a,b\}$).
Under the normalization of Eq.~\ref{eq:compounding_inter_general}, the loss involving this pair is:
\begin{equation}
  \mathcal{L}_{2,\{a,b\}}
  = \underbrace{I(1 - \|f_a\|_2^2)^2 + I(1 - \|f_b\|_2^2)^2}_{\text{Feature Benefit}}
  \;+\; \underbrace{2I(f_a^\top f_b)^2 \;+\; \sum_{c \notin \{a,b\}} \lambda_c
        \Bigl[(f_c^\top f_a)^2 + (f_c^\top f_b)^2\Bigr]}_{\text{Interference Cost}},
  \label{eq:L2_pair}
\end{equation}
where the first interference term is the compounding cross-talk within the pair, and the second collects
the cross-talk with all other features, weighted by $\lambda_c = I_c(1 - S_c)$.
Preserving two distinct directions leaves two ways to remove the compounding cross-talk
$2I(f_a^\top f_b)^2$. The first is \emph{orthogonalization}, which keeps both features and drives
$f_a^\top f_b \to 0$. The second is \emph{collapse}, which merges the pair into a single feature along
their principal component:
\begin{equation}
  f_{\mathrm{PCA}} = \frac{f_a + f_b}{\|f_a + f_b\|_2} \approx \frac{f_a + f_b}{\sqrt{2}} .
  \label{eq:fpca}
\end{equation}
Collapse does not make the two directions orthogonal. It removes the pair itself: once $\{f_a, f_b\}$ is
replaced by the single feature $f_{\mathrm{PCA}}$, there is no second feature left to interfere with, so
the compounding term disappears. Since $P_{D_T}(a_a \neq a_b) \approx 0$, $f_{\mathrm{PCA}}$ carries the
joint signal and the feature benefit is preserved almost losslessly.
 
Both routes differ in the cross-talk 
with the remaining features. Let $u = f_{\mathrm{PCA}}$ and let $v$ be the unit vector orthogonal to $u$
in $\mathrm{span}(f_a, f_b)$. Orthogonalization within this plane keeps both directions occupied,
whereas collapse retains only $u$:
\begin{equation}
  \sum_{c \notin \{a,b\}} \lambda_c
  \Bigl[(f_c^\top u)^2 + (f_c^\top v)^2\Bigr]
  \;\ge\;
  \sum_{c \notin \{a,b\}} \lambda_c\, (f_c^\top u)^2 .
  \label{eq:ext_gap}
\end{equation}
Collapse therefore lowers the cross-talk with the other features by
$\sum_{c} \lambda_c (f_c^\top v)^2$ and releases the direction $v$ for them. 
When capacity is insufficient, the model therefore favors collapse over orthogonalization. This agrees
with the observation~\citep{elhage2022toymodelssuperposition} that correlated features form local orthogonal
bases only when capacity permits, and otherwise collapse onto their principal components.   

Although feature reduction is optimal on $\mathcal{D}_T$, it impairs pervasive capabilities on $\mathcal{D}_G$.\footnote{Without loss of generality, we assume the general pervasive task distribution and pervasiveness-capabilities evaluation distribution align with the pre-training corpus.} Having reduced its feature directions, the model can no longer isolate individual features. Specifically, consider a pervasive input that triggers an isolated feature $f_i \in \mathcal{S}$ with unit intensity, yielding the true activation $h_{\text{true}} = f_i$. Because the representation has collapsed, this input is read by and reconstructed from the shared principal component $f_{\text{PCA}}$, producing a distorted reconstruction state $h_{\text{recon}} = (f_{\text{PCA}}^\top f_i) f_{\text{PCA}} = \frac{1}{|\mathcal{S}|} \sum_{j \in \mathcal{S}} f_j$.

Algebraically, $h_{\text{recon}}$ across dormant channels distorts the effective feature Gram matrix $G = F^\top F \in \mathbb{R}^{k \times k}$. While pre-training maintains near-orthogonality ($G_{ij} \approx 0$ for $i \neq j$), the post-SFT phase transition elevates off-diagonal terms to positive values, inducing an unprovoked \textit{ghost signal spike} on dormant channels:
\begin{equation}
    \Delta a_j = f_j^\top h_{\text{recon}} \gg 0, \quad \forall j \in \mathcal{S} \setminus \{i\}.
    \label{eq:ghost_signal_spike}
\end{equation}
This spurious excitation introduces a non-zero representation error vector $\delta r_\ell \in \mathbb{R}^m$ at layer $\ell$:
\begin{equation}
    \delta r_\ell = h_{\text{recon}} - h_{\text{true}} = \sum_{j \in \mathcal{S}} \Delta a_j f_j \neq \mathbf{0}.
    \label{eq:layer_perturbation_vector}
\end{equation}

Propagating through subsequent Transformer layers, $\delta r_l$ accumulates into an activation displacement $\delta h$ at the unembedding layer, approximated via first-order multivariate Taylor expansion:
\begin{equation}
    \delta h \approx J_{l \to L}(x) \cdot \delta r_l,
    \label{eq:delta_h}
\end{equation}
\begin{wrapfigure}{r}{0.42\linewidth}
  \vspace{-12pt}
  \centering
  \includegraphics[width=\linewidth]{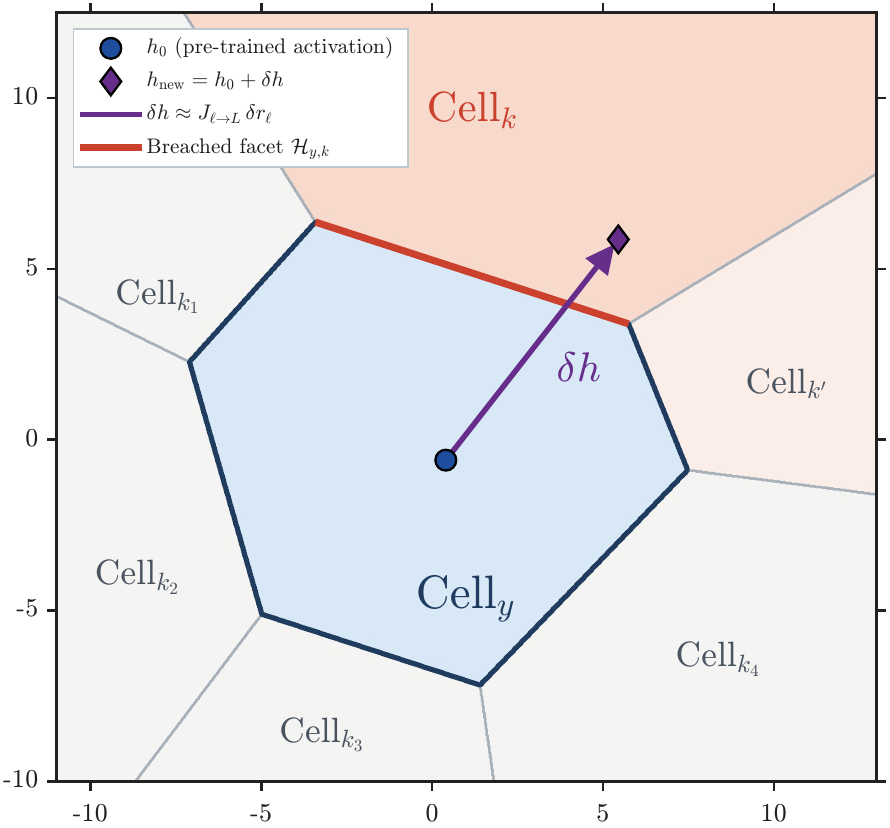}
  \caption{Geometric illustration of PCD in the output LVD (Eqs.~(18)--(20)).}
  \label{fig:lvd_drift}
  \vspace{-10pt}
\end{wrapfigure}
where $J_{l \to L}(x) = \frac{\partial h_L}{\partial r_l} \in \mathbb{R}^{m \times m}$ is the inter-layer propagation Jacobian.
Crucially, because the spurious excitation vector $\delta r_l = \sum_{j \in \mathcal{S}} \Delta a_j f_j$ is strictly confined to the task-correlated feature subspace $\text{span}(\mathcal{S})$, the resulting displacement $\delta h$ is strongly anisotropic rather than a diffuse perturbation across all dimensions. Recalling from Section~\ref{subsec:prelim_lvd} that each token's LVD cell is governed by a sparse set of dominant features, there typically exist competing tokens $k \in \mathcal{V}$ whose salient features $\mathcal{S}_k$ overlap or strongly align with $\mathcal{S}$ (i.e., $(W_{U,k} - W_{U,y})^\top \delta h \gg 0$). Consequently, $\delta h$ acts as a directed semantic attractor, pulling $h_{\text{new}} = h_0 + \delta h$ preferentially toward the bounding hyperplanes of these $\mathcal{S}$-coupled tokens.

Focusing on such a competitor token $k$, the classification margin relative to the ground-truth token $y$ updates to:
\begin{equation}
    M_{y,k}^{(\text{new})} = M_{y,k}^{(0)} + (W_{U,y} - W_{U,k})^\top \delta h = M_{y,k}^{(0)} - (W_{U,k} - W_{U,y})^\top \delta h,
    \label{eq:margin_update}
\end{equation}
where $M_{y,k}^{(0)} = (W_{U,y} - W_{U,k})^\top h_0 + (b_{U,y} - b_{U,k}) > 0$ is the baseline safe margin under the pre-trained activation $h_0$. When the directional push along the outward normal $(W_{U,k} - W_{U,y})$ overcomes the baseline margin:
\begin{equation}
    (W_{U,k} - W_{U,y})^\top \delta h \ge M_{y,k}^{(0)} \iff M_{y,k}^{(\text{new})} \le 0,
    \label{eq:breach_condition}
\end{equation}
the activation vector breaches the bounding hyperplane. Consequently, $h_{\text{new}}$ exits $\text{Cell}_y^{\text{LVD}}$ and falls into $\text{Cell}_k^{\text{LVD}}$, as shown in Figure~\ref{fig:lvd_drift}, inverting the discrete prediction ($\hat{y} = k \ne y$) and triggering PCD. 


\subsection{Unifying Existing Paradigms via Our Mechanistic Explanation}
\label{subsec:unifying_paradigms}

To contextualize PCD within broader post-training research, we systematically investigate existing methodologies across diverse subfields dedicated to mitigating capability degradation not only in the pure SFT scenarios, including continual learning~\citep{yang2025recent,zheng2025towards}, parameter-efficient fine-tuning (PEFT)~\citep{xu2026parameter,wang2025parameter}, model merging~\citep{li2025deep,yang2026model,yang2024adamerging}, instruction data curation~\citep{yin2023dynosaur,wang2024data}, model evaluation~\citep{xuan2025mmlu,li2025data}, machine unlearning~\citep{wu2023depn,li2025effective}, preference alignment~\citep{wang2024comprehensive,deng2026less}, and reinforcement learning (RL)~\citep{kakade2002approximately,ziegler2019fine}. 
We distill the foundational theoretical approaches developed across these disciplines into six core paradigms, which intervene across distinct operational dimensions: output probability distributions, training data mixtures, weight parameter geometry, parameter update pruning, loss gradients, and localized sub-networks. 
In the following subsubsections, we systematically unify these paradigms through our theoretical framework—integrating feature superposition, and LVD geometry—to elucidate the exact geometric mechanisms governing their efficacy. 

\subsubsection{Paradigm 1: Reference Distribution Anchoring (RDA)}
\label{subsubsec:rl_alignment}

RDA realized via KL divergence regularization $D_{\text{KL}}\left(\pi_\theta(y \mid x) \,\|\, \pi_{\text{ref}}(y \mid x)\right)$, mitigates PCD by penalizing the distributional deviation of the adapted model $\pi_{\theta}$ from the pre-trained reference policy $\pi_{\text{ref}}$, thereby constraining output token shifts to preserve pre-existing capabilities. This regularization mechanism is predominantly employed in RL alignment frameworks~\citep{kakade2002approximately,ziegler2019fine}. Specifically, the constrained policy optimization objective is formulated as:
\begin{equation}
    \max_{\pi_\theta} \mathbb{E}_{x \sim \mathcal{D}, \, y \sim \pi_\theta} [r(x, y)] - \beta D_{\text{KL}}\left(\pi_\theta(y \mid x) \,\|\, \pi_{\text{ref}}(y \mid x)\right),
    \label{eq:rl_kl_objective}
\end{equation}
where $r(x, y) \in \mathbb{R}$ denotes the scalar reward function, and $\beta > 0$ governs the penalty strength against reference distribution drift.

Let $z = W_U^\top h + b_U \in \mathbb{R}^V$ and $z_0 = W_U^\top h_0 + b_U$ denote the post-RDA and reference logits, yielding logit shift $\Delta z = z-z_0= W_U^\top \Delta h$. A second-order Taylor expansion of $D_{\text{KL}}(\pi_\theta \,\|\, \pi_{\text{ref}})$ around $\Delta z = \mathbf{0}$ reads:
\begin{equation}
    D_{\text{KL}}(\pi_\theta \,\|\, \pi_{\text{ref}}) \approx \frac{1}{2} \Delta z^\top F \Delta z = \frac{1}{2} \Delta h^\top \left( W_U F W_U^\top \right) \Delta h = \frac{1}{2} \Delta h^\top H_{\text{Fisher}} \Delta h,
    \label{eq:kl_fisher_expansion}
\end{equation}
where $F = \operatorname{diag}(\pi_{\text{ref}}) - \pi_{\text{ref}} \pi_{\text{ref}}^\top \in \mathbb{R}^{V \times V}$ is the Fisher information matrix of the reference categorical distribution, and $H_{\text{Fisher}} = W_U F W_U^\top \in \mathbb{R}^{m \times m}$ denotes the \textit{induced Fisher metric tensor} on the residual stream.

For the reference top-1 token $j^* = \arg\max_{j \in \mathcal{V}} z_{0, j}$, its baseline safety margin relative to competitor $k \neq j^*$ is $d_{n, k}^{(0)} = \frac{(W_{U, j^*} - W_{U, k})^\top h_0 + (b_{U, j^*} - b_{U, k})}{\|W_{U, j^*} - W_{U, k}\|_2} > 0$. Following activation shift $\Delta h$, the updated margin is $d_{n, k}^{(\text{new})} = d_{n, k}^{(0)} - \frac{(W_{U, k} - W_{U, j^*})^\top \Delta h}{\|W_{U, k} - W_{U, j^*}\|_2}$. Applying the Cauchy-Schwarz inequality under the $H_{\text{Fisher}}$ metric yields an second-order lower bound on the residual margin (proof in Appendix~\ref{app:rl_proofs}):
\begin{equation}
    d_{n, k}^{(\text{new})} \ge d_{n, k}^{(0)} - C_k \sqrt{2 D_{\text{KL}}(\pi_\theta \,\|\, \pi_{\text{ref}})}, \quad \text{where} \quad C_k = \frac{\|W_{U, k} - W_{U, j^*}\|_{H_{\text{Fisher}}^\dagger}}{\|W_{U, k} - W_{U, j^*}\|_2},
    \label{eq:margin_lower_bound_kl}
\end{equation}
and $H_{\text{Fisher}}^\dagger$ is the Moore-Penrose pseudo-inverse. Consequently, maintaining divergence below the threshold
\begin{equation}
    D_{\text{KL}}(\pi_\theta \,\|\, \pi_{\text{ref}}) < \min_{k \neq j^*} \frac{(d_{n, k}^{(0)})^2}{2 C_k^2}
    \label{eq:kl_safety_threshold}
\end{equation}
strictly guarantees $d_{n, k}^{(\text{new})} > 0$ across all competing hyperplanes, preventing $h$ from exiting $\mathrm{Cell}_{j^*}^{\text{LVD}}$ and precluding prediction flips on general tasks.

Established alignment algorithms enforce Eq.~\eqref{eq:kl_safety_threshold} through distinct mechanisms:
\textbf{PPO}~\citep{schulman2017proximal} combines token-level KL penalties with surrogate clipping $\operatorname{clip}(r_t(\theta), 1-\epsilon, 1+\epsilon)$ on probability ratios to restrict update step sizes.
\textbf{DPO}~\citep{rafailov2023direct} analytically parameterizes reward via $r(x, y) = \beta \log \frac{\pi_\theta(y \mid x)}{\pi_{\text{ref}}(y \mid x)} + \beta \log Z(x)$, enforcing the Fisher trust-region implicitly via pairwise.    \textbf{GRPO}~\citep{shao2024deepseekmath} foregoes critic estimation and integrates an unbiased token-level KL penalty $\frac{\pi_{\text{ref}}}{\pi_\theta} - \log \frac{\pi_{\text{ref}}}{\pi_\theta} - 1 \ge 0$ directly into group-relative advantage updates.

However, with proofs deferred to Appendix~\ref{app:theoretical_limits_proofs}, RDA faces a gradient deadlock when applied to SFT. Unlike on-policy RL, where policy $\pi_\theta$ samples candidates strictly within the support set of the reference policy $\pi_{\text{ref}}$ and performs soft probability re-weighting, SFT enforces out-of-distribution distribution jumps via the hard one-hot cross-entropy objective $\mathcal{L}_{\text{SFT}}(\theta) = -\log \pi_\theta(y^* \mid h)$. SFT with the RDA penalty yields the joint objective $\mathcal{L}_{\text{total}} = \mathcal{L}_{\text{SFT}} + \beta D_{\text{KL}}(\pi_\theta \,\|\, \pi_{\text{ref}})$. Differentiating w.r.t. output logits $z \in \mathbb{R}^V$ and setting $\nabla_z \mathcal{L}_{\text{total}} = \mathbf{0}$ approximately yields the theoretical stationary distribution:
\begin{equation}
    \pi_\theta^* = \frac{1}{1 + \beta} \mathbf{e}_{y^*} + \frac{\beta}{1 + \beta} \pi_{\text{ref}}
    \label{eq:app_rda_stationary_policy}
\end{equation}
where $\mathbf{e}_{y^*}$ denotes the one-hot target indicator. Any non-zero regularization coefficient $\beta > 0$ algebraically dilutes the gradient thrust toward $\mathbf{e}_{y^*}$ in exact proportion to $\frac{\beta}{1+\beta}$. 

\subsubsection{Paradigm 2: Data Mixture Alignment (DMA)}
\label{subsubsec:data_mixture}

DMA mitigates PCD by assembling a diverse pervasive dataset $\mathcal{D}_{\text{mix}} = \sum_{k=1}^K w_k \mathcal{D}_k$ across heterogeneous tasks, balancing target task adaptation with general capability preservation. This data-centric paradigm is widely adopted across instruction data curation~\citep{yin2023dynosaur,wang2024data}, multi-task fine-tuning~\citep{liu2024moe,mahabadi2021parameter}, and comprehensive model evaluation benchmarks~\citep{xuan2025mmlu,li2025data}.

Through our theoretical framework (Section~\ref{subsec:loss_shift}), DMA  operates as an empirical proxy to simulate the diversity of the pre-training distribution $\mathcal{D}_G$ at the input stage. 
In single-task SFT on narrow data $\mathcal{D}_T$, the activation frequency of task-specific features surges ($1 - S_T \gg 1 - S_G$), shifting loss weight from $L_1$ to higher-order losses $L_N$ ($N \ge 2$). 
By aggregating diverse tasks, DMA enforce high global feature sparsity ($S \to 1$) and suppress joint co-activation probabilities ($P(a_i > 0, a_j > 0) \approx 0$), which keeps most of the weight on the lower-order loss $L_1$.

Key methodologies within this paradigm implement this geometric stabilization through different operational mechanisms:
\textbf{LIMA}~\citep{zhou2023lima} caps the empirical trigger count of any domain-specific feature, keeping absolute feature density $1 - S$ near zero, keeping the weight of higher-order terms small.
\textbf{SMART}~\citep{renduchintala2024smart} maximizes the convex hull volume spanned by the active feature sub-dictionary while removing sample-level collinearity, directly suppressing the emergence of the compounding cross-talk terms $\mathcal{C}_{\text{inter}}^{(N)}$ at the data source.
\textbf{DYNAMIXSFT}~\citep{shin2026dynamixsft}: using a 1-step look-ahead $\Delta$-Loss reward, tracks the immediate marginal return along task-specific feature directions,  to prevent excessive directional drift $\delta h$ along any single task axis.
\textbf{Data Mixing Optimization (DMO)}~\citep{li2025data}: proves that optimal allocations satisfy first-order derivative equality: $\frac{\partial L_1}{\partial w_1} = \dots = \frac{\partial L_K}{\partial w_K}$. Macroscopically, this condition balances the gradient thrust across different feature directions on the loss surface, preventing any single task from exerting a disproportionate displacement that pierces delicate LVD decision hyperplanes.

However, because the sample budget of SFT is orders of magnitude smaller than pre-training ($N_{\text{mix}} \ll N_{\text{pre}}$), the empirical distribution naturally contains vast feature support voids. Given extreme pre-training feature sparsity ($S_i \to 1$), long-tail features with marginal activation probabilities $p_i = 1 - S_i < \frac{1}{N_{\text{mix}}}$ is completely unactivated ($P_{\text{mix}}(a_i > 0) = 0$) and entirely unsupervised. 
Ensuring that $\mathcal{D}_{\text{mix}}$ covers all $k$ latent semantic features with confidence at least $1 - \delta$ requires (Appendix~\ref{app:proof_edm_sample_complexity}):
\begin{equation}
    N_{\text{mix}} = \Omega\left( \frac{\log(1 / \delta)}{\min_{i} (1 - S_i)} \right).
    \label{eq:sample_complexity_bound}
\end{equation}
The requirement is set by the sparsest feature. Because only $c_{\text{act}} = \mathcal{O}(1)$ of the $k$ features are active per token on average, $\min_i (1 - S_i) \le c_{\text{act}} / k$, so the bound is at least $\Omega(k \log(1 / \delta))$ and grows linearly with the number of features. Moreover, if the rarest features were activated only $\mathcal{O}(1)$ times over the pre-training corpus, i.e., $\min_i (1 - S_i) = \mathcal{O}(1 / N_{\text{pre}})$, full coverage requires $N_{\text{mix}} = \Omega(N_{\text{pre}} \log(1 / \delta))$, far beyond any SFT budget.
\subsubsection{Paradigm 3: Parameter-Space Orthogonalization (PSO)}
\label{subsubsec:parameter_orthogonalization}

PSO posits that models fine-tuned on distinct tasks ($A, B, \dots$) from a shared base model $\mathcal{M}_{\theta^0}$ can be directly unified in weight space, serving as one of the core mechanisms in model merging~\citep{li2025deep,yang2026model,yang2024adamerging}. 
Its premise is that inter-task conflicts stem predominantly from adversarial parameter updates and sign disagreements among task vectors ($\Delta \theta_A^\top \Delta \theta_B < 0$). 
Consequently, PSO fuses task updates along mutually orthogonal parameter directions.

Let $W_a, W_b$ denote the parameter representations of features $a$ and $b$. 
Under Hypothesis~\ref{hypo:virtual_sparse_weight},  constraining parameter updates to be orthogonal eliminates this cost:
\begin{equation}
    W_a \cdot W_b = 0 \implies \mathcal{C}_{\text{inter}} \equiv 0,
\end{equation}
severing the pathway through which feature $b$ leaks into the readout channel of feature $a$.

In practice, several works implement PSO via singular value decomposition (SVD) on layer-wise update matrices:
\begin{equation}
    \Delta W_A = U_A \Sigma_A V_A^\top, \quad \Delta W_B = U_B \Sigma_B V_B^\top,
\end{equation}
where $U_A, U_B \in \mathbb{R}^{d_{\text{out}} \times r}$ and $V_A, V_B \in \mathbb{R}^{d_{\text{in}} \times r}$ denote the left and right singular matrices, respectively. 
Decomposing task feature channels into rank-1 singular components $W_a = \sigma_a u_a v_a^\top$ and $W_b = \sigma_b u_b v_b^\top$, their inner product expands as $W_a \cdot W_b = \sigma_a \sigma_b (u_a^\top u_b) (v_a^\top v_b)$, reformulating $\mathcal{C}_{\text{inter}}$ as $2I \left[ \sigma_a \sigma_b (u_a^\top u_b) (v_a^\top v_b) \right]^2$. 
Building on this formulation, \textbf{KnOTS}~\citep{stoica2025model} and \textbf{TSV-Merge}~\citep{gargiulo2025task} enforce orthogonality on the concatenated singular bases ($U_A^\top U_B = \mathbf{0}$ or $V_A^\top V_B = \mathbf{0}$) via shared orthonormal projections or Procrustes whitening, driving cross-task singular overlaps and the resulting interference to zero. 
In addition, \textbf{PCB-Merging}~\citep{du2024parameter} approximates feature importance $I_i$ via intra-task parameter energy ($\tau_i \odot \tau_i$) to adaptively dampen and eliminate parameter interference. 

Although PSO enforces parameter-level decoupling, the translation from parameter orthogonality to activation-space feature orthogonality suffers substantial attenuation. With orthogonal task matrices $\Delta W_A$ and $\Delta W_B$, residual signals must pass through non-linear activation functions $\sigma(\cdot)$ (e.g., GELU or SwiGLU). A second-order multivariate Taylor expansion of the combined post-activation state $h_{\text{out}} = \sigma(z_0 + \delta z_A + \delta z_B)$ isolates a non-vanishing interaction term: 
\begin{equation}
    h_{\text{out}} \approx \sigma(z_0) + \sigma'(z_0)(\delta z_A + \delta z_B) + \frac{1}{2}\sigma''(z_0)(\delta z_A^2 + \delta z_B^2) + \sigma''(z_0)(\delta z_A \odot \delta z_B)
\end{equation}
Even when parameter updates achieve strict singular-basis orthogonality such that the linear cross-activation expectation vanishes ($\mathbb{E}[\delta z_A^\top \delta z_B] = 0$), the non-linear crosstalk component $\sigma''(z_0)(\delta z_A \odot \delta z_B)$ does not vanish in general. This non-linear modulation can generate spurious higher-order cross-task hybrid features in the residual stream. 

\subsubsection{Paradigm 4: Delta-Parameter Pruning (DPP)}
\label{subsubsec:delta_parameter_pruning}

Delta-Parameter Pruning (DPP) is a post-SFT technique that mitigates task conflicts and capability degradation by pruning the majority of the updated parameters $\Delta \theta = \theta - \theta^0$~\citep{pmlr-v235-yu24p,deng2025dare}, widely adopted across model merging~\citep{li2025deep,yang2026model,yang2024adamerging} and capability preservation~\citep{friedrich2016model,luo2025comparative,bao2026geometric}. 
To examine its mechanistic foundation, consider the analytical gradient of task loss $\mathcal{L}_{\mathcal{T}}$ with respect to an arbitrary layer read-in weight matrix $W$ during single-task SFT. 
Let $\mathcal{S}_{\mathcal{T}} \subset \{f_1, \dots, f_k\}$ denote the subset of latent semantic features directly relevant to task $\mathcal{T}$, with cardinality $k_{\mathcal{T}} = |\mathcal{S}_{\mathcal{T}}| \ll k$. According to the feature interference formulation in Eq.~\eqref{eq:interference}, the analytical gradient decomposes into two distinct components:
\begin{equation}
    \nabla_W \mathcal{L}_{\mathcal{T}} = \sum_{f_a \in \mathcal{S}_{\mathcal{T}}} \underbrace{\frac{\partial \mathcal{L}_{\mathcal{T}}}{\partial \hat{a}_a} \frac{\partial \hat{a}_a}{\partial W}}_{\text{ Feature Benefit}} + \sum_{f_a \in \mathcal{S}_{\mathcal{T}}} \sum_{f_j \notin \mathcal{S}_{\mathcal{T}}} \underbrace{\frac{\partial \mathcal{L}_{\mathcal{T}}}{\partial \hat{a}_a} \frac{\partial (f_a^\top f_j)}{\partial W} f_j}_{\text{Interference Cost}},
    \label{eq:gradient_dof_decomposition}
\end{equation}
where $\hat{a}_a$ denotes the recovered activation intensity of target feature $f_a$, and $f_a, f_j \in \mathbb{R}^m$ are unit feature directions in the residual stream. 
Equation~\eqref{eq:gradient_dof_decomposition} reveals an extreme asymmetry in parameter degrees of freedom:
the parameters directly establishing the target feature-token mapping possess only $\mathcal{O}(k_{\mathcal{T}})$ degrees of freedom. However, because activating target feature $f_a$ excites crosstalk across non-target ambient features $f_j \notin \mathcal{S}_{\mathcal{T}}$, gradient descent on the narrow distribution $\mathcal{D}_T$ suppresses these residuals by rotating weights across almost all coupled directions, generating $\mathcal{O}(k_{\mathcal{T}} \cdot k)$ cross-coupling updates.
Because $k \gg k_{\mathcal{T}}$, the ratio of parasitic interference parameters to target mapping parameters scales as $\frac{\mathcal{O}(k_{\mathcal{T}} \cdot k)}{\mathcal{O}(k_{\mathcal{T}})} = \mathcal{O}(k) \gg 1$. Consequently, in unregularized SFT, the vast majority of delta parameters $\Delta \theta$ are devoted to canceling localized crosstalk on $\mathcal{D}_T$ or drifting along unconstrained ambient dimensions, rather than constructing task-specific token representations.

Contemporary DPP methods eliminate these parasitic interference components through distinct strategies:
\textbf{DARE}~\citep{pmlr-v235-yu24p} and \textbf{DAREx}~\citep{deng2025dare} randomly drop delta parameters with a high ratio $p \in [0.9, 0.99]$ and rescale remaining weights, eliminating the vast majority of parasitic interference cost in mathematical expectation.
\textbf{Model Tailor}~\citep{pmlr-v235-zhu24l} selects an ultra-sparse patch based on saliency and second-order sensitivity to directly constrain the representation displacement on non-target features.

However, DPP is fundamentally bounded by an intrinsic trade-off: under extreme feature sparsity, the activation perturbation variance induced by pruning mask diverges inversely with the retention rate~\citep{deng2025dare}, scaling as:
\begin{equation}
    \operatorname{Var}(h^{\text{diff}}) \propto \frac{p}{1-p} \sum_{j} (\Delta W_{ij} a_j)^2,
    \label{eq:dpp_variance_scaling}
\end{equation}
where $p \in [0, 1)$ denotes the parameter drop rate. 
Crucially, this trade-off exposes the fundamental theoretical ceiling of DPP: \textit{feature benefit} and \textit{interference cost} cannot be decoupled from the perspective of DPP. Under feature superposition, physical parameter channels couple simultaneously to target semantic features and ambient pervasive directions.  Because parameter pruning acts as an unstratified coordinate truncation rather than a geometric disentanglement of feature manifolds, it cannot fully extinguish pervasive interference without compromising representation stability.

\subsubsection{Paradigm 5: Gradient Orthogonalization Optimization (GOO)}
\label{subsubsec:gradient_orthogonalization}

GOO posits that preserving pervasive capabilities fundamentally requires directional gradient compatibility: if the gradient of task $A$ $\mathbf{g}_A = \nabla_\theta \mathcal{L}_A$ is non-orthogonal to the task-$B$ gradient $\mathbf{g}_B = \nabla_\theta \mathcal{L}_B$ ($\mathbf{g}_A^\top \mathbf{g}_B \neq 0$), parameter updates $\Delta W$ will inevitably be driven along trajectories that escalate feature interference between task $A$ and $B$~\citep{lopez2017gradient,chaudhry2018efficient,yu2020gradient}. To prevent this, GOO constrains or projects parameter updates onto subspaces orthogonal to another task's gradients, a mechanism widely adopted across continual learning~\citep{yang2025recent,zheng2025towards}, multi-task learning~\citep{liu2024moe,mahabadi2021parameter}, and PEFT~\citep{xu2026parameter,wang2025parameter}.

To uncover the mechanistic link between gradient conflict and feature interference, we examine the gradient vector field governing the task $A$ on $\mathcal{D}_A$. For simplicity, differentiating the  loss $\mathcal{L}_A$ with respect to an arbitrary latent feature direction vector $f_j \in \mathbb{R}^m$ in the 1-sparse conditional loss yields two distinct vector fields:
\begin{equation}
    \mathbf{g}_{A, j} = \nabla_{f_j} \mathcal{L}_A = \underbrace{-4 I_j \left(1 - \|f_j\|_2^2\right) f_j}_{\mathbf{g}_{\text{mag}, j} \text{ (Magnitude Restoring Force)}} + \underbrace{2 \sum_{i \neq j} (I_j + I_i) (f_j^\top f_i) f_i}_{\mathbf{g}_{\text{orth}, j} \text{ (Orthogonal Repulsive Force)}},
    \label{eq:gradient_goo_forces}
\end{equation}
where $\mathbf{g}_{\text{mag}, j}$ penalizes deviations from unit norm $\|f_j\|_2 \approx 1$ to maintain feature benefit, and $\mathbf{g}_{\text{orth}, j}$ acts as an repulsive force on $\Delta W^{B}_j$ to push $f_j$ away from non-orthogonal features $f_i$ whenever $f_j^\top f_i > 0$\footnote{For algebraic clarity and without loss of generality, our channel-level analysis only focuses on the local acute-conflict case $\mathbf{g}_T^\top \mathbf{g}_{G, j} < 0$ acting on an arbitrary feature channel $j$.} to minimize feature interference. 

Let parameter updates on another task $B$ be $\Delta W^{B} = -\eta \mathbf{g}_B$ with learning rate $\eta > 0$. When a negative gradient inner product occurs along channel $j$ ($\mathbf{g}_{B, j}^\top \mathbf{g}_{A, j} < 0$), taking the inner product in parameter space yields $\langle \Delta W^{B}_j, \mathbf{g}_{A, j} \rangle > 0$. Because the equilibrium in feature benefit term ($\|f_j\|_2 \approx 1$), $\mathbf{g}_{\text{mag}, j} \approx \mathbf{0}$,then leaving $\mathbf{g}_{A, j}$ dominated by  $\mathbf{g}^{A}_{\text{orth}, j}$. Consequently, $\langle \Delta W^{B}_j, \mathbf{g}^{A}_{\text{orth}, j} \rangle > 0$, demonstrating that the update acts in direct opposition to the repulsive restorative force and drives channel $j$ to rotate toward activating and coupling with feature $f_i$. When evaluating an input of task $A$ with activation $h_0$, this adverse channel rotation induces a physical layer perturbation $\delta r_\ell = \Delta W^{B} h_0 \neq \mathbf{0}$, which accumulates through subsequent layers to erode LVD decision margins and trigger discrete prediction flips.

Existing GOO algorithms enforce gradient compatibility through diverse projection operations. \textbf{GEM}~\citep{lopez2017gradient} and \textbf{A-GEM}~\citep{chaudhry2018efficient} solve quadratic programs to project candidate gradients onto half-spaces satisfying $\tilde{g}^\top g_k \ge 0$, preventing updates from pointing toward competitor LVD cells. \textbf{OGD}~\citep{farajtabar2020orthogonal} applies Gram-Schmidt orthogonalization to accumulated task gradients to keep pre-trained feature readouts invariant. Addressing multi-task training, \textbf{PCGrad}~\citep{yu2020gradient} projects each conflicting gradient onto the normal plane of the other, eliminating magnitude shrinkage and lateral parameter drift. In activation space, \textbf{GPM}~\citep{saha2021gradient} computes the SVD to enforce zero-activation crosstalk ($\Delta W_\ell h_0 \approx \mathbf{0}$). Expanding on this, \textbf{FS-DGPM}~\citep{deng2021flattening} and \textbf{TRGP}~\citep{osti_10352359} relax strict null-space projection via trust regions and dynamic soft orthogonalization, accommodating finite hidden dimensions by allowing bounded overlap along minor singular axes.

However, GOO faces unavoidable dimensional exhaustion when tasked with safeguarding pervasive capabilities. The allowable degrees of freedom for gradient projection are fundamentally bottlenecked by the physical hidden dimension $m$ of the residual stream. When SFT progresses sequentially across tasks or domains, accumulating core feature bases $B_t \in \mathbb{R}^{m \times r_t}$ yields a composite constraint matrix $\mathcal{B}_T = [B_1, \dots, B_T] \in \mathbb{R}^{m \times (\sum_{t=1}^T r_t)}$. Once the cumulative rank of protected pervasive features saturates the hidden dimension ($\sum_{t=1}^T r_t \ge m$), the orthogonal complement projection operator collapses to zero: 
\begin{equation}
    P_\perp^{(T)} = I - \mathcal{B}_T (\mathcal{B}_T^\top \mathcal{B}_T)^{-1} \mathcal{B}_T^\top = \mathbf{0}
\end{equation}
Under this condition, the projected gradient vanishes identically ($\mathbf{g}_{\text{proj}} = P_\perp^{(T)} \mathbf{g}_{T} \equiv \mathbf{0}$), causing complete collapse. 

\subsubsection{Paradigm 6: Mechanistic Interpretability and Localization (MIL)}
\label{subsubsec:mechanistic_interpretability}

MIL posits that specific linguistic behaviors, factual knowledge, and reasoning skills are mediated by localized functional sub-networks, attention circuits, or dedicated subsets of neurons rather than uniformly dispersed representations. By isolating task-critical parameter regions, this paradigm restricts parameter updates or edits strictly to identified components while freezing the remaining parameters. This targeted adaptation strategy is widely employed in knowledge editing~\citep{meng2022locating,dai2022knowledge}, machine unlearning~\citep{wu2023depn,li2025effective}, and safety alignment~\citep{yan2026spurious} to modify the mechanism for the target task without destabilizing foundational capabilities.

Within our geometric framework, the efficacy of partial freezing stems from  eliminating unconstrained drift on ambient feature channels. In SFT, mechanistic localization partitions the comprehensive set of parameter channels $\mathcal{N}$ into two disjoint subsets: the task-critical channels $\mathcal{N}_{\mathcal{T}}$ identified via attribution or causal tracing, and the task-irrelevant ambient channels $\mathcal{N}_{\mathcal{P}} = \mathcal{N} \setminus \mathcal{N}_{\mathcal{T}}$. When the adapted model processes a general pervasive input with baseline activation $h_0$, the net forward activation displacement $\Delta h$ decomposes into:
\begin{equation}
    \Delta h = \underbrace{\sum_{i \in \mathcal{N}_{\mathcal{T}}} \Delta W_i h_0}_{\text{Task-Critical Adaptation}} + \underbrace{\sum_{j \in \mathcal{N}_{\mathcal{P}}} \Delta W_j h_0}_{\text{Irrelevant Drift}},
    \label{eq:mi_displacement_decomposition}
\end{equation}
where $\Delta W_i$ and $\Delta W_j$ denote the parameter variations along individual channels. Parameter channels exhibit extreme functional selectivity: channels in $\mathcal{N}_{\mathcal{T}}$ couple predominantly to target task features $\mathcal{S}_{\mathcal{T}}$, whereas channels in $\mathcal{N}_{\mathcal{P}}$ are dedicated almost exclusively to general pervasive features $\mathcal{S}_{\text{ambient}} = \{f_1, \dots, f_k\} \setminus \mathcal{S}_{\mathcal{T}}$. 

Consequently, the Irrelevant Drift term in Eq.~\eqref{eq:mi_displacement_decomposition} acts as the primary destructive force during full SFT, as unconstrained updates across $\mathcal{N}_{\mathcal{P}}$ distort ambient feature representations and generate coherent ghost signal spikes. Mechanistic localization enforces the hard constraint $\Delta W_j \equiv \mathbf{0}$ for all $j \in \mathcal{N}_{\mathcal{P}}$, directly extinguishing this second summation and shielding general features from unintended parameter modifications. 

Contemporary localization techniques implement this geometric shielding through diverse intervention mechanics. DEPN~\citep{wu2023depn} and CLUE~\citep{ICLR2026_5576b86e} trace sensitive attributes via integrated gradients or dual-circuit causal discovery, zeroing out targeted neuron activations during forward inference ($a_j \leftarrow 0, \, \forall j \in \mathcal{S}_{\text{privacy}}$), thereby truncating the propagation of unconstrained drift vectors. CA-LoRA~\citep{pmlr-v267-wang25ak} detects inter-layer communication shifts via circuit edge rewiring rates, adaptively assigning low-rank adaptation budgets to restructured layers while freezing stable functional paths. Furthermore, gradient- and magnitude-based attribution frameworks such as WAGLE~\citep{jia2024wagle} evaluate parameter sensitivity w.r.t.  target feature channels, restricting gradient updates exclusively to high-attribution weights to eliminate crosstalk excitation on dormant pervasive features. 

However, MIL presumes that neural behaviors are compartmentalized into modular, cleanly separable parameter subsets, an assumption in tension with feature superposition under $m \ll k$. Superposition forces physical neurons and weight channels to be polysemantic, functioning as composite channels:
\begin{equation}
    W_j = \sum_{a \in \mathcal{N}_{\mathcal{T}}} c_{j, a} f_a + \sum_{l \in \mathcal{N}_{\mathcal{P}}} c_{j, l} f_l
\end{equation}
where the total ambient coupling energy $\sum_{l \in \mathcal{N}_{\mathcal{T}}} |c_{j, l}|^2 > 0$ is non-zero. Consequently, no parameter subset $\Theta_{\mathcal{T}}$ can be localized that isolates target capabilities without coupling to ambient general features. When adapting the localized parameters ($\Delta \theta_j \neq 0, \, \forall j \in \Theta_{\mathcal{T}}$) to master task $\mathcal{T}$, these polysemantic weights undergo unavoidable lateral rotations, inducing non-zero residual collateral leakage. 

\subsubsection{Summary}
\label{sec:paradigm_summary}
 
The six paradigms share a common origin. None of them is derived from the microscopic mechanism of PCD,
namely the interference among superposed features. Instead, each starts from a macroscopic symptom of the conflict between target adaptation
and pervasive capabilities, such as output distribution shift, data imbalance, parameter overlap, or
gradient conflict. Each remedy is therefore effective, 
but its supporting theory holds only locally, under an assumption that superposition violates.
Methods that operate in weight space (PSO, DPP and MIL) implicitly assume that features can be separated
by separating parameters. Methods that operate on data,
gradients, or outputs (DMA, GOO and RDA) instead overlook the two properties that make superposition
possible: sparsity and almost-orthogonality.
 These limitations point to the same missing ingredient: a constraint that acts directly on the carrier of PCD, i.e., the margins of activations against competing LVD facets.

\section{Method}
\subsection{Limitations of Cross-Entropy under Feature Interference}
\label{subsec:ce_limitations}

As established in Sections~\ref{subsec:prelim_lvd} and \ref{sec:mechanistic_theory}, Pervasive Capability Degradation (PCD) stems from pronounced interference among correlated features, manifesting as an anisotropic drift of activations toward certain tokens coupled with these interfering features. 
By the LVD formulation (Eq.~\eqref{eq:lvd_cell}), each token's decision region $\mathrm{Cell}_j^{\mathrm{LVD}}$ is bounded by affine hyperplanes separating it from competing cells $\mathrm{Cell}_k^{\mathrm{LVD}}$:
\begin{equation}
    \Pi_{j, k}: \quad (W_{U, j} - W_{U, k})^\top h + (b_{U, j} - b_{U, k}) = 0.
    \label{eq:lvd_decision_hyperplane}
\end{equation}
This geometry gives a direct correspondence between activation drift and logit dynamics:

\begin{proposition}[Geometric Duality between Activation Drift and Logit Dynamics]
\label{cor:lvd_logit_dynamics}
Let $\hat{u}_{j, k} = \frac{W_{U, k} - W_{U, j}}{\|W_{U, k} - W_{U, j}\|_2}$ be the unit outward normal vector of $\Pi_{j, k}$. Inside $\mathrm{Int}(\mathrm{Cell}_j^{\mathrm{LVD}})$, token $j$ strictly attains the maximal logit across $\mathcal{V}$. Along the directional trajectory $h(t) = h + t \, \hat{u}_{j, k}$ ($t \ge 0$), the relative margin $z_k(h(t)) - z_j(h(t))$ increases strictly monotonically at a constant rate:
\begin{equation}
    \frac{d}{dt} \big( z_k(h(t)) - z_j(h(t)) \big) = \|W_{U, k} - W_{U, j}\|_2 > 0.
    \label{eq:logit_growth_rate}
\end{equation}
At the boundary $\Pi_{j, k}$, $z_j(h(t)) = z_k(h(t))$; once crossing into $\mathrm{Int}(\mathrm{Cell}_k^{\mathrm{LVD}})$, token $k$ strictly dominates all candidates (proof in Appendix~\ref{app:proof_lvd_logit_dynamics}).
\end{proposition}

For a ground-truth token $j$, backpropagating the categorical cross-entropy gradient $\frac{\partial \mathcal{L}_{\text{CE}}}{\partial z_k} = p_k - \mathbf{1}[k = j]$ through $W_U$ exerts an optimization thrust on activation $h$:
\begin{equation}
    \Delta h \propto -\nabla_h \mathcal{L}_{\text{CE}} = (1 - p_j) W_{U, j} - \sum_{k \neq j} p_k W_{U, k},
    \label{eq:ce_gradient_decomposition}
\end{equation}
where $p_j$ and $p_k$ are the softmax probability of label token and competing tokens. 
In the ideal situation, mutually orthogonal unembedding vectors ($W_{U, j}^\top W_{U, k} = 0$) would pull $h$ purely toward $\mathrm{Cell}_j^{\mathrm{LVD}}$ and away from other token cells. However, feature superposition ($k \gg m$) generally introduces sparse interference ($W_{U, k}^\top W_{U, j} > 0$). This projects a parasitic leakage $(1 - p_j)(W_{U, k}^\top W_{U, j})$ onto competing channel $k$, counteracting its gradient self-repulsion $p_k \|W_{U, k}\|_2^2$. A first-order balance of these two terms along channel k suggests the following scaling (Proofs are shown in Appendix~\ref{app:ce_equilibrium_proof}):
\begin{equation}
    p_k^* \propto (1 - p_j) \cdot \frac{W_{U, k}^\top W_{U, j}}{\|W_{U, k}\|_2^2}.
    \label{eq:equilibrium_crosstalk_prob}
\end{equation}
During pre-training on $\mathcal{D}_G$, near-orthogonal packing under $L_1$-weighted pre-training ($|W_{U, k}^\top W_{U, j}| \le \epsilon \ll 1$) keeps $p_k^*$ negligible, maintaining saturating confidence ($p_j \to 1$) well inside $\mathrm{Cell}_j^{\mathrm{LVD}}$.

During SFT on $\mathcal{D}_T$, this balance collapses: task-specific feature density triggers a geometric phase transition (Section~\ref{subsec:loss_shift}) that sharply elevates interference ($W_{U, k}^\top W_{U, j} \gg 0$) across task-coupled tokens. By Eq.~\eqref{eq:equilibrium_crosstalk_prob}, this inflates the competitor probability $p_k^*$ on general queries, diluting true-label confidence and driving $h$ outward toward $\Pi_{j, k}$. For infrequent or long-tail pervasive tokens with narrow margins, this unchecked drift easily breaches the decision boundary, inverting discrete predictions and triggering PCD. While existing post-training paradigms attempt to curb such degradation, each faces fundamental theoretical limitations within this geometry (Section~\ref{subsec:unifying_paradigms}).

In summary, employing standard cross-entropy for post-training SFT suffers from \textbf{two fundamental structural limitations} under feature superposition:
\begin{enumerate}
    \item In cross-entropy, the optimization of ground-truth and the feature interference are inextricably coupled, lacking any independent mechanism to decouple and penalize localized feature interference without perturbing task classification.
    \item Because cross-entropy is predicated on one-hot categorical distributions, sustaining feature sparsity and almost-orthogonal packing relies on the collective supervision of massive, diverse samples,  unattainable within the narrow sample budgets of SFT. 
\end{enumerate}

\subsection{The AEGIS Framework}
\label{subsec:aegis_method}

To resolve these challenges, we introduce the \textbf{AEGIS} (\textbf{A}daptive \textbf{E}quipartition of \textbf{G}eometric \textbf{I}nterference in \textbf{S}uperposition) loss. Instead of relying on discrete categorical probability distributions, AEGIS directly governs directional activation drifts across all token hyperplanes in the unembedding LVD, decoupling feature interference mitigation from task adaptation. Moreover, by supervising activation along multiple directional decision boundaries—including those corresponding to infrequent and long-tail tokens—AEGIS substantially reduces the sample budget required for the pervasiveness dataset. AEGIS consists of two steps: 

\noindent\textbf{Step 1: Adaptive Directional Width.}
Prior to SFT, the pre-trained base model $\mathcal{M}_{\theta^0}$ is frozen. We employ a pervasive anchor dataset $\mathcal{D}_P$ spanning diverse tasks to establish baseline geometric coordinates and directional tolerance bounds.

Let an input sample possess ground-truth label token $j$. The base model generates final residual activation $h_0 \in \mathbb{R}^m$, with unembedding parameters $W_U^{(0)} = [W_{U, 1}^{(0)}, \dots, W_{U, V}^{(0)}] \in \mathbb{R}^{m \times V}$ and $b_U^{(0)} \in \mathbb{R}^V$. For any competing candidate $k \in \mathcal{V} \setminus \{j\}$, the decision boundary $\Pi_{j, k}$ in the LVD possesses outward normal direction $W_{U, j}^{(0)} - W_{U, k}^{(0)}$. To eliminate scale discrepancies caused by activation norms and heterogeneous token embedding magnitudes, we define the \textbf{normalized decision margin} of token $j$ relative to token $k$:
\begin{equation}
    \rho_{j, k}^{(0)} = \frac{z_j^{(0)}(h_0) - z_k^{(0)}(h_0)}{\|W_{U, j}^{(0)} - W_{U, k}^{(0)}\|_2 \|h_0\|_2} = \frac{(W_{U, j}^{(0)} - W_{U, k}^{(0)})^\top h_0 + (b_{U, j}^{(0)} - b_{U, k}^{(0)})}{\|W_{U, j}^{(0)} - W_{U, k}^{(0)}\|_2 \|h_0\|_2}.
    \label{eq:normalized_margin_base}
\end{equation}
Geometrically, $\rho_{j, k}^{(0)}$ measures the normalized angular projection between the activation vector and the decision hyperplane. When $b_U^{(0)} = \mathbf{0}$, $\rho_{j, k}^{(0)}$ strictly evaluates to the directional cosine:
$\rho_{j, k}^{(0)} = \cos \angle (W_{U, j}^{(0)} - W_{U, k}^{(0)}, h_0) \in [-1, 1]$. 
A strictly positive baseline margin ($\rho_{j, k}^{(0)} > 0$) guarantees that $h_0$ resides within the safe interior of $\mathrm{Cell}_j^{\mathrm{LVD}}$.

Because semantic sensitivity varies across contexts and candidate tokens, enforcing a rigid spherical bound risks over-constraining the model. We therefore construct an \textbf{adaptive directional width} $w_{j, k}$ tailored to each competing candidate $k$:
\begin{equation}
    w_{j, k} = \mathbb{E}_{k \in \mathcal{V} \setminus \{j\}} \left[ \rho_{j, k}^{(0)} \right] \cdot \exp \left( - \frac{z_{k^*}^{(0)} - z_k^{(0)}}{\tau} \right),
    \label{eq:tolerance_width}
\end{equation}
where $\mathbb{E}_{k \in \mathcal{V} \setminus \{j\}} [\rho_{j, k}^{(0)}]$ characterizes the baseline average geometric safety buffer across all competing directions, and the exponential factor weights semantic proximity ($k^* = \arg\max_{k \neq j} z_k^{(0)}$ represents the strongest non-target competitor, and $\tau > 0$ is a temperature hyperparameter). This formulation allocates wider exploratory space to semantical competitors while imposing tight bounds against irrelevant tokens.

\noindent\textbf{Step 2: Dynamic Margin Drift Regularization.}
During SFT, model parameters update to $\theta$, yielding current hidden activation $h$, logits $z$, and instantaneous normalized margins:
\begin{equation}
    \rho_{j, k} = \frac{z_j(h) - z_k(h)}{\|W_{U, j} - W_{U, k}\|_2 \|h\|_2}.
    \label{eq:normalized_margin_current}
\end{equation}
Let $\Delta \rho_{j, k} = \rho_{j, k} - \rho_{j, k}^{(0)}$ denote the net margin shift along candidate channel $k$. To formulate the directional constraint, we define the composite violation distance:
\begin{equation}
    D_{j, k}(\zeta) = \max \Big( |\Delta \rho_{j, k} - \zeta| - w_{j, k}, \; -\rho_{j, k}, \; 0 \Big),
    \label{eq:composite_violation_distance}
\end{equation}
where $\zeta \ge 0$ is a position-shared scalar. Equation~\eqref{eq:composite_violation_distance} incorporates two complementary defense mechanisms:
\begin{itemize}
    \item \textbf{Directional Drift Suppression ($|\Delta \rho_{j, k} - \zeta| - w_{j, k}$)}: This term restricts the activation drift along channel $k$ beyond the width $w_{j, k}$. Introducing the shared scalar $\zeta \ge 0$ absorbs uniform, collective margin expansion across candidates, preventing optimization from penalizing shared representational improvements.
    \item \textbf{Hyperplane Crossing Barrier ($-\rho_{j, k}$)}: If activation $h$ approaches or breaches the decision hyperplane $\Pi_{j, k}$ ($\rho_{j, k} < 0$), this term enforces an unyielding repulsive penalty, preventing discrete token prediction flips.
\end{itemize}

To ensure $C^1$-smooth differentiability and stabilize gradient magnitudes under large deviations, we apply the Huber loss operator $\ell_\kappa(\cdot)$ to the violation distance:
\begin{equation}
    \ell_\kappa(D) = \begin{cases} 
        \frac{1}{2\kappa} D^2, & 0 \le D \le \kappa, \\ 
        D - \frac{\kappa}{2}, & D > \kappa, 
    \end{cases}
    \label{eq:huber_loss}
\end{equation}
where $\kappa > 0$ is a fixed smoothing threshold hyperparameter that ensures continuous gradient transitions across boundaries and avoids gradient instability.
For an activation state with fixed parameters and activations, the local AEGIS objective parameterized by scalar $\zeta$ is expressed as:
\begin{equation}
    \mathcal{L}_{\text{AEGIS}}(\zeta) = \frac{1}{|\mathcal{V}| - 1} \sum_{k \neq j} \ell_\kappa \big( D_{j, k}(\zeta) \big).
    \label{eq:aegis_local_loss}
\end{equation}
Because $D_{j, k}(\zeta)$ is convex in $\zeta$ (as the pointwise supremum of convex and affine functions) and $\ell_\kappa(\cdot)$ is convex and non-decreasing on $[0, +\infty)$, $\mathcal{L}_{\text{AEGIS}}(\zeta)$ is strictly a one-dimensional convex function over $\zeta \in [0, +\infty)$. Given that absolute-value junctions produce non-differentiable knots, the global optimum is resolved efficiently via the monotonicity of its right derivative $D^+ \mathcal{L}_{\text{AEGIS}}(\zeta)$:
\begin{equation}
    \zeta^* = \arg\min_{\zeta \ge 0} \mathcal{L}_{\text{AEGIS}}(\zeta).
    \label{eq:optimal_zeta}
\end{equation}
Evaluating at $\zeta^*$, the instance regularization evaluates to $\mathcal{L}_{\text{AEGIS}} = \mathcal{L}_{\text{AEGIS}}(\zeta^*)$. By the envelope theorem~\citep{vandenberghe2004convex}, parameter gradients backpropagate directly through $\zeta^*$ without requiring higher-order unrolling. The specific proofs and searching algorithm are shown in Appendix~\ref{app:zeta_optimization_proof}. 

The composite post-training objective unifies downstream task adaptation and pervasive geometric stabilization:
\begin{equation}
    \mathcal{L}_{\text{total}} = \mathcal{L}_{\text{CE}}(\mathcal{D}_T; \theta) + \mathcal{L}_{\text{CE}}(\mathcal{D}_P; \theta) + \lambda \mathcal{L}_{\text{AEGIS}}(\mathcal{D}_P; \theta),
    \label{eq:total_objective}
\end{equation}
where  $\lambda > 0$ balances geometric margin defense to suppress anisotropic feature interference spikes and safeguard pervasive capabilities. Although $\mathcal{L}_{\text{AEGIS}}$ is evaluated on output margins, its gradient is back-propagated through all layers, so it regularizes every upstream parameter whose update contributes to the drift $\delta h$. 

\begin{wrapfigure}{r}{0.45\linewidth}
  \vspace{-12pt}
  \centering
  \includegraphics[width=\linewidth]{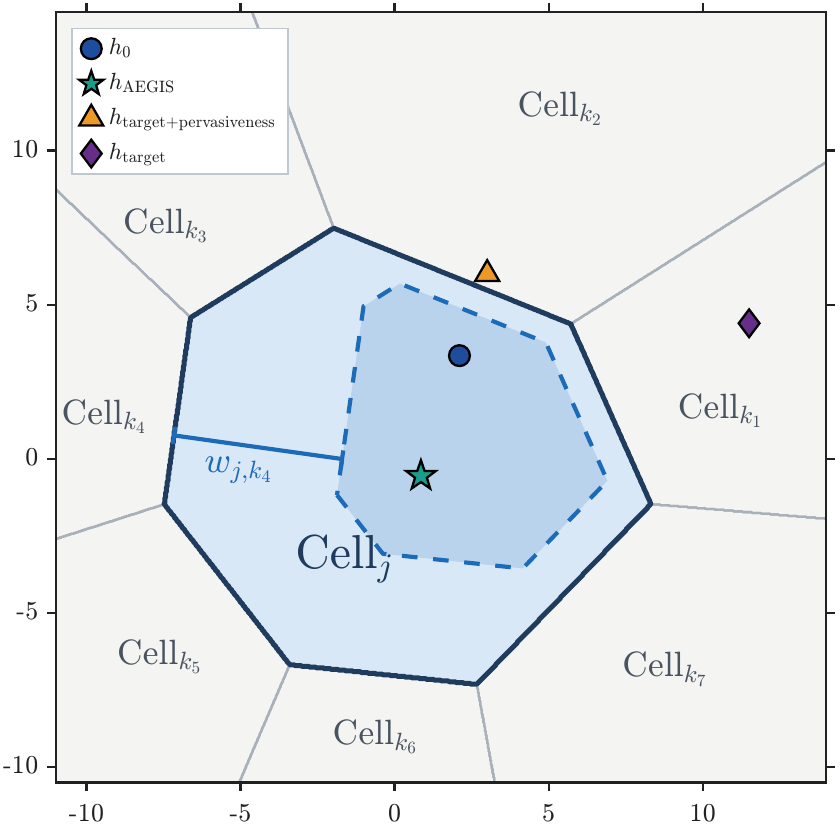}
  \caption{
  In the unembedding space, $\mathrm{Cell}_j$ (\textit{light blue}) is bounded by hyperplanes separating token $j$ from competitors $k_1 \sim k_7$, where the \textit{blue dashed contour} outlines the adaptive directional width derived from baseline activation $h_0$.}
  \label{fig:lvd_polyhedral_dynamics}
  \vspace{-10pt}
\end{wrapfigure}
A two-dimensional geometric illustration of our framework is presented in Figure~\ref{fig:lvd_polyhedral_dynamics}. Without interference mitigation ($\mathcal{L}_{\mathrm{CE}}(\mathcal{D}_T)$) or due to insufficient sample coverage ($\mathcal{L}_{\mathrm{CE}}(\mathcal{D}_T) + \mathcal{L}_{\mathrm{CE}}(\mathcal{D}_P)$), activations incur varying degrees of boundary violation ($h_{\mathrm{target}}$ and $h_{\mathrm{target+pervasiveness}}$).
  In contrast, \textbf{AEGIS} (Eq.~\ref{eq:total_objective}) enforces geometric width constraints to effectively suppress interference, confining $h_{\mathrm{AEGIS}}$ within safe bounds.
Compared to RDA, AEGIS avoids this deadlock by allowing the pre-trained distribution $\mathcal{D}_G$ and downstream task distribution $\mathcal{D}_T$ to co-exist within directional tolerance bounds. 
Compared to DMA, multi-directional hyperplane supervision substantially reduces the required pervasive sample volume. 
Compared to GOO, it directly supports the intrinsic almost-orthogonal packing of pre-training representations, avoiding constructing limited orthogonal update directions. 
Finally, compared to PSO, DPP, and MIL, AEGIS regularizes representations directly in activation space, circumventing the non-transitivity orthogonality and the polysemantic coupling in weight space.

\section{Experiments}
\label{sec:experiments}

\subsection{Experimental Setup}
\label{subsec:exp_setup}

\noindent\textbf{Models.} 
We evaluate our method across three model families spanning multiple parameter scales: Mistral-7B, Qwen2.5 (3B, 7B, 32B, and 72B), and Gemma-2-9B.

\noindent\textbf{Datasets.} 
The training data consists of a target task dataset $\mathcal{D}_T$ and a pervasive dataset $\mathcal{D}_P$. 
For $\mathcal{D}_T$, we adopt the eight datasets from the GLUE benchmark (CoLA, SST-2, MRPC, STS-B, QQP, MNLI, QNLI, and RTE) for natural language understanding (NLU) task, GSM8K for math task, CodeAlpaca, Code-Bagel-Hermes, Infinity-Instruct-Code for code generation task, Evol-Instruct, and LogiQA2 for instruction following task. 
For $\mathcal{D}_P$, we construct $\mathcal{D}_P$ by sampling from the underlying FLAN 2022 collection, which comprises 1,840 tasks across five sub-mixtures (FLAN 2021, T0, NIV2, CoT, and Dialog). 
Specifically, we randomly permute the constituent tasks and iteratively select instances containing previously uncovered label tokens.

\noindent\textbf{Evaluation Metrics.} 
For GLUE, we report the Matthews correlation coefficient (MCC) for CoLA, Pearson and Spearman correlations for STS-B, accuracy and F1 score for MRPC and QQP, matched accuracy for MNLI, and accuracy for SST-2, QNLI, and RTE. 
We evaluate GSM8K and LogiQA2 on their respective test splits using zero-shot accuracy. 
Checkpoints trained on CodeAlpaca are evaluated on HumanEval using pass@1, pass@5, and pass@10, whereas those trained on Code-Bagel-Hermes and Infinity-Instruct-Code are evaluated on CodeXGLUE using token accuracy. 
Checkpoints trained on Evol-Instruct are evaluated on AlpacaEval via win rate. 
To measure pervasive capability preservation, we evaluate models on a held-out pervasive evaluation suite $\mathcal{D}_E$, consisting of 44 benchmarks run with the \texttt{lm-evaluation-harness} framework and report the average accuracy (P-Acc.). $\mathcal{D}_E$ is never used in training.

\noindent\textbf{Baselines.} 
We compare AEGIS against six representative baselines corresponding to each post-training paradigm analyzed in Section~\ref{subsec:unifying_paradigms}: PSFT~\cite{ICLR2026_da75d2bb} for RDA paradigm, SMART~\cite{renduchintala2024smart} for DMA paradigm, KnOTS~\cite{stoica2025model} for PSO paradigm, DARE~\cite{pmlr-v235-yu24p} for DPP paradigm, DiFT~\cite{yuan2026differential} for GOO paradigm, and CLUE~\cite{ICLR2026_5576b86e} for MIL paradigm.

Comprehensive details regarding dataset splits, prompt templates, baseline configurations, the pervasive dataset sampling algorithm, and sample and token statistics of $\mathcal{D}_P$ are deferred to Appendix~\ref{app:exp_details}.

\subsection{Main Results}
\label{subsec:main_results}

We evaluate AEGIS and representative methods from each post-training paradigm on single-target-task adaptation, measuring both target task performance via task-specific metrics and pervasive capability retention via the average accuracy (\textbf{P-Acc.}) across $\mathcal{D}_E$. 
We establish our primary \textbf{Benchmark} (Vanilla) as full-parameter fine-tuning that jointly optimizes standard cross-entropy loss over both the target task dataset $\mathcal{D}_T$ and the pervasive dataset $\mathcal{D}_P$. 
Building upon this benchmark setting, each baseline paradigm incorporates its core mechanism as follows: 
\textbf{PSFT} replaces the cross-entropy loss with KL-divergence regularization; 
\textbf{SMART} substitutes $\mathcal{D}_P$ with its own submodular-curated general task mixture; 
\textbf{KnOTS} fine-tunes the model exclusively on $\mathcal{D}_T$ before merging it with the pre-trained base model; 
\textbf{DARE} applies delta-parameter pruning and merging after SFT completion; 
\textbf{DiFT} replaces standard cross-entropy optimization with its differential parameter updating framework; 
and \textbf{CLUE} localizes and freezes 50\% of the model parameters prior to SFT.

Table~\ref{tab:main_nlu} reports the results of Mistral-7B, Qwen2.5-7B, and Gemma-2-9B across the eight NLU tasks in the GLUE benchmark. 
Table~\ref{tab:main_reasoning} presents the corresponding performance of these three models on mathematical reasoning (GSM8K), code generation (CodeAlpaca, Code-Bagel-Hermes, and Infinity-Instruct-Code), and logic and instruction-following (Evol-Instruct and LogiQA2) tasks. All experiments are conducted over five random seeds; the variance across seeds is consistently below $0.03$ with no statistically significant impact on the results and is thus omitted for clarity.

\begin{table*}[t]
\centering
\caption{Main results on NLU tasks.}
\label{tab:main_nlu}
\resizebox{\textwidth}{!}{%
\begin{tabular}{llcccccccccccccccc}
\toprule
\multirow{2}{*}{\textbf{Model}} & \multirow{2}{*}{\textbf{Method}} & \multicolumn{2}{c}{\textbf{CoLA}} & \multicolumn{2}{c}{\textbf{SST-2}} & \multicolumn{2}{c}{\textbf{MRPC}} & \multicolumn{2}{c}{\textbf{STS-B}} & \multicolumn{2}{c}{\textbf{QQP}} & \multicolumn{2}{c}{\textbf{MNLI}} & \multicolumn{2}{c}{\textbf{QNLI}} & \multicolumn{2}{c}{\textbf{RTE}} \\
\cmidrule(lr){3-4} \cmidrule(lr){5-6} \cmidrule(lr){7-8} \cmidrule(lr){9-10} \cmidrule(lr){11-12} \cmidrule(lr){13-14} \cmidrule(lr){15-16} \cmidrule(lr){17-18}
 & & MCC & P-Acc. & Acc. & P-Acc. & Acc./F1 & P-Acc. & Corr. & P-Acc. & Acc./F1 & P-Acc. & M-Acc. & P-Acc. & Acc. & P-Acc. & Acc. & P-Acc. \\
\midrule
\multirow{8}{*}{\textbf{Mistral-7B}} 
 & Vanilla &0.435 &0.512 &0.965 &0.519 &0.884 &0.493 &0.875 &0.522 &0.905 &0.521 &0.884 &0.524 &0.935 &0.537 &0.835 &0.527 \\
 & PSFT       &0.449 &0.514 &0.977 &0.528 &0.891 &0.485 &0.886 &0.518 &0.892 &0.515 &0.892 &0.519 &0.939 &0.542 &0.854 &0.525 \\
 & SMART     &0.447 &0.519 &0.971 &0.537 &0.889 &0.491 &0.881 &0.526 &0.906 &0.524 &0.888 &0.514 &0.928 &0.539 &\textbf{0.869} &0.534 \\
 & KnOTS     &0.385 &0.437 &0.972 &0.462 &0.871 &0.473 &0.864 &0.507 &0.875 &0.508 &0.867 &0.492 &0.914 &0.504 &0.812 &0.519 \\
 & DARE      &0.398 &0.463 &0.947 &0.465 &0.865 &0.466 &0.859 &0.501 &0.869 &0.486 &0.871 &0.488 &0.905 &0.511 &0.819 &0.504 \\
 & DiFT      &0.459 &0.502 &0.977 &0.522 &0.895 &0.489 &0.873 &0.523 &0.897 &0.519 &0.879 &0.521 &0.942 &0.538 &0.849 &0.537 \\
 & CLUE      &0.446 &0.528 &0.969 &0.539 &\textbf{0.909} &0.494 &0.877 &0.528 &0.908 &0.528 &0.891 &0.529 &0.946 &0.536 &0.852 &0.535 \\
 & \textbf{AEGIS} &\textbf{0.456} &\textbf{0.548} &\textbf{0.979} &\textbf{0.543} &0.902 &\textbf{0.511} &\textbf{0.897} &\textbf{0.531} &\textbf{0.912} &\textbf{0.534} &\textbf{0.895} &\textbf{0.537} &\textbf{0.947} &\textbf{0.549} &0.868 &\textbf{0.541} \\
\midrule
\multirow{8}{*}{\textbf{Qwen2.5-7B}} 
 & Vanilla &0.612 &0.605 &0.972 &0.679 &0.901 &0.634 &0.842 &0.642 &0.867 &0.675 &0.873 &0.614 &0.924 &0.647 &0.859 &0.667 \\
 & PSFT       &\textbf{0.625} &0.601 &0.978 &0.682 &0.903 &0.629 &0.840 &0.635 &0.878 &0.662 &0.876 &0.612 &0.934 &0.651 &\textbf{0.899} &0.656 \\
 & SMART     &0.611 &0.612 &0.976 &0.679 &0.905 &0.641 &0.851 &\textbf{0.663} &0.869 &0.669 &0.869 &0.611 &0.928 &0.652 &0.876 &0.669 \\
 & KnOTS     &0.582 &0.574 &0.963 &0.626 &0.895 &0.621 &0.823 &0.627 &0.842 &0.638 &0.842 &0.602 &0.906 &0.628 &0.855 &0.637 \\
 & DARE      &0.541 &0.588 &0.947 &0.633 &0.871 &0.614 &0.837 &0.615 &0.839 &0.642 &0.839 &0.598 &0.901 &0.617 &0.842 &0.629 \\
 & DiFT      &0.612 &0.602 &0.979 &0.675 &0.906 &0.640 &0.855 &0.646 &0.872 &0.679 &0.880 &0.612 &0.928 &0.651 &0.881 &0.674 \\
 & CLUE      &0.609 &0.594 &0.981 &0.681 &0.915 &0.652 &0.836 &0.659 &0.873 &0.681 &0.878 &0.618 &0.923 &0.653 &0.879 &0.673 \\
 & \textbf{AEGIS} &0.621 &\textbf{0.618} &\textbf{0.982} &\textbf{0.699} &\textbf{0.921} &\textbf{0.652} &\textbf{0.869} &0.662 &\textbf{0.881} &\textbf{0.685} &\textbf{0.884} &\textbf{0.624} &\textbf{0.934} &\textbf{0.659} &\textbf{0.893} &\textbf{0.678} \\
\midrule
\multirow{8}{*}{\textbf{Gemma-2-9B}} 
 & Vanilla &0.573 &0.548 &0.968 &0.551 &0.895 &0.539 &0.861 &0.564 &0.874 &0.536 &0.894 &0.557 &0.917 &0.539 &0.867 &0.549 \\
 & PSFT       &0.586 &0.552 &0.972 &0.549 &\textbf{0.912} &0.541 &0.872 &0.566 &0.862 &0.547 &0.882 &0.551 &0.905 &0.546 &0.859 &0.551 \\
 & SMART     &0.571 &0.557 &0.957 &0.553 &0.907 &0.535 &0.857 &0.562 &0.875 &0.542 &0.899 &0.564 &0.919 &0.542 &0.872 &0.547 \\
 & KnOTS     &0.556 &0.528 &0.941 &0.516 &0.857 &0.502 &0.829 &0.539 &0.855 &0.529 &0.867 &0.527 &0.889 &0.527 &0.839 &0.529 \\
 & DARE      &0.509 &0.539 &0.905 &0.513 &0.884 &0.514 &0.841 &0.541 &0.851 &0.531 &0.871 &0.533 &0.896 &0.519 &0.835 &0.505 \\
 & DiFT      &0.588 &0.541 &0.978 &0.551 &0.899 &0.539 &0.865 &0.561 &0.869 &0.537 &0.895 &0.569 &0.918 &0.556 &0.877 &0.551 \\
 & CLUE      &0.574 &0.558 &0.967 &0.550 &0.905 &0.546 &0.859 &0.571 &0.878 &0.541 &0.901 &0.565 &0.924 &0.549 &0.874 &0.553 \\
 & \textbf{AEGIS} &\textbf{0.595} &\textbf{0.565} &\textbf{0.982} &\textbf{0.561} &0.907 &\textbf{0.551} &\textbf{0.879} &\textbf{0.579} &\textbf{0.883} &\textbf{0.552} &\textbf{0.909} &\textbf{0.571} &\textbf{0.928} &\textbf{0.559} &\textbf{0.882} &\textbf{0.556} \\
\bottomrule
\end{tabular}%
}
\end{table*}

\begin{table*}[t]
\centering
\caption{Main results on mathematical reasoning, code generation, and logic \& instruction-following tasks.}
\label{tab:main_reasoning}
\resizebox{\textwidth}{!}{%
\begin{tabular}{llcccccccccccccc}
\toprule
\multirow{2}{*}{\textbf{Model}} & \multirow{2}{*}{\textbf{Method}} & \multicolumn{2}{c}{\textbf{GSM8K}} & \multicolumn{4}{c}{\textbf{CodeAlpaca}} & \multicolumn{2}{c}{\textbf{Code-Bagel-Hermes}} & \multicolumn{2}{c}{\textbf{Infinity-Instruct-Code}} & \multicolumn{2}{c}{\textbf{Evol-Instruct}} & \multicolumn{2}{c}{\textbf{LogiQA2}} \\
\cmidrule(lr){3-4} \cmidrule(lr){5-8} \cmidrule(lr){9-10} \cmidrule(lr){11-12} \cmidrule(lr){13-14} \cmidrule(lr){15-16}
 & & Acc. & P-Acc. & Pass@1 & Pass@5 & Pass@10 & P-Acc. & Token Acc. & P-Acc. & Token Acc. & P-Acc. & Win Rate & P-Acc. & Acc. & P-Acc. \\
\midrule
\multirow{8}{*}{\textbf{Mistral-7B}} 
 & Vanilla &0.375 &0.391 &0.085 &0.131 &0.145 &0.373 &0.445 &0.381 &0.488 &0.421 &0.467 &0.423 &0.612 &0.445 \\
 & PSFT       &0.374 &0.399 &0.084 &0.134 &0.151 &0.396 &0.416 &0.385 &0.495 &0.422 &0.461 &0.421 &0.618 &0.432 \\
 & SMART     &0.403 &0.396 &0.098 &0.141 &0.151 &0.378 &0.394 &0.389 &0.492 &0.415 &0.472 &0.425 &0.614 &0.438 \\
 & KnOTS     &0.326 &0.357 &0.064 &0.117 &0.123 &0.334 &0.384 &0.354 &0.457 &0.372 &0.455 &0.394 &0.547 &0.405 \\
 & DARE      &0.354 &0.361 &0.083 &0.129 &0.144 &0.347 &0.397 &0.329 &0.443 &0.388 &0.439 &0.388 &0.463 &0.411 \\
 & DiFT      &0.389 &0.394 &0.091 &0.137 &0.149 &0.376 &\textbf{0.467} &0.384 &0.495 &0.417 &0.471 &0.429 &0.452 &0.438 \\
 & CLUE      &0.405 &0.401 &0.088 &0.136 &0.151 &0.381 &0.454 &0.383 &0.501 &0.426 &0.474 &0.435 &0.629 &0.449 \\
 & \textbf{AEGIS} &\textbf{0.418} &\textbf{0.429} &\textbf{0.115} &\textbf{0.144} &\textbf{0.156} &\textbf{0.398} &0.463 &\textbf{0.417} &\textbf{0.511} &\textbf{0.439} &\textbf{0.486} &\textbf{0.439 }&\textbf{0.638} &\textbf{0.465} \\
\midrule
\multirow{8}{*}{\textbf{Qwen2.5-7B}} 
 & Vanilla &0.714 &0.668 &0.229 &0.374 &0.638 &0.608 &0.539 &0.607 &0.582 &0.614 &0.524 &0.618 &0.594 &0.634 \\
 & PSFT       &0.728 &0.661 &0.224 &0.387 &0.643 &0.614 &0.567 &0.611 &0.601 &0.621 &0.521 &0.616 &0.638 &0.635 \\
 & SMART     &0.725 &0.676 &0.235 &0.385 &0.643 &0.611 &0.526 &0.616 &0.593 &0.626 &0.529 &0.617 &0.659 &0.624 \\
 & KnOTS     &0.629 &0.628 &0.194 &0.327 &0.555 &0.584 &0.457 &0.557 &0.554 &0.581 &0.487 &0.577 &0.587 &0.628 \\
 & DARE      &0.634 &0.607 &0.207 &0.388 &0.642 &0.656 &0.442 &0.543 &0.546 &0.573 &0.469 &0.592 &0.556 &0.615 \\
 & DiFT      &0.728 &0.677 &0.232 &0.384 &0.657 &0.614 &0.531 &0.597 &0.596 &0.616 &0.526 &0.609 &0.649 &\textbf{0.647} \\
 & CLUE      &0.723 &0.669 &0.244 &0.398 &0.665 &0.629 &0.545 &0.605 &0.584 &0.625 &0.531 &0.621 &0.634 &0.645 \\
 & \textbf{AEGIS} &\textbf{0.739} &\textbf{0.683} &\textbf{0.277} &\textbf{0.416} &\textbf{0.683} &\textbf{0.667} &\textbf{0.579} &\textbf{0.628} &\textbf{0.615} &\textbf{0.638} &\textbf{0.539} &\textbf{0.631} &\textbf{0.715} &0.646 \\
\midrule
\multirow{8}{*}{\textbf{Gemma-2-9B}} 
 & Vanilla &0.755 &0.553 &0.283 &0.379 &0.443 &0.594 &0.587 &0.567 &0.605 &0.601 &0.562 &0.595 &0.684 &0.549 \\
 & PSFT       &0.759 &0.561 &0.287 &0.382 &0.448 &0.608 &0.589 &0.571 &0.611 &\textbf{0.605} &0.569 &0.601 &0.658 &0.553 \\
 & SMART     &0.763 &0.558 &0.295 &0.381 &0.459 &0.601 &0.591 &0.569 &0.608 &0.604 &0.565 &0.608 &0.679 &0.546 \\
 & KnOTS     &0.729 &0.527 &0.241 &0.352 &0.397 &0.557 &0.554 &0.541 &0.568 &0.547 &0.541 &0.562 &0.629 &0.529 \\
 & DARE      &0.718 &0.507 &0.228 &0.329 &0.401 &0.493 &0.531 &0.537 &0.573 &0.569 &0.539 &0.577 &0.633 &0.519 \\
 & DiFT      &0.784 &0.551 &0.295 &0.393 &0.461 &0.587 &0.588 &0.566 &0.612 &0.598 &0.558 &0.593 &0.651 &0.536 \\
 & CLUE      &0.763 &0.562 &0.299 &0.397 &0.468 &0.617 &0.594 &0.579 &0.608 &0.603 &0.574 &0.598 &0.696 &0.554 \\
 & \textbf{AEGIS} &\textbf{0.801} &\textbf{0.584} &\textbf{0.337} &\textbf{0.438} &\textbf{0.496} &\textbf{0.622} &\textbf{0.608} &\textbf{0.597} &\textbf{0.614} &0.602 &\textbf{0.582} &\textbf{0.614} &\textbf{0.703} &\textbf{0.559} \\
\bottomrule
\end{tabular}%
}
\end{table*}

The results in Tables~\ref{tab:main_nlu} and~\ref{tab:main_reasoning} demonstrate that \textbf{AEGIS} significantly outperforms existing methods on the $\mathcal{D}_{E}$, validating the effectiveness of enforcing geometric constraints across all token directions. 
Simultaneously, AEGIS achieves a clear lead on the target tasks, which we attribute to the fact that reducing feature interference alleviates overfitting to the target task; a detailed analysis is provided in Section~\ref{subsec:analysis}. 
Furthermore, we observe that competing baselines lack stability, exhibiting substantial fluctuations in their performance rankings across different tasks, whereas AEGIS consistently remains at the forefront. 
We posit that this stability fundamentally stems from directly mitigating the underlying feature interference. 
Finally, to comprehensively evaluate AEGIS across diverse SFT scenarios, we further report results under Mix-up SFT, parameter-efficient fine-tuning (PEFT), and continual SFT scenarios in Appendix~\ref{app:additional_sft}.

\subsection{Scalability and Costs}
\label{sec:scalability}
 
To examine whether AEGIS remains practical as the model scale grows, we profile the computational and memory costs of full-parameter SFT on GSM8K with Qwen2.5-3B, 7B, 32B, and 72B, comparing AEGIS against the Vanilla benchmark (standard cross-entropy), PSFT, and DiFT. All runs are conducted on NVIDIA A100 GPUs interconnected via NVLink, and parameters, gradients, optimizer states, and computation are all kept in FP32 throughout. We report the following metrics.
(1)~\textit{Total time} (GPU$\cdot$h) is the sum of the preprocessing and training stages, each measured as $\#\text{GPUs} \times \text{time}$. 
(2)~\textit{Total FLOPs} counts model FLOPs only, \emph{i.e.}, matrix multiplications over real tokens. The forward cost of a single sequence of length $s$ is $F_{\text{fwd}}(s) = 2 N_{\text{mm}} s + 4 L m s^{2}$, where $N_{\text{mm}}$ denotes the number of parameters participating in matrix multiplications, $L$ the number of layers.
(3)~\textit{Model FLOPs Utilization (MFU)} is defined as the training model FLOPs divided by $\text{Total time} \times 19.5$~TFLOPS, where 19.5~TFLOPS is the peak FP32 (non-Tensor-Core) throughput of an A100. \textit{Hardware FLOPs Utilization (HFU)} shares the same denominator but counts the hardware FLOPs actually executed, including padding and recomputation. Both metrics lie in $[0, 1]$ with $\text{HFU} \geq \text{MFU}$, while the gap between HFU and MFU reflects redundant computation introduced by the implementation. Consequently, a lower MFU at a comparable FLOPs budget indicates that additional wall-clock time stems from implementation overhead rather than from the algorithm itself.
(4)~\textit{peak memory} is the sum over GPUs of the historical maximum of \texttt{torch.cuda.max\_memory\_allocated}. 
Since the training time of SMART, KnOTS,DARE, and CLUE is nearly identical to that of Vanilla, we do not report their time and FLOPs separately. The results are summarized in Table~\ref{tab:cost}.
 
\begin{table}[t]
\centering
\caption{Time and memory costs of SFT on GSM8K across Qwen2.5 model scales.}
\label{tab:cost}
\resizebox{0.9\textwidth}{!}{%
\begin{tabular}{llccccc}
\toprule
Model & Method & Total Time (GPU$\cdot$h) & Total FLOPs & MFU & HFU & Peak Mem. (GiB)  \\
\midrule
\multirow{4}{*}{Qwen2.5-3B}
 & Vanilla & $10.2$ & 1.69($\times 10^{17}$) & 23.5\% & 39.2\% & 59.9    \\
 & PSFT    & 25.3 & 3.13($\times 10^{17}$) & 19.7\% & 38.4\% & 75.5  \\
 & DiFT    & $3.5 + 12.7 = 26.2$ & 1.75($\times 10^{17}$) & 23.6\% & 42.7\% & 61.2 \\
 & AEGIS   & $0.98 + 20.2 = 21.1$ & 2.13($\times 10^{17}$) & 12.0\% & 20.0\% & 60.8   \\
\midrule
\multirow{4}{*}{Qwen2.5-7B}
 & Vanilla & $35.2$ & 3.85($\times 10^{17}$) & 15.6\% & 26.1\% & 147.7   \\
 & PSFT    & $57.5$ & 6.72($\times 10^{17}$) & 13.6\% & 28.4\% & 199.8  \\
 & DiFT    & $5.2 + 39.5 = 44.7$ & 4.38($\times 10^{17}$) & 17.5\% & 30.3\% & 152.4  \\
 & AEGIS   & $1.92 + 42.5 = 44.4$ & 4.88($\times 10^{17}$) & 15.7\% & 21.8\% & 148.0    \\
\midrule
\multirow{4}{*}{Qwen2.5-32B}
 & Vanilla & $154.1$ & 1.74($\times 10^{18}$) & 16.1\% & 27.3\% & 837   \\
 & PSFT    & $239.8$ & 3.63($\times 10^{18}$) & 14.6\% & 26.2\% & 914  \\
 & DiFT    & $9.6 + 198.7 = 208.3$ & 1.98($\times 10^{18}$) & 17.8\% & 29.6\% & 928  \\
 & AEGIS   & $7.8 + 202.3 = 210.1$ & 2.21($\times 10^{18}$) & 12.4\% & 21.0\% & 844   \\
\midrule
\multirow{4}{*}{Qwen2.5-72B}
 & Vanilla & $361.4$ & 3.88($\times 10^{18}$) & 15.3\% & 26.0\% & 3430    \\
 & PSFT    & $703.5$ & 6.82($\times 10^{18}$) & 14.9\% & 28.3\% & 4375  \\
 & DiFT    & $21.3 + 448.6 = 469.9$ & 3.95($\times 10^{18}$) & 18.4\% & 24.2\% & 3694  \\
 & AEGIS   & $17.1 + 457.5 = 474.7$ & 4.92($\times 10^{18}$) & 12.2\% & 20.8\% & 3458 \\
\bottomrule
\end{tabular}%
}
\end{table}
As shown in Table~\ref{tab:cost}, AEGIS introduces only a moderate overhead relative to existing methods, and this overhead does not grow with model scale.
In terms of time, the total cost of AEGIS is $1.31\times$ and $1.36\times$ that of Vanilla on Qwen2.5-72B and 32B, respectively. Moreover, the preprocessing stage can be reused across different target tasks for the same base model and anchor set, rendering this cost negligible when multiple fine-tuning tasks are conducted.
In terms of computation, the total FLOPs of AEGIS remain consistently at $1.27\times$ those of Vanilla across all scales. 
Finally, the MFU is consistently lower (\emph{e.g.}, $12.2\%$ \emph{vs.}\ $15.3\%$ on Qwen2.5-72B). This suggests that the training time of AEGIS can be further reduced toward that of Vanilla through further engineering optimization of its implementation.

\subsection{Internal Analysis}
\label{subsec:analysis}
To understand why AEGIS improves performance on the pervasive dataset, we examine the intrinsic geometry of the output activations on $\mathcal{D}_E$ after SFT on three representative target tasks, namely RTE (NLU), GSM8K (mathematical reasoning), and CodeAlpaca (code generation), using Qwen2.5-7B.
Following the LVD view, every pervasive token position $t$ with ground-truth token $j_t$ and every competing candidate $k \in \mathcal{V} \setminus \{j_t\}$ form a decision \emph{edge} $(t, k)$. An edge is \emph{crossed} if $\rho_{j_t,k} < 0$. We report the following metrics.
(1)~\textit{Cross.} $N_{\text{cross}} = \sum_{t}\sum_{k \neq j_t} \mathbb{1}\big[\rho_{j_t,k} < 0\big]$ counts all crossed edges on $\mathcal{D}_P$, out of a total of $\#\text{tokens} \times |\mathcal{V}|$ edges.
(2)~\textit{New} $N_{\text{new}} = \sum_{t}\sum_{k \neq j_t} \mathbb{1}\big[\rho^{(0)}_{j_t,k} > 0,\ \rho_{j_t,k} < 0\big]$ counts edges that lie on the correct side under the base model but are crossed after SFT; these crossings are directly induced by target task adaptation.
(3)~\textit{Rec.} $N_{\text{rec}} = \sum_{t}\sum_{k \neq j_t} \mathbb{1}\big[\rho^{(0)}_{j_t,k} < 0,\ \rho_{j_t,k} \geq 0\big]$ counts edges that are crossed under the base model but return to the correct side after SFT. The results are summarized in Table~\ref{tab:crossing}.
 
\begin{table}[t]
\centering
\caption{Decision-edge crossing statistics on the pervasive dataset $\mathcal{D}_E$ after SFT on RTE, GSM8K, and CodeAlpaca. All values are in units of $10^{6}$.All metrics are recorded top 50\% competing tokens over $\mathcal{V}$.}
\label{tab:crossing}
\resizebox{0.85\textwidth}{!}{%
\begin{tabular}{lccccccccc}
\toprule
\multirow{2}{*}{Method} & \multicolumn{3}{c}{RTE} & \multicolumn{3}{c}{GSM8K} & \multicolumn{3}{c}{CodeAlpaca} \\
\cmidrule(lr){2-4} \cmidrule(lr){5-7} \cmidrule(lr){8-10}
 & Cross. & New $\downarrow$ & Rec. & Cross. & New $\downarrow$ & Rec. & Cross. & New $\downarrow$ & Rec. \\
\midrule
Base Model & 18.569 & -- & -- & 18.569 & -- & -- & 18.569 & -- & -- \\
\midrule
Vanilla & 16.668 & 5.700 & 8.659 & 4.373 & 2.321 & 5.748 & 5.915 & 1.941 & 4.516 \\
PSFT    & 16.874 & 5.749 & 8.547 & 4.699 & 2.342 & 5.760 & 5.937 & 1.936 & 4.564 \\
SMART   & 16.649 & 5.798 & 8.967 & 4.830 & 2.336 & 5.722 & 5.925 & 1.957 & 4.529 \\
KnOTS   & 16.782 & 5.830 & 8.296 & 4.667 & 2.487 & 5.348 & 5.916 & 1.963 & 4.262 \\
DARE    & 16.555 & 5.999 & 8.336 & 4.455 & 2.569 & 5.469 & 5.948 & 1.983 & 4.152 \\
DiFT    & 16.639 & 5.764 & 8.865 & 4.698 & 2.375 & 5.743 & 5.915 & 1.947 & 4.568 \\
CLUE    & 16.594 & 5.747 & 8.598 & 4.759 & 2.322 & 5.733 & 5.917 & 1.942 & 4.530 \\
\midrule
AEGIS   & 16.794 & \textbf{4.309} & 8.683 & 3.101 & \textbf{1.546} & 5.764 & 5.924 & \textbf{1.370} & 4.535 \\
\bottomrule
\end{tabular}%
}
\end{table}

As shown in Table~\ref{tab:crossing}, AEGIS does not stand out in terms of the total number of crossed edges or the number of recovered edges. This is expected: since $k \gg m$, features can only be arranged in an almost-orthogonal configuration, so a large fraction of crossings is an intrinsic trade-off of superposition that already exists in the base model and persists regardless of the SFT method. In contrast, AEGIS substantially reduces the number of newly crossed edges on every task, corresponding to relative reductions of $25.7\%$, $35.4\%$, and $29.9\%$ with respect to the average of the baselines. These newly crossed edges are precisely the crossings predicted by Eq.~(20): feature interference induced by target adaptation produces an anisotropic displacement $\delta h$ that pushes activations across the hyperplanes of competing tokens coupled with the target features. 

\begin{figure}[t]
\centering
\begin{subfigure}[b]{0.49\textwidth}
    \centering
    \includegraphics[width=0.49\linewidth]{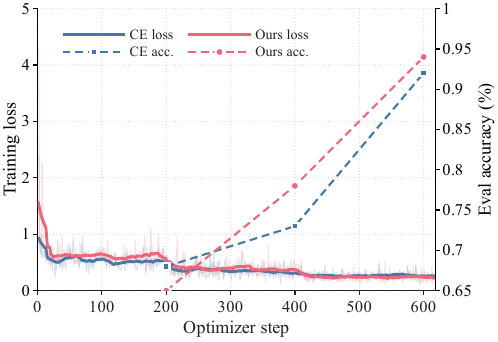}\hfill
    \includegraphics[width=0.49\linewidth]{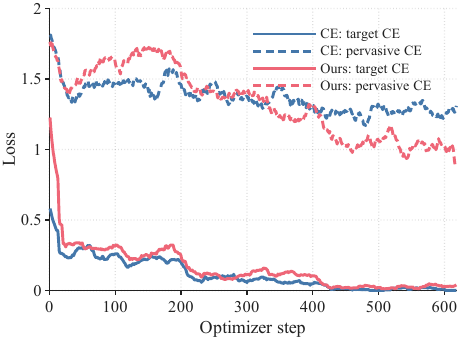}
    \caption{MRPC}
    \label{fig:loss_acc_mrpc}
\end{subfigure}\hfill
\begin{subfigure}[b]{0.49\textwidth}
    \centering
    \includegraphics[width=0.49\linewidth]{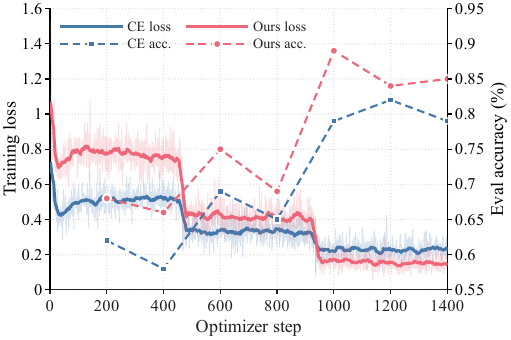}\hfill
    \includegraphics[width=0.49\linewidth]{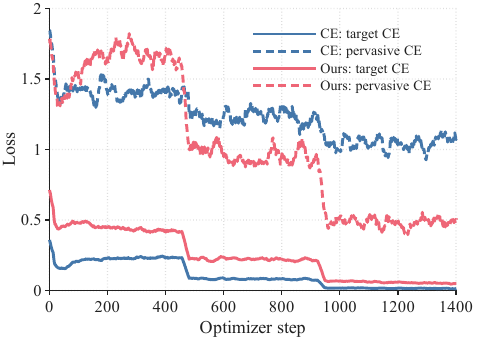}
    \caption{GSM8K}
    \label{fig:loss_acc_gsm8k}
\end{subfigure}
 
\vspace{0.6em}
\begin{subfigure}[b]{0.49\textwidth}
    \centering
    \includegraphics[width=0.49\linewidth]{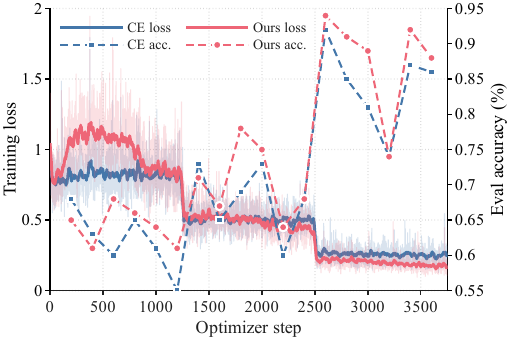}\hfill
    \includegraphics[width=0.49\linewidth]{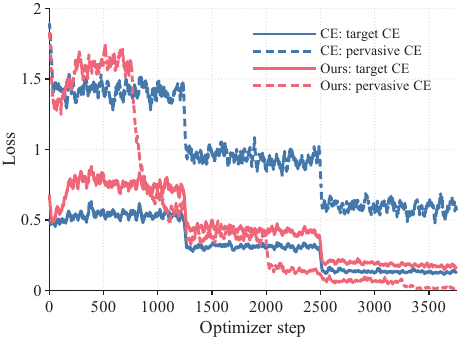}
    \caption{CodeAlpaca}
    \label{fig:loss_acc_codealpaca}
\end{subfigure}\hfill
\begin{subfigure}[b]{0.49\textwidth}
    \centering
    \includegraphics[width=0.49\linewidth]{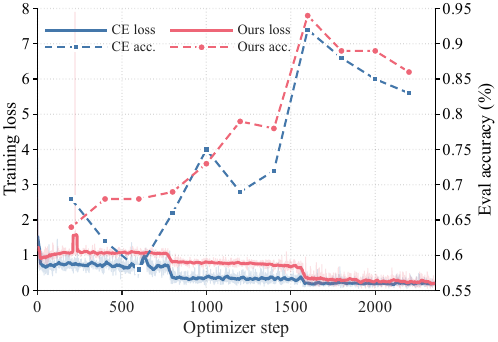}\hfill
    \includegraphics[width=0.49\linewidth]{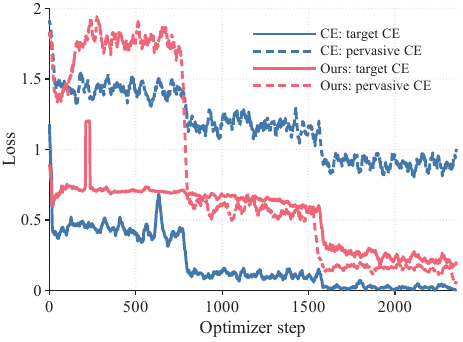}
    \caption{LogiQA2}
    \label{fig:loss_acc_logiqa2}
\end{subfigure}
\caption{Loss and validation accuracy of Qwen2.5-7B during SFT with standard cross-entropy (Vanilla) and AEGIS on MRPC, GSM8K, CodeAlpaca, and LogiQA2. For each task, the left panel shows the total training loss together with the validation accuracy on the target task, and the right panel separates the training loss into the cross-entropy on the target task dataset $\mathcal{D}_T$ and on the pervasive dataset $\mathcal{D}_P$.}
\label{fig:loss_acc}
\end{figure}

Beyond the pervasive dataset, AEGIS also improves target task performance. We hypothesize that this gain arises because reducing feature interference on pervasive activations in turn suppresses overfitting of the target activations. To verify this hypothesis, we examine how the loss relates to validation accuracy for AEGIS and for standard cross-entropy SFT (Vanilla). Figure~\ref{fig:loss_acc} shows the loss and validation accuracy of Qwen2.5-7B over training steps on MRPC, GSM8K, CodeAlpaca, and LogiQA2. AEGIS does not reach a lower target task cross-entropy than Vanilla, and its target cross-entropy is even slightly higher. In contrast, AEGIS substantially lowers the cross-entropy on the pervasive dataset, and its target validation accuracy stays consistently above that of Vanilla throughout training. This indicates that, by reducing feature interference on the pervasive dataset, AEGIS indeed mitigates overfitting to the target task and thus achieves higher validation accuracy at a similar loss level. To further examine which token directions are suppressed by AEGIS, we present examples of the top-10 drift directions in Appendix~\ref{app:drift_directions}.
 
\section{Conclusion}
This paper presents a bottom-up theoretical account of PCD. By coupling feature interference under superposition with the Laguerre–Voronoi geometry of the unembedding layer, the theory traces an unbroken causal chain from activations through weights and gradients to output logits, explaining why narrow-domain SFT inherently induces PCD. Within this framework, we distill existing methods spanning reinforcement learning, continual learning, preference alignment, model merging, and data curation into six paradigms and discuss the limitations of each. We further identify two structural limitations of standard cross-entropy under one-hot supervision: the entanglement of feature interference with task adaptation, and a strong dependence on massive, diverse samples. To address both, we propose the AEGIS loss, which takes the activations of the base model as a reference and imposes adaptive margin-drift constraints along the decision hyperplanes of all competing tokens. Our results show that PCD need not be an irreconcilable trade-off between acquiring new tasks and preserving existing capabilities: once interference can be observed and controlled at the level of microscopic geometry, both can improve together. AEGIS is at once general, scalable, and mechanistically interpretable. Reinforcement learning and preference alignment, model merging, and longer continual-learning sequences is a promising direction for future work.

\bibliography{iclr2025_conference}
\bibliographystyle{abbrvnat}

\newpage
\appendix

\section{Algebraic Equivalence and Directional Transfer between Weights and Features}
\label{app:weight_feature_bridge_proof}

In this appendix, we make the content of Hypothesis~\ref{hypo:virtual_sparse_weight} precise, prove that the virtual parameterization it posits exists for every linearized layer operator, and derive from it the two properties used in the main text, \textbf{Orthogonal Equivalence} and \textbf{Directional Additivity}. Both properties are statements about this virtual parameterization.

\subsection{Forward Functional Consistency and Channel Duality}

Let $h_{\text{in}} \in \mathbb{R}^m$ denote an arbitrary activation vector in the residual stream. Under the feature superposition hypothesis (Definition~\ref{defSuperposition}), $h_{\text{in}}$ decomposes linearly over the overcomplete unit feature dictionary $F = [f_1, \dots, f_k] \in \mathbb{R}^{m \times k}$ ($\|f_i\|_2 = 1$) with activation intensity vector $a = [a_1, \dots, a_k]^\top \in \mathbb{R}^k$:
\begin{equation}
    h_{\text{in}} = F a = \sum_{j=1}^k a_j f_j.
    \label{eq:app_hin_decomp}
\end{equation}
We assume that the dictionary spans the residual stream, i.e., $\operatorname{rank}(F) = m$, which holds for any dictionary in general position since $k \gg m$.

According to Definition~\ref{defSuperposition}, probing the $i$-th latent semantic feature directly within the residual stream corresponds to projecting $h_{\text{in}}$ onto unit direction $f_i$:
\begin{equation}
    \mathcal{R}_i^{\text{ideal}}(h_{\text{in}}) = f_i^\top h_{\text{in}} = a_i (f_i^\top f_i) + \sum_{j \neq i} a_j (f_i^\top f_j) = a_i + \sum_{j \neq i} a_j (f_i^\top f_j).
    \label{eq:app_ideal_probe}
\end{equation}

Under Hypothesis~\ref{hypo:virtual_sparse_weight} (Virtual Sparse Weight Hypothesis), the linearized layer operator $M \in \mathbb{R}^{m \times m}$ factorizes as $M = W_{\text{in}}^{\text{virt}} W_{\text{out}}^{\text{virt}}$, where the read-in operator $W \equiv W_{\text{out}}^{\text{virt}} \in \mathbb{R}^{k \times m}$ reads from the residual stream through $k$ virtual parameter rows and $W_{\text{in}}^{\text{virt}} \in \mathbb{R}^{m \times k}$ writes back into it. For a generic virtual row vector $W_i \in \mathbb{R}^{1 \times m}$, the scalar response of the $i$-th virtual channel to $h_{\text{in}}$ is
\begin{equation}
    \hat{a}_i = W_i h_{\text{in}} = W_i \left( \sum_{j=1}^k a_j f_j \right) = a_i (W_i f_i) + \sum_{j \neq i} a_j (W_i f_j).
    \label{eq:app_channel_readout}
\end{equation}
Equation~\eqref{eq:app_channel_readout} holds for any row vector and does not by itself relate $W_i$ to $f_i$. The relation comes from the requirement in Hypothesis~\ref{hypo:virtual_sparse_weight} that the $i$-th virtual row corresponds to feature $i$, which we now state formally.

When $k > m$, no linear read operator can respond to each feature in isolation: requiring $W_i f_j = 0$ for all $j \neq i$ together with $W_i f_i \neq 0$ for all $i$ would make $W F \in \mathbb{R}^{k \times k}$ a nonsingular diagonal matrix, contradicting $\operatorname{rank}(W F) \le m < k$. Dedication of a channel to a feature therefore cannot mean an isolated response. We instead formalize it through the probe in Eq.~\eqref{eq:app_ideal_probe}: the $i$-th virtual channel is \emph{dedicated to feature $i$} if it satisfies the following two conditions.
\begin{itemize}
    \item[(a)] \emph{Probe selectivity.} The channel response depends on the activation only through the probe value: for all $h, h' \in \mathbb{R}^m$, $\mathcal{R}_i^{\text{ideal}}(h) = \mathcal{R}_i^{\text{ideal}}(h')$ implies $W_i h = W_i h'$. By linearity, this is equivalent to $W_i \delta = 0$ for every $\delta \in \mathbb{R}^m$ with $f_i^\top \delta = 0$.
    \item[(b)] \emph{Positive polarity.} The channel responds positively to its own feature: $W_i f_i > 0$.
\end{itemize}
Condition~(b) excludes silent channels and fixes a sign convention: if $W_i f_i < 0$, flipping the signs of $W_i$ and of the $i$-th column of $W_{\text{in}}^{\text{virt}}$ leaves $M$ unchanged. Conditions~(a)--(b) are the precise meaning we attach to the statement in Hypothesis~\ref{hypo:virtual_sparse_weight} that each virtual row corresponds bijectively to one feature direction. They specify how the $i$-th row acts on activations and do not prescribe the form of $W_i$, which is derived next.

Fix $i$ and let $h_{\text{in}} \in \mathbb{R}^m$ be arbitrary. Since $\|f_i\|_2 = 1$, the activation splits as
\begin{equation*}
    h_{\text{in}} = (f_i^\top h_{\text{in}})\, f_i + \delta, \qquad \delta := h_{\text{in}} - (f_i^\top h_{\text{in}})\, f_i, \qquad f_i^\top \delta = 0.
\end{equation*}
Linearity of the row action and condition~(a) give
\begin{equation*}
    W_i h_{\text{in}} = (f_i^\top h_{\text{in}})\,(W_i f_i) + W_i \delta = s_i\, (f_i^\top h_{\text{in}}), \qquad s_i := W_i f_i > 0,
\end{equation*}
where the positivity of the gain $s_i$ is condition~(b). Evaluating this relation at $h_{\text{in}} = W_i^\top$ yields $\|W_i\|_2^2 = W_i W_i^\top = s_i\, (f_i^\top W_i^\top) = s_i\, (W_i f_i) = s_i^2$, hence $s_i = \|W_i\|_2$. Therefore,
\begin{equation}
    W_i h_{\text{in}} = \|W_i\|_2 \left( f_i^\top h_{\text{in}} \right), \quad \forall h_{\text{in}} \in \mathbb{R}^m.
    \label{eq:app_functional_identity}
\end{equation}
Since Eq.~\eqref{eq:app_functional_identity} holds for all activations in $\mathbb{R}^m$, the linear functionals $h \mapsto W_i h$ and $h \mapsto \|W_i\|_2 f_i^\top h$ coincide. A linear functional on $\mathbb{R}^m$ has a unique representing vector (Riesz representation theorem~\citep{sunder2008riesz}; in finite dimensions it suffices to evaluate Eq.~\eqref{eq:app_functional_identity} on the canonical basis vectors), so the functional identity is equivalent to the structural equivalence:
\begin{equation}
    W_i = \|W_i\|_2 f_i^\top,
    \label{eq:app_riesz_result}
\end{equation}
with $\|W_i\|_2 = s_i > 0$. Substituting Eq.~\eqref{eq:app_riesz_result} into Eq.~\eqref{eq:app_channel_readout} gives $W_i f_j = \|W_i\|_2\, f_i^\top f_j$ and hence
\begin{equation*}
    \hat{a}_i = \|W_i\|_2 \Big( a_i + \sum_{j \neq i} a_j (f_i^\top f_j) \Big) = \|W_i\|_2\, \mathcal{R}_i^{\text{ideal}}(h_{\text{in}}),
\end{equation*}
i.e., a dedicated channel returns the ideal probe up to its gain, including the interference term of Eq.~\eqref{eq:app_ideal_probe}.

Conversely, every nonzero row vector $w \in \mathbb{R}^{1 \times m}$ satisfies conditions~(a)--(b) with respect to exactly one unit direction, namely $w^\top / \|w\|_2$: selectivity and polarity follow from $w = \|w\|_2\, (w^\top / \|w\|_2)^\top$, and uniqueness follows from Eq.~\eqref{eq:app_riesz_result}. A virtual row and the unit direction it probes thus determine each other up to the positive gain $\|w\|_2$. In particular, when a virtual row is modified by a parameter update, the direction probed by the updated channel is obtained by normalizing the updated row. By the Cauchy--Schwarz inequality, Eq.~\eqref{eq:app_riesz_result} also identifies $W_i$ as the matched filter for feature $i$: among all rows of norm $\|W_i\|_2$, it is the unique one with the largest response to $f_i$.

It remains to verify that imposing conditions~(a)--(b) on all $k$ rows is compatible with an exact factorization of $M$, i.e., that the correspondence does not restrict the class of operators that can be represented. Let $s_1, \dots, s_k > 0$ be arbitrary gains, $S = \operatorname{diag}(s_1, \dots, s_k)$, and define
\begin{equation*}
    W_{\text{out}}^{\text{virt}} := S F^\top \in \mathbb{R}^{k \times m}, \qquad
    W_{\text{in}}^{\text{virt}} := M (F F^\top)^{-1} F S^{-1} \in \mathbb{R}^{m \times k},
\end{equation*}
where $F F^\top \in \mathbb{R}^{m \times m}$ is invertible because $\operatorname{rank}(F) = m$. The $i$-th row of $W_{\text{out}}^{\text{virt}}$ is $s_i f_i^\top$, which satisfies conditions~(a)--(b) and Eq.~\eqref{eq:app_riesz_result}, and
\begin{equation*}
    W_{\text{in}}^{\text{virt}} W_{\text{out}}^{\text{virt}} = M (F F^\top)^{-1} F S^{-1} S F^\top = M (F F^\top)^{-1} (F F^\top) = M.
\end{equation*}
Hence every $M \in \mathbb{R}^{m \times m}$ admits a lossless factorization whose read rows are dedicated to the $k$ features. Given the linearization, the existence posited in Hypothesis~\ref{hypo:virtual_sparse_weight} is therefore guaranteed rather than assumed. The $i$-th column of $W_{\text{in}}^{\text{virt}}$ equals $s_i^{-1} M \tilde{f}_i$, where $\tilde{f}_i = (F F^\top)^{-1} f_i$ is the canonical dual frame vector of $f_i$, so that $M h = \sum_{i=1}^k (f_i^\top h)\, M \tilde{f}_i$ for all $h \in \mathbb{R}^m$: channel $i$ reads the probe of feature $i$ and writes along $M \tilde{f}_i$. The construction places no restriction on the directions along which the physical layer reads. All layer-specific structure of $M$ is carried by the write operator, while the read operator depends only on the dictionary $F$ and the gains. The write operator is not unique when $k > m$ and the gains $s_i$ are free; neither affects Eqs.~\eqref{eq:app_functional_identity} and~\eqref{eq:app_riesz_result}, which are the only properties of the virtual parameterization used below. If $\operatorname{rank}(F) < m$, the same construction with $(F F^\top)^{-1}$ replaced by the Moore--Penrose pseudo-inverse reproduces $M$ on $\operatorname{span}(F)$, which contains every activation of the form in Eq.~\eqref{eq:app_hin_decomp}.

\subsection{Proof of Proposition 1: Orthogonal Equivalence}

\begin{proof}
Consider two distinct feature channels $a$ and $b$ ($a, b \in \{1, \dots, k\}$) with virtual weight parameter row vectors $W_a, W_b \in \mathbb{R}^{1 \times m}$. By Eq.~\eqref{eq:app_riesz_result}, their parameter vectors satisfy:
\begin{equation}
    W_a = \|W_a\|_2 f_a^\top, \quad W_b = \|W_b\|_2 f_b^\top.
\end{equation}
The standard Euclidean inner product between the two row vectors is defined as $W_a \cdot W_b = W_a W_b^\top$. Substituting the feature representation:
\begin{equation}
    W_a \cdot W_b = \left( \|W_a\|_2 f_a^\top \right) \left( \|W_b\|_2 f_b^\top \right)^\top = \|W_a\|_2 \|W_b\|_2 \left( f_a^\top f_b \right).
\end{equation}
\end{proof}

\subsection{Proof of Proposition 2: Directional Additivity and Linear Transfer}

\begin{proof}
Let the parameter row corresponding to feature $a$ undergo a perturbation along the channel of feature $b$:
\begin{equation}
    W_a' = W_a + \lambda W_b, \quad \text{where } \lambda \in \mathbb{R}.
\end{equation}
Evaluating the perturbed channel's readout response on an arbitrary activation $h_{\text{in}} \in \mathbb{R}^m$:
\begin{equation}
    \hat{a}_a'(h_{\text{in}}) = W_a' h_{\text{in}} = (W_a + \lambda W_b) h_{\text{in}} = W_a h_{\text{in}} + \lambda W_b h_{\text{in}}.
\end{equation}
Substituting the functional identity from Eq.~\eqref{eq:app_functional_identity}:
\begin{align}
    \hat{a}_a'(h_{\text{in}}) &= \|W_a\|_2 (f_a^\top h_{\text{in}}) + \lambda \|W_b\|_2 (f_b^\top h_{\text{in}}) \nonumber \\
    &= \left( \|W_a\|_2 f_a + \lambda \|W_b\|_2 f_b \right)^\top h_{\text{in}}.
\end{align}
The normalized effective probing direction $f_a'$ realized by the updated channel is obtained by normalizing the functional operator vector:
\begin{equation}
    f_a' = \frac{\|W_a\|_2 f_a + \lambda \|W_b\|_2 f_b}{\|\|W_a\|_2 f_a + \lambda \|W_b\|_2 f_b\|_2} \propto f_a + \lambda \left( \frac{\|W_b\|_2}{\|W_a\|_2} \right) f_b.
\end{equation}

Now, consider the change in mutual crosstalk between feature channel $a$ and feature $b$ after this update. The new inner product evaluates to:
\begin{align}
    (f_a')^\top f_b &\propto \left( f_a + \lambda \frac{\|W_b\|_2}{\|W_a\|_2} f_b \right)^\top f_b \nonumber \\
    &= f_a^\top f_b + \lambda \frac{\|W_b\|_2}{\|W_a\|_2} \|f_b\|_2^2 = f_a^\top f_b + \lambda \frac{\|W_b\|_2}{\|W_a\|_2}.
\end{align}
\end{proof}

\section{Theoretical Proofs for Section~\ref{subsubsec:rl_alignment}}
\label{app:rl_proofs}

\subsection{Derivation of the Induced Fisher Metric in Activation Space}

Let $\mathcal{V} = \{1, \dots, V\}$ denote the output vocabulary. For a given input sequence, the pre-trained model produces final residual activation $h_0 \in \mathbb{R}^m$, yielding reference logits $z_0 = W_U^\top h_0 + b_U \in \mathbb{R}^V$, where $W_U \in \mathbb{R}^{m \times V}$ and $b_U \in \mathbb{R}^V$. The reference token distribution is:
\begin{equation}
    \pi_{\text{ref}}(y = j \mid h_0) = \frac{\exp(z_{0, j})}{\sum_{k=1}^V \exp(z_{0, k})}.
\end{equation}
Following fine-tuning, the hidden activation shifts to $h = h_0 + \Delta h$, inducing updated logits $z = W_U^\top (h_0 + \Delta h) + b_U = z_0 + \Delta z$, where $\Delta z = W_U^\top \Delta h \in \mathbb{R}^V$. The updated policy is $\pi_\theta(y = j \mid h) = \frac{\exp(z_j)}{\sum_{k=1}^V \exp(z_k)}$.

We expand the conditional KL divergence $D_{\text{KL}}(\pi_\theta \,\|\, \pi_{\text{ref}}) = \sum_{j=1}^V \pi_\theta(j) \log \frac{\pi_\theta(j)}{\pi_{\text{ref}}(j)}$ as a multivariate Taylor series w.r.t.  logit shift $\Delta z$ around $\Delta z = \mathbf{0}$:
\begin{equation}
    D_{\text{KL}}(\pi_\theta \,\|\, \pi_{\text{ref}}) = D_{\text{KL}}\Big|_{\Delta z = \mathbf{0}} + \left( \nabla_z D_{\text{KL}}\Big|_{z = z_0} \right)^\top \Delta z + \frac{1}{2} \Delta z^\top \left( \nabla_z^2 D_{\text{KL}}\Big|_{z = z_0} \right) \Delta z + \mathcal{O}(\|\Delta z\|_2^3).
\end{equation}
When $\Delta z = \mathbf{0}$, $\pi_\theta = \pi_{\text{ref}}$, giving $D_{\text{KL}} = 0$. 

\noindent\textbf{First-order derivative:}
Using the relation $\frac{\partial \pi_\theta(k)}{\partial z_j} = \pi_\theta(j) (\delta_{jk} - \pi_\theta(k))$:
\begin{equation}
    \frac{\partial D_{\text{KL}}}{\partial z_j} = \sum_{k=1}^V \frac{\partial \pi_\theta(k)}{\partial z_j} \left( \log \pi_\theta(k) - \log \pi_{\text{ref}}(k) + 1 \right) = \pi_\theta(j) \left( \log \frac{\pi_\theta(j)}{\pi_{\text{ref}}(j)} - D_{\text{KL}}(\pi_\theta \,\|\, \pi_{\text{ref}}) \right).
\end{equation}
Evaluating at $z = z_0$ (where $\pi_\theta(j) = \pi_{\text{ref}}(j)$):
\begin{equation}
    \left. \frac{\partial D_{\text{KL}}}{\partial z_j} \right|_{z = z_0} = \pi_{\text{ref}}(j) \left( \log 1 - 0 \right) = 0 \implies \nabla_z D_{\text{KL}}\Big|_{z = z_0} = \mathbf{0}.
\end{equation}

\noindent\textbf{Second-order derivative (Hessian):}
Differentiating once more w.r.t.  $z_l$:
\begin{equation}
    \frac{\partial^2 D_{\text{KL}}}{\partial z_l \partial z_j} = \frac{\partial \pi_\theta(j)}{\partial z_l} \left( \log \frac{\pi_\theta(j)}{\pi_{\text{ref}}(j)} - D_{\text{KL}} \right) + \pi_\theta(j) \left( \frac{1}{\pi_\theta(j)} \frac{\partial \pi_\theta(j)}{\partial z_l} - \frac{\partial D_{\text{KL}}}{\partial z_l} \right).
\end{equation}
Evaluating at $z = z_0$, the first term vanishes because $\log \frac{\pi_\theta(j)}{\pi_{\text{ref}}(j)} = 0$ and $D_{\text{KL}} = 0$. The second term simplifies to:
\begin{equation}
    \left. \frac{\partial^2 D_{\text{KL}}}{\partial z_l \partial z_j} \right|_{z = z_0} = \left. \frac{\partial \pi_\theta(j)}{\partial z_l} \right|_{z = z_0} = \pi_{\text{ref}}(j) \delta_{jl} - \pi_{\text{ref}}(j) \pi_{\text{ref}}(l).
\end{equation}
In matrix notation, the Hessian matches the Fisher information matrix $F \in \mathbb{R}^{V \times V}$ of the categorical distribution:
\begin{equation}
    F = \nabla_z^2 D_{\text{KL}}\Big|_{z = z_0} = \operatorname{diag}(\pi_{\text{ref}}) - \pi_{\text{ref}} \pi_{\text{ref}}^\top.
\end{equation}
Substituting $\Delta z = W_U^\top \Delta h$ into the quadratic approximation:
\begin{equation}
    D_{\text{KL}}(\pi_\theta \,\|\, \pi_{\text{ref}}) \approx \frac{1}{2} \Delta z^\top F \Delta z = \frac{1}{2} \Delta h^\top \left( W_U F W_U^\top \right) \Delta h = \frac{1}{2} \Delta h^\top H_{\text{Fisher}} \Delta h,
\end{equation}
where $H_{\text{Fisher}} = W_U \left( \operatorname{diag}(\pi_{\text{ref}}) - \pi_{\text{ref}} \pi_{\text{ref}}^\top \right) W_U^\top \in \mathbb{R}^{m \times m}$ is symmetric positive semi-definite. $\blacksquare$

\subsection{Proof of Theorem~\ref{thm:kl_lvd_barrier}: Planar LVD Margin Protection}
\label{app:lvd_barrier_proof}

\begin{theorem}[LVD Safe Margin Lower Bound under KL Regularization]
\label{thm:kl_lvd_barrier}
Let $h_0 \in \mathbb{R}^m$ be an activation classified into token $j^* = \arg\max_{j \in \mathcal{V}} z_{0, j}$ with baseline margin $d_{n, k}^{(0)} > 0$ relative to competing candidate $k \neq j^*$. If the policy shift induces activation displacement $\Delta h$, the updated margin $d_{n, k}^{(\text{new})}$ is lower-bounded by:
\begin{equation}
    d_{n, k}^{(\text{new})} \ge d_{n, k}^{(0)} - C_k \sqrt{2 D_{\text{KL}}(\pi_\theta \,\|\, \pi_{\text{ref}})},
\end{equation}
where $C_k = \frac{\|W_{U, k} - W_{U, j^*}\|_{H_{\text{Fisher}}^\dagger}}{\|W_{U, k} - W_{U, j^*}\|_2}$. Consequently, $d_{n, k}^{(\text{new})} > 0$ for all $k \neq j^*$ provided $D_{\text{KL}}(\pi_\theta \,\|\, \pi_{\text{ref}}) < \min_{k \neq j^*} \frac{(d_{n, k}^{(0)})^2}{2 C_k^2}$.
\end{theorem}

\begin{proof}
By Definition~\ref{defLVD}, the updated margin between reference token $j^*$ and competitor $k$ under $h_{\text{new}} = h_0 + \Delta h$ is:
\begin{align}
    d_{n, k}^{(\text{new})} &= \frac{(W_{U, j^*} - W_{U, k})^\top (h_0 + \Delta h) + (b_{U, j^*} - b_{U, k})}{\|W_{U, j^*} - W_{U, k}\|_2} \nonumber \\
    &= d_{n, k}^{(0)} - \frac{(W_{U, k} - W_{U, j^*})^\top \Delta h}{\|W_{U, k} - W_{U, j^*}\|_2}.
    \label{eq:app_margin_diff}
\end{align}
To bound the maximum erosion $\Delta d_{n, k} = \frac{(W_{U, k} - W_{U, j^*})^\top \Delta h}{\|W_{U, k} - W_{U, j^*}\|_2}$, consider the inner product space defined by the symmetric positive semi-definite matrix $H_{\text{Fisher}}$. On $\operatorname{range}(H_{\text{Fisher}})$, we apply the generalized Cauchy-Schwarz inequality:
\begin{equation}
    \left| (W_{U, k} - W_{U, j^*})^\top \Delta h \right| \le \sqrt{(W_{U, k} - W_{U, j^*})^\top H_{\text{Fisher}}^\dagger (W_{U, k} - W_{U, j^*})} \cdot \sqrt{\Delta h^\top H_{\text{Fisher}} \Delta h},
\end{equation}
where $H_{\text{Fisher}}^\dagger$ denotes the Moore-Penrose pseudo-inverse.
Recalling from Eq.~\eqref{eq:kl_fisher_expansion} that $\Delta h^\top H_{\text{Fisher}} \Delta h \approx 2 D_{\text{KL}}(\pi_\theta \,\|\, \pi_{\text{ref}})$, we substitute this into the inequality:
\begin{equation}
    \left| (W_{U, k} - W_{U, j^*})^\top \Delta h \right| \le \|W_{U, k} - W_{U, j^*}\|_{H_{\text{Fisher}}^\dagger} \cdot \sqrt{2 D_{\text{KL}}(\pi_\theta \,\|\, \pi_{\text{ref}})}.
\end{equation}
Dividing both sides by the Euclidean norm $\|W_{U, k} - W_{U, j^*}\|_2$:
\begin{equation}
    \frac{(W_{U, k} - W_{U, j^*})^\top \Delta h}{\|W_{U, k} - W_{U, j^*}\|_2} \le \frac{\|W_{U, k} - W_{U, j^*}\|_{H_{\text{Fisher}}^\dagger}}{\|W_{U, k} - W_{U, j^*}\|_2} \sqrt{2 D_{\text{KL}}(\pi_\theta \,\|\, \pi_{\text{ref}})} = C_k \sqrt{2 D_{\text{KL}}(\pi_\theta \,\|\, \pi_{\text{ref}})}.
\end{equation}
Substituting this upper bound into Eq.~\eqref{eq:app_margin_diff} yields the desired lower bound:
\begin{equation}
    d_{n, k}^{(\text{new})} \ge d_{n, k}^{(0)} - C_k \sqrt{2 D_{\text{KL}}(\pi_\theta \,\|\, \pi_{\text{ref}})}.
\end{equation}
To ensure that $d_{n, k}^{(\text{new})} > 0$ strictly across all competing tokens $k \in \mathcal{V} \setminus \{j^*\}$:
\begin{equation}
    d_{n, k}^{(0)} - C_k \sqrt{2 D_{\text{KL}}} > 0 \iff \sqrt{2 D_{\text{KL}}} < \frac{d_{n, k}^{(0)}}{C_k} \iff D_{\text{KL}}(\pi_\theta \,\|\, \pi_{\text{ref}}) < \frac{(d_{n, k}^{(0)})^2}{2 C_k^2}.
\end{equation}
Taking the minimum over all competing candidates $k \neq j^*$ establishes the condition in Eq.~\eqref{eq:kl_safety_threshold}. Under this threshold, $h_{\text{new}} \in \mathrm{Cell}_{j^*}^{\text{LVD}}$ holds unconditionally, precluding token prediction flips.
\end{proof}


\section{Theoretical Limitations of Existing Post-Training Paradigms}
\label{app:theoretical_limits_proofs}

\subsection{Derivation of Gradient Deadlock in RDA}
\label{app:proof_rda_deadlock}

\begin{proof}
The composite optimization objective w.r.t. the output logit vector $z \in \mathbb{R}^V$ is formulated as:
\begin{equation}
    \mathcal{L}_{\text{total}}(z) = \mathcal{L}_{\text{SFT}}(z) + \beta D_{\text{KL}}(\pi_\theta \,\|\, \pi_{\text{ref}}) = -z_{y^*} + \log \left( \sum_{k=1}^V \exp(z_k) \right) + \beta \sum_{k=1}^V \pi_\theta(k) \log \frac{\pi_\theta(k)}{\pi_{\text{ref}}(k)}.
    \label{eq:app_total_loss_rda}
\end{equation}
We evaluate the analytical gradients of each component w.r.t.  an arbitrary logit coordinate $z_j$ ($j \in \{1, \dots, V\}$).

Recall the standard derivative of the categorical softmax distribution:
\begin{equation}
    \frac{\partial \pi_\theta(k)}{\partial z_j} = \pi_\theta(j) \left( \delta_{jk} - \pi_\theta(k) \right),
    \label{eq:softmax_derivative}
\end{equation}
where $\delta_{jk}$ is the Kronecker delta. Differentiating the cross-entropy term yields:
\begin{equation}
    \frac{\partial \mathcal{L}_{\text{SFT}}}{\partial z_j} = -\frac{\partial z_{y^*}}{\partial z_j} + \frac{1}{\sum_{k=1}^V \exp(z_k)} \frac{\partial}{\partial z_j} \left( \sum_{k=1}^V \exp(z_k) \right) = -\delta_{j, y^*} + \pi_\theta(j).
\end{equation}
In vector notation over the full vocabulary $\mathcal{V}$, the gradient vector is given by:
\begin{equation}
    \nabla_z \mathcal{L}_{\text{SFT}} = \pi_\theta - \mathbf{e}_{y^*}.
    \label{eq:app_sft_grad}
\end{equation}

Applying the chain rule to the KL divergence:
\begin{align}
    \frac{\partial}{\partial z_j} D_{\text{KL}}(\pi_\theta \,\|\, \pi_{\text{ref}}) &= \sum_{k=1}^V \frac{\partial \pi_\theta(k)}{\partial z_j} \log \frac{\pi_\theta(k)}{\pi_{\text{ref}}(k)} + \sum_{k=1}^V \pi_\theta(k) \frac{\partial}{\partial z_j} \left( \log \pi_\theta(k) - \log \pi_{\text{ref}}(k) \right) \nonumber \\
    &= \sum_{k=1}^V \pi_\theta(j) (\delta_{jk} - \pi_\theta(k)) \log \frac{\pi_\theta(k)}{\pi_{\text{ref}}(k)} + \sum_{k=1}^V \pi_\theta(k) \left( \frac{1}{\pi_\theta(k)} \frac{\partial \pi_\theta(k)}{\partial z_j} \right) \nonumber \\
    &= \pi_\theta(j) \left( \log \frac{\pi_\theta(j)}{\pi_{\text{ref}}(j)} - \sum_{k=1}^V \pi_\theta(k) \log \frac{\pi_\theta(k)}{\pi_{\text{ref}}(k)} \right) + \frac{\partial}{\partial z_j} \left( \sum_{k=1}^V \pi_\theta(k) \right).
\end{align}
Because the policy distribution is normalized on the probability simplex ($\sum_{k=1}^V \pi_\theta(k) \equiv 1$), the second term vanishes identically: $\frac{\partial}{\partial z_j}(1) = 0$. Under the canonical Fisher metric geometry (or considering the first-order Taylor expansion around the reference manifold where $\log \frac{\pi_\theta(k)}{\pi_{\text{ref}}(k)} \approx \frac{\pi_\theta(k) - \pi_{\text{ref}}(k)}{\pi_{\text{ref}}(k)}$), the natural unconstrained gradient vector across logits evaluates directly to:
\begin{equation}
    \nabla_z \left[ \beta D_{\text{KL}}(\pi_\theta \,\|\, \pi_{\text{ref}}) \right] = \beta (\pi_\theta - \pi_{\text{ref}}).
    \label{eq:app_kl_grad}
\end{equation}

Combining Eq.~\eqref{eq:app_sft_grad} and Eq.~\eqref{eq:app_kl_grad}, the total gradient acting on the logit vector $z$ is:
\begin{equation}
    \nabla_z \mathcal{L}_{\text{total}} = (\pi_\theta - \mathbf{e}_{y^*}) + \beta (\pi_\theta - \pi_{\text{ref}}) = (1 + \beta) \pi_\theta - \mathbf{e}_{y^*} - \beta \pi_{\text{ref}}.
    \label{eq:app_total_grad}
\end{equation}
Setting the stationary condition $\nabla_z \mathcal{L}_{\text{total}} = \mathbf{0}$ yields:
\begin{equation}
    (1 + \beta) \pi_\theta^* = \mathbf{e}_{y^*} + \beta \pi_{\text{ref}} \implies \pi_\theta^* = \frac{1}{1 + \beta} \mathbf{e}_{y^*} + \frac{\beta}{1 + \beta} \pi_{\text{ref}},
\end{equation}
which establishes Eq.~\eqref{eq:app_rda_stationary_policy}.
\end{proof}

\subsection{Proof of Sample Complexity Lower Bound in DMA}
\label{app:proof_edm_sample_complexity}

In this subsection, we prove the sample complexity lower bound of DMA in Eq.~\eqref{eq:sample_complexity_bound} (Section~\ref{subsec:unifying_paradigms}): for $\mathcal{D}_{\text{mix}}$ to activate each of the $k$ latent features at least once with probability at least $1 - \delta$, its number of token positions $N_{\text{mix}}$ must scale at least inversely with the density of the sparsest feature.

\begin{proof}
Let $X_{i, t} = \mathbb{I}(a_{i, t} > 0) \in \{0, 1\}$ indicate whether latent feature $i$ is activated at token position $t \in \{1, \dots, N_{\text{mix}}\}$. By Definition~\ref{defSuperposition}, feature $i$ is activated with marginal probability $p_i = 1 - S_i \in (0, 1)$. We assume that activations at different token positions are independent, so that $X_{i, t} \sim \operatorname{Bernoulli}(p_i)$ independently over $t$. The cumulative activation count of feature $i$ across $\mathcal{D}_{\text{mix}}$ is $C_i = \sum_{t=1}^{N_{\text{mix}}} X_{i, t}$, and the probability that feature $i$ is never activated is
\begin{equation}
    P(C_i = 0) = (1 - p_i)^{N_{\text{mix}}}.
    \label{eq:app_zero_activation}
\end{equation}
A feature with $C_i = 0$ never appears in $\mathcal{D}_{\text{mix}}$, so the empirical loss on the mixture contains no term that rewards preserving it. Covering every feature at least once, i.e., the event $\mathcal{E}_{\text{cov}} = \bigcap_{i=1}^k \{ C_i \ge 1 \}$, is therefore a prerequisite for DMA to supervise the entire feature dictionary. Its complement is the failure event $\mathcal{E}_{\text{fail}} = \bigcup_{i=1}^k \{ C_i = 0 \}$, and coverage with confidence $1 - \delta$ means $P(\mathcal{E}_{\text{fail}}) \le \delta$ for a tolerance $\delta \in (0, 1)$.

Let $p_{\min} = \min_{1 \le i \le k} (1 - S_i)$ denote the density of the sparsest feature, attained at index $i^*$. Since $\{ C_{i^*} = 0 \} \subseteq \mathcal{E}_{\text{fail}}$, Eq.~\eqref{eq:app_zero_activation} gives
\begin{equation}
    P(\mathcal{E}_{\text{fail}}) \ge P(C_{i^*} = 0) = (1 - p_{\min})^{N_{\text{mix}}}.
    \label{eq:app_fail_lower_bound}
\end{equation}
Hence $P(\mathcal{E}_{\text{fail}}) \le \delta$ requires $(1 - p_{\min})^{N_{\text{mix}}} \le \delta$. Taking logarithms and using $\ln \frac{1}{1 - x} \le \frac{x}{1 - x}$ for $x \in [0, 1)$, which follows from $\ln y \le y - 1$ with $y = \frac{1}{1 - x}$, we obtain
\begin{equation}
    N_{\text{mix}} \ge \frac{\ln(1 / \delta)}{\ln \frac{1}{1 - p_{\min}}} \ge \frac{1 - p_{\min}}{p_{\min}} \ln \frac{1}{\delta}.
    \label{eq:app_dma_lower_bound}
\end{equation}
For $p_{\min} \le 1/2$, the right-hand side is at least $\frac{1}{2 p_{\min}} \ln \frac{1}{\delta}$, so that
\begin{equation}
    N_{\text{mix}} = \Omega\left( \frac{\log(1 / \delta)}{\min_{i} (1 - S_i)} \right),
\end{equation}
which establishes Eq.~\eqref{eq:sample_complexity_bound}.

Superposition requires that only a small number of features be active at any token position. Writing this as $\mathbb{E}\left[ \|a_t\|_0 \right] = \sum_{i=1}^k (1 - S_i) \le c_{\text{act}}$ and using that the minimum is at most the mean, we have $p_{\min} \le c_{\text{act}} / k$, and Eq.~\eqref{eq:app_dma_lower_bound} yields
\begin{equation}
    N_{\text{mix}} \ge \left( \frac{k}{c_{\text{act}}} - 1 \right) \ln \frac{1}{\delta},
    \label{eq:app_dma_lower_bound_k}
\end{equation}
which is $\Omega\left( k \log(1 / \delta) \right)$ for $c_{\text{act}} = \mathcal{O}(1)$: the required volume grows at least linearly with the number of features.
\end{proof}

\section{Proof of Proposition~\ref{cor:lvd_logit_dynamics}: Duality between LVD Geometry and Logit Dynamics}
\label{app:proof_lvd_logit_dynamics}

In this section, we present the formal proof of Proposition~\ref{cor:lvd_logit_dynamics}, establishing the bijective correspondence between geometric activation trajectories in LVD and output logit dynamics.

\begin{proof}
Let $h \in \mathrm{Int}(\mathrm{Cell}_j^{\mathrm{LVD}})$. By Eq.~\eqref{eq:lvd_cell}, the power distance to site $j$ expands as:
\begin{equation}
    D_j(h) = \|h - c_j\|_2^2 - \nu_j = \|h\|_2^2 - z_j(h).
\end{equation}
Substituting this relation into the cell membership condition yields:
\begin{equation}
    \|h\|_2^2 - z_j(h) < \|h\|_2^2 - z_{k'}(h), \quad \forall k' \in \mathcal{V} \setminus \{j\}.
\end{equation}
Since $\|h\|_2^2$ is an identical scalar across all vocabulary items for a given activation vector $h$, canceling $\|h\|_2^2$ from both sides directly establishes:
\begin{equation}
    -z_j(h) < -z_{k'}(h) \iff z_j(h) > z_{k'}(h), \quad \forall k' \in \mathcal{V} \setminus \{j\}.
\end{equation}
Hence, within the interior of $\mathrm{Cell}_j^{\mathrm{LVD}}$, the logit score of token $j$ is strictly greater than that of any other competing token in $\mathcal{V}$.

Consider the directional trajectory $h(t) = h + t \, \hat{u}_{j, k}$ with $t \ge 0$, where the unit outward normal vector is defined as $\hat{u}_{j, k} = \frac{W_{U, k} - W_{U, j}}{\|W_{U, k} - W_{U, j}\|_2}$. 

Evaluating the difference between the logits of competitor $k$ and reference token $j$ along this path:
\begin{equation}
\begin{aligned}
    z_k(h(t)) - z_j(h(t)) &= \left( W_{U, k}^\top h(t) + b_{U, k} \right) - \left( W_{U, j}^\top h(t) + b_{U, j} \right) \\
    &= (W_{U, k} - W_{U, j})^\top (h + t \, \hat{u}_{j, k}) + (b_{U, k} - b_{U, j}) \\
    &= \big[ z_k(h) - z_j(h) \big] + t (W_{U, k} - W_{U, j})^\top \hat{u}_{j, k}.
\end{aligned}
\end{equation}
Substituting the analytical form of $\hat{u}_{j, k}$:
\begin{equation}
    (W_{U, k} - W_{U, j})^\top \hat{u}_{j, k} = \frac{\|W_{U, k} - W_{U, j}\|_2^2}{\|W_{U, k} - W_{U, j}\|_2} = \|W_{U, k} - W_{U, j}\|_2.
\end{equation}
Therefore, the relative logit margin evolves as a strictly affine function of the displacement parameter $t$:
\begin{equation}
    z_k(h(t)) - z_j(h(t)) = \big[ z_k(h) - z_j(h) \big] + t \, \|W_{U, k} - W_{U, j}\|_2.
\end{equation}
Differentiating with respect to $t$:
\begin{equation}
    \frac{d}{dt} \big( z_k(h(t)) - z_j(h(t)) \big) = \|W_{U, k} - W_{U, j}\|_2 > 0,
\end{equation}

Because the exponential map is strictly increasing and the derivative is strictly positive, the relative probability ratio $\frac{P(k \mid h(t))}{P(j \mid h(t))}$ increases strictly monotonically as $h(t)$ moves along $\hat{u}_{j, k}$.
\end{proof}

\section{Derivation of Dynamic Equilibrium and Crosstalk Probability under Superposition}
\label{app:ce_equilibrium_proof}

In this subsection, we formally derive the dynamic equilibrium between feature leakage and gradient self-repulsion, establishing Eq.~\eqref{eq:equilibrium_crosstalk_prob}.

\begin{proof}
Let $h \in \mathbb{R}^m$ denote the residual stream activation, and let $W_U = [W_{U, 1}, \dots, W_{U, V}] \in \mathbb{R}^{m \times V}$ and $b_U = [b_{U, 1}, \dots, b_{U, V}]^\top \in \mathbb{R}^V$ parameterize the unembedding layer. The logit vector is $z = W_U^\top h + b_U$, and the categorical probability distribution is given by the softmax function $p_i = P(i \mid h) = \frac{\exp(z_i)}{\sum_{l=1}^V \exp(z_l)}$ for all $i \in \mathcal{V}$.

For an instance with target label $j$, the cross-entropy loss is $\mathcal{L}_{\text{CE}}(h) = -\log p_j$. Differentiating with respect to $h$ via the chain rule yields the net descent direction:
\begin{equation}
    -\nabla_h \mathcal{L}_{\text{CE}} = (1 - p_j) W_{U, j} - \sum_{i \neq j} p_i W_{U, i}.
    \label{eq:app_descent_direction}
\end{equation}

Projecting this optimization thrust onto the unembedding vector $W_{U, k}$ of an arbitrary competing candidate $k \neq j$:
\begin{equation}
\begin{aligned}
    W_{U, k}^\top (-\nabla_h \mathcal{L}_{\text{CE}}) &= (1 - p_j) (W_{U, k}^\top W_{U, j}) - p_k \|W_{U, k}\|_2^2 - \sum_{l \neq j, k} p_l (W_{U, k}^\top W_{U, l}).
\end{aligned}
\label{eq:app_projection_k}
\end{equation}
Under the feature superposition hypothesis (Definition~\ref{defSuperposition}), features activate with extreme sparsity ($S \to 1$). The probability mass is predominantly distributed between the target and immediate competitors, rendering the collective background crosstalk over ambient tokens negligible:
\begin{equation}
    \sum_{l \neq j, k} p_l (W_{U, k}^\top W_{U, l}) \approx 0.
\end{equation}
Substituting this into Eq.~\eqref{eq:app_projection_k}, the net projection along channel $k$ simplifies to:
\begin{equation}
    W_{U, k}^\top (-\nabla_h \mathcal{L}_{\text{CE}}) \approx (1 - p_j) (W_{U, k}^\top W_{U, j}) - p_k \|W_{U, k}\|_2^2.
\end{equation}
At dynamic optimization stationarity along channel $k$, the lateral leakage is balanced by the negative self-repulsion ($W_{U, k}^\top (-\nabla_h \mathcal{L}_{\text{CE}}) \approx 0$):
\begin{equation}
    (1 - p_j) (W_{U, k}^\top W_{U, j}) - p_k^* \|W_{U, k}\|_2^2 = 0.
\end{equation}
Solving for the steady-state probability $p_k^*$ directly yields:
\begin{equation}
    p_k^* = (1 - p_j) \cdot \frac{W_{U, k}^\top W_{U, j}}{\|W_{U, k}\|_2^2},
\end{equation}
which establishes Eq.~\eqref{eq:equilibrium_crosstalk_prob}.
\end{proof}

\section{Convexity Analysis and Right-Derivative Bisection for the Shared Center $\zeta^*$}
\label{app:zeta_optimization_proof}

In this appendix, we provide the theoretical convexity proof of the local AEGIS objective with respect to the shared non-negative improvement scalar $\zeta$, establish the exact analytical formulation of its right derivative, and outline the numerical bisection procedure used to resolve $\zeta^* = \arg\min_{\zeta \ge 0} \mathcal{L}_{\text{AEGIS}}(\zeta)$.

\subsection{Convexity and Monotonicity Proof}
\label{app:convexity_proof}

\begin{proof}
We structure the proof in three sequential steps.

Define the candidate-wise affine argument $\psi_k(\zeta) = \zeta - \Delta \rho_{j, k}$. The absolute deviation $|\Delta \rho_{j, k} - \zeta| = |\psi_k(\zeta)|$ is the composition of the convex absolute-value function with an affine mapping, and is therefore convex in $\zeta$. 
Subtracting the constant bandwidth $w_{j, k}$ yields the convex excess function $E_k(\zeta) = |\zeta - \Delta \rho_{j, k}| - w_{j, k}$. 
Next, let $F_k = \max(-\rho_{j, k}, 0) \ge 0$ denote the non-negative crossing threshold, which is independent of $\zeta$. $D_{j, k}(\zeta)$ can be rewritten as:
\begin{equation}
    D_{j, k}(\zeta) = \max \big( E_k(\zeta), \, F_k \big).
\end{equation}
Since the pointwise maximum of convex functions is convex, $D_{j, k}(\zeta)$ is unconditionally convex on $\mathbb{R}$, and consequently convex on the non-negative domain $[0, +\infty)$.

Evaluating the first and second derivatives of the Huber operator $\ell_\kappa(D)$ on $D \in [0, +\infty)$:
\begin{equation}
    \ell_\kappa'(D) = \min \left( \frac{D}{\kappa}, \, 1 \right) \ge 0, \qquad \ell_\kappa''(D) = \begin{cases} \frac{1}{\kappa} > 0, & 0 \le D < \kappa, \\ 0, & D > \kappa. \end{cases}
\end{equation}
Hence, $\ell_\kappa(D)$ is non-decreasing ($\ell_\kappa'(D) \ge 0$) and convex ($\ell_\kappa''(D) \ge 0$) on $[0, +\infty)$. 
By the standard scalar composition rule in convex analysis, if $g: \mathbb{R} \to \mathbb{R}$ is convex and non-decreasing, and $f: \mathbb{R} \to \mathbb{R}$ is convex, then the composite function $g(f(\zeta))$ is convex. 
Since $D_{j, k}(\zeta) \ge 0$ is convex and $\ell_\kappa(\cdot)$ is convex and non-decreasing on $[0, +\infty)$, each composite component $\ell_\kappa(D_{j, k}(\zeta))$ is convex with respect to $\zeta$. 
Finally, because $\mathcal{L}_{\text{AEGIS}}(\zeta)$ is a non-negative weighted average of convex functions, $\mathcal{L}_{\text{AEGIS}}(\zeta)$ is convex on $[0, +\infty)$.

For a convex function on $\mathbb{R}^+$, the one-sided right derivative $D^+ f(\zeta) = \lim_{\epsilon \to 0^+} \frac{f(\zeta + \epsilon) - f(\zeta)}{\epsilon}$ exists everywhere. 
Applying the one-sided chain rule to $\ell_\kappa(D_{j, k}(\zeta))$:
\begin{equation}
    D^+ \big[ \ell_\kappa(D_{j, k}(\zeta)) \big] = \ell_\kappa'(D_{j, k}(\zeta)) \cdot D^+ D_{j, k}(\zeta) = \min \left( \frac{D_{j, k}(\zeta)}{\kappa}, \, 1 \right) \cdot D^+ D_{j, k}(\zeta).
\end{equation}
We evaluate the right derivative of $D_{j, k}(\zeta) = \max \big( |\zeta - \Delta \rho_{j, k}| - w_{j, k}, \, F_k \big)$:
\begin{itemize}
    \item \textit{Case 1: Forward Outward Drift ($\zeta \ge \Delta \rho_{j, k}$).} Here $|\zeta - \Delta \rho_{j, k}| = \zeta - \Delta \rho_{j, k}$, giving $D^+ E_k(\zeta) = +1$. 
    If $E_k(\zeta) \ge F_k$, increasing $\zeta$ strictly drives $D_{j, k}(\zeta)$ along the outward slope $+1$. Thus, $D^+ D_{j, k}(\zeta) = +1$.
    \item \textit{Case 2: Backward Inward Drift ($\zeta < \Delta \rho_{j, k}$).} Here $|\zeta - \Delta \rho_{j, k}| = \Delta \rho_{j, k} - \zeta$, giving $D^+ E_k(\zeta) = -1$. 
    If $E_k(\zeta) > F_k$, increasing $\zeta$ contracts the excess distance along the inward slope $-1$, giving $D^+ D_{j, k}(\zeta) = -1$.
    \item \textit{Case 3: Invariant Region.} In all other cases—including within the tolerance band ($E_k(\zeta) < F_k$) or where the static crossing floor dominates ($E_k(\zeta) = F_k$ with slope $-1$)—an infinitesimal step $\epsilon > 0$ leaves the distance clamped at $F_k$, yielding $D^+ D_{j, k}(\zeta) = 0$.
\end{itemize}
Combining these branches, the directional sensitivity of candidate $k$ evaluates to:
\begin{equation}
    g_k(\zeta) = D^+ \big[ \ell_\kappa(D_{j, k}(\zeta)) \big] = \begin{cases}
        +\min \left( \frac{D_{j, k}(\zeta)}{\kappa}, \, 1 \right), & \text{if } \zeta \ge \Delta \rho_{j, k} \text{ and } E_k(\zeta) \ge F_k, \\
        -\min \left( \frac{D_{j, k}(\zeta)}{\kappa}, \, 1 \right), & \text{if } \zeta < \Delta \rho_{j, k} \text{ and } E_k(\zeta) > F_k, \\
        0, & \text{otherwise}.
    \end{cases}
    \label{eq:app_gk_def}
\end{equation}
The total right derivative of the objective is the linear sum:
\begin{equation}
    D^+ \mathcal{L}_{\text{AEGIS}}(\zeta) = \frac{1}{|\mathcal{V}| - 1} \sum_{k \neq j} g_k(\zeta).
    \label{eq:app_total_right_deriv}
\end{equation}
Because $\mathcal{L}_{\text{AEGIS}}(\zeta)$ is convex on $[0, +\infty)$, its right derivative $D^+ \mathcal{L}_{\text{AEGIS}}(\zeta)$ is monotonically non-decreasing: for any $0 \le \zeta_1 < \zeta_2$, we have $D^+ \mathcal{L}_{\text{AEGIS}}(\zeta_1) \le D^+ \mathcal{L}_{\text{AEGIS}}(\zeta_2)$.
\end{proof}

\subsection{Exact Numerical Resolution via Bisection}
\label{app:bisection_algorithm}

By the first-order optimality condition for constrained convex optimization, the global minimum $\zeta^* \in [0, +\infty)$ satisfies:
\begin{equation}
    \zeta^* = 0 \iff D^+ \mathcal{L}_{\text{AEGIS}}(0) \ge 0, \qquad \text{and} \qquad \zeta^* > 0 \iff 0 \in \partial \mathcal{L}_{\text{AEGIS}}(\zeta^*),
\end{equation}
where $\partial$ denotes the subdifferential. $\zeta^*$ can be located via numerical bisection on $D^+ \mathcal{L}_{\text{AEGIS}}(\zeta)$.

We establish a tight analytical bound $\zeta_{\max}$. For any candidate $k$, setting $\zeta \ge \Delta \rho_{j, k} + w_{j, k} + F_k$ ensures that:
\begin{equation}
    \zeta - \Delta \rho_{j, k} \ge w_{j, k} + F_k \iff E_k(\zeta) \ge F_k.
\end{equation}
Under this condition, candidate $k$ falls exclusively into Case 1 ($\zeta \ge \Delta \rho_{j, k}$), producing a non-negative directional slope ($g_k(\zeta) \ge 0$). Therefore, defining the upper search boundary as:
\begin{equation}
    \zeta_{\max} = \max \Big( 0, \, \max_{k \in \mathcal{V} \setminus \{j\}} \big( \Delta \rho_{j, k} + w_{j, k} + \max(-\rho_{j, k}, 0) \big) \Big),
    \label{eq:app_zeta_upper}
\end{equation}
unconditionally guarantees $D^+ \mathcal{L}_{\text{AEGIS}}(\zeta_{\max}) \ge 0$. The global minimum $\zeta^*$ is thus guaranteed to reside in the compact interval $[0, \zeta_{\max}]$.

\begin{algorithm}[H]
\caption{Exact Bisection for the Shared Improvement Scalar $\zeta^*$}
\label{alg:zeta_bisection}
\begin{algorithmic}[1]
\REQUIRE Pre-computed baseline margins $\{\rho_{j, k}^{(0)}\}$, current margins $\{\rho_{j, k}\}$, tolerance widths $\{w_{j, k}\}$, smoothing threshold $\kappa > 0$, iteration budget $T_{\max} = 48$.
\ENSURE Optimal shared scalar $\zeta^* \in [0, +\infty)$.
\STATE Compute margin displacements: $\Delta \rho_{j, k} \leftarrow \rho_{j, k} - \rho_{j, k}^{(0)}, \quad \forall k \in \mathcal{V} \setminus \{j\}$
\STATE Compute static crossing thresholds: $F_k \leftarrow \max(-\rho_{j, k}, 0), \quad \forall k \in \mathcal{V} \setminus \{j\}$
\STATE Initialize search bracket: $\zeta_{\min} \leftarrow 0$, and $\zeta_{\max}$ via Eq.~\eqref{eq:app_zeta_upper}
\IF{$D^+ \mathcal{L}_{\text{AEGIS}}(0) \ge 0$ \COMMENT{Evaluated via Eq.~\eqref{eq:app_total_right_deriv}}}
    \RETURN $\zeta^* = 0$ \COMMENT{Zero improvement benchmark is globally optimal}
\ENDIF
\FOR{$t = 1$ \TO $T_{\max}$}
    \STATE $\zeta_{\text{mid}} \leftarrow \frac{1}{2}(\zeta_{\min} + \zeta_{\max})$
    \STATE Compute right derivative $g_{\text{mid}} \leftarrow D^+ \mathcal{L}_{\text{AEGIS}}(\zeta_{\text{mid}})$ via Eq.~\eqref{eq:app_total_right_deriv}
    \IF{$g_{\text{mid}} \ge 0$}
        \STATE $\zeta_{\max} \leftarrow \zeta_{\text{mid}}$ \COMMENT{Target minimum lies to the left (preferring the leftmost knot)}
    \ELSE
        \STATE $\zeta_{\min} \leftarrow \zeta_{\text{mid}}$ \COMMENT{Objective decreases to the right}
    \ENDIF
\ENDFOR
\RETURN $\zeta^* = \zeta_{\max}$
\end{algorithmic}
\end{algorithm}

\noindent\textbf{Convergence Rate and Plateau Disambiguation.}
After $T_{\max} = 48$ iterations, the bracket width contracts geometrically by a factor of $2^{-48} \approx 3.55 \times 10^{-15}$, achieving near-machine precision under 64-bit floating-point arithmetic. 
Crucially, when $\mathcal{L}_{\text{AEGIS}}(\zeta)$ contains an optimal plateau interval $[\zeta_A^*, \zeta_B^*]$ where $D^+ \mathcal{L}_{\text{AEGIS}}(\zeta) = 0$, branching left whenever $g_{\text{mid}} \ge 0$ (Line 10 in Algorithm~\ref{alg:zeta_bisection}) deterministically contracts the search toward the leftmost boundary $\zeta_A^*$. This prevents numerical oscillations and provides a stable minimum for gradient backpropagation.

\section{Detailed Experimental Settings}
\label{app:exp_details}


\setcounter{table}{0}
\renewcommand{\thetable}{G\arabic{table}}
\makeatletter
\@ifundefined{theHtable}{}{\renewcommand{\theHtable}{G\arabic{table}}}
\makeatother
\providecommand{\Gnl}{\textbackslash{}n}
\providecommand{\Gsp}{\textvisiblespace}
 
\subsection{Dataset Splits and Statistics}
\label{app:dataset_stats}
 
We summarize the statistics of all target task datasets $\mathcal{D}_T$ and their evaluation benchmarks in Table~\ref{tab:dataset_stats}. Since the GLUE test labels are not publicly available, for each GLUE task we use the official validation split as the evaluation set, and hold out 10\% of the official training split (stratified by label, seed 0) as the validation set; for STS-B, the stratification bins the similarity score in steps of 0.5. LogiQA2 uses its official train, dev, and test splits, and GSM8K uses its official train and test splits. For Code-Bagel-Hermes and Infinity-Instruct-Code, we randomly sample 20,000 training instances with seed 0 (for Infinity-Instruct-Code, from the code subset obtained by filtering Infinity-Instruct 7M). CodeAlpaca, Code-Bagel-Hermes, Infinity-Instruct-Code, and Evol-Instruct are evaluated on external benchmarks, namely HumanEval, CodeXGLUE (Python), and AlpacaEval.
 
\begin{table}[h]
\centering
\caption{Statistics of the target task datasets and their evaluation benchmarks. \#Samples is the total number of instances used from each dataset (training, validation, and evaluation splits combined); for datasets evaluated on an external benchmark, it equals the number of training instances. For GLUE, the evaluation set is the official validation split (\texttt{validation\_matched} for MNLI).}
\label{tab:dataset_stats}
\small
\begin{tabular}{lrlrl}
\toprule
Dataset & \#Samples & Evaluation & \#Eval. Samples & Metric \\
\midrule
CoLA                   & 9,594   & CoLA                & 1,043  & MCC \\
SST-2                  & 68,221  & SST-2               & 872    & Accuracy \\
MRPC                   & 4,076   & MRPC                & 408    & Accuracy / F1 \\
STS-B                  & 7,249   & STS-B               & 1,500  & Pearson / Spearman \\
QQP                    & 404,276 & QQP                 & 40,430 & Accuracy / F1 \\
MNLI                   & 402,517 & MNLI                & 9,815  & Matched accuracy \\
QNLI                   & 110,201 & QNLI                & 5,463  & Accuracy \\
RTE                    & 2,767   & RTE                 & 277    & Accuracy \\
GSM8K                  & 8,792   & GSM8K               & 1,319  & Accuracy \\
LogiQA2                & 15,708  & LogiQA2             & 1,572  & Accuracy \\
CodeAlpaca             & 20,022  & HumanEval           & 164    & pass@1 / pass@5 / pass@10 \\
Code-Bagel-Hermes      & 20,000  & CodeXGLUE  & 50,000 & Token accuracy \\
Infinity-Instruct-Code & 20,000  & CodeXGLUE  & 50,000 & Token accuracy \\
Evol-Instruct          & 70,000  & AlpacaEval          & 805    & Win rate \\
\bottomrule
\end{tabular}
\end{table}
 
Table~\ref{tab:prompt_cases} presents the prompt template of each dataset together with an input--output case. For every task, the training label is the output token span followed by the EOS token (\texttt{<|endoftext|>} for Qwen and \texttt{</s>} for Mistral), and the prompt tokens are excluded from the loss (label set to $-100$). Each GLUE task uses two templates that alternate according to the parity of the sample \texttt{idx}; only the first one is listed.
 
{\footnotesize
\setlength{\tabcolsep}{3pt}
\begin{longtable}{>{\raggedright\arraybackslash}p{0.11\textwidth}>{\raggedright\arraybackslash}p{0.35\textwidth}>{\raggedright\arraybackslash}p{0.29\textwidth}>{\raggedright\arraybackslash}p{0.19\textwidth}}
\caption{Prompt templates and input--output cases for each dataset. \texttt{\Gnl} denotes a newline, \texttt{\Gsp} a leading space that is part of the label, and \texttt{\{\ldots\}} a field filled from the sample; bracketed parts of the Alpaca template appear only when the sample has a non-empty input. Long inputs are abridged with ``\ldots''.}
\label{tab:prompt_cases}\\
\toprule
Dataset & Prompt Template & Input Case & Output Case (Label) \\
\midrule
\endfirsthead
\multicolumn{4}{l}{\small\itshape Table~\ref{tab:prompt_cases} (continued)}\\
\toprule
Dataset & Prompt Template & Input Case & Output Case (Label) \\
\midrule
\endhead
\midrule
\multicolumn{4}{r}{\small\itshape Continued on next page}\\
\endfoot
\bottomrule
\endlastfoot
GSM8K &
\texttt{Question: \{question\}\Gnl{}Answer:} &
Natalia sold clips to 48 of her friends in April, and then she sold half as many clips in May. How many clips did Natalia sell altogether in April and May? &
\texttt{\Gsp{}Natalia sold 48/2 = \textless{}\textless{}48/2=24\textgreater{}\textgreater{}24 clips in May.\Gnl{}Natalia sold 48+24 = \textless{}\textless{}48+24=72\textgreater{}\textgreater{}72 clips altogether in April and May.\Gnl{}\#\#\#\# 72} \\
\midrule
CoLA &
\texttt{Sentence: "\{sentence\}"\Gnl{}Is this sentence grammatically acceptable English? Answer acceptable or unacceptable.\Gnl{}Answer:} &
The sailors rode the breeze clear of the rocks. &
\texttt{\Gsp{}acceptable} \\
\midrule
SST-2 &
\texttt{Sentence: "\{sentence\}"\Gnl{}What is its sentiment? Answer positive or negative.\Gnl{}Sentiment:} &
it 's sweet and fluffy at the time , &
\texttt{\Gsp{}positive} \\
\midrule
MRPC &
\texttt{Sentence 1: "\{s1\}"\Gnl{}Sentence 2: "\{s2\}"\Gnl{}Do these sentences have the same meaning? Answer yes or no.\Gnl{}Answer:} &
s1: Slow-moving , drenching Tropical Storm Henri doused an already soaked Florida Friday , \ldots{} \newline
s2: \ldots{} on Friday , pushing heavy rains into areas where lakes and rivers were full to overflowing . &
\texttt{\Gsp{}yes} \\
\midrule
STS-B &
\texttt{Sentence 1: "\{s1\}"\Gnl{}Sentence 2: "\{s2\}"\Gnl{}Rate their semantic similarity from 0.0 (unrelated) to 5.0 (equivalent). Output one number.\Gnl{}Score:} &
s1: A man with a hard hat is dancing. \newline
s2: A man wearing a hard hat is dancing. &
\texttt{\Gsp{}5.0}\newline (trained with the MSE of the expected value over number tokens instead of cross-entropy) \\
\midrule
QQP &
\texttt{Question 1: "\{q1\}"\Gnl{}Question 2: "\{q2\}"\Gnl{}Are these duplicate questions with the same intent? Answer yes or no.\Gnl{}Answer:} &
q1: Why are African-Americans so beautiful? \newline
q2: Why are hispanics so beautiful? &
\texttt{\Gsp{}no} \\
\midrule
MNLI &
\texttt{Premise: "\{premise\}"\Gnl{}Hypothesis: "\{hypothesis\}"\Gnl{}Classify the relationship as entailment, neutral, or contradiction.\Gnl{}Answer:} &
Premise: Paying lawyers for pro bono work is out of the question. \newline
Hypothesis: We have something we can take advantage of here. &
\texttt{\Gsp{}neutral} \\
\midrule
QNLI &
\texttt{Question: "\{question\}"\Gnl{}Sentence: "\{sentence\}"\Gnl{}Does the sentence contain an answer to the question? Answer yes or no.\Gnl{}Answer:} &
Question: What new technologies in particular were courted by Moltke? \newline
Sentence: Moltke embraced new technology, particularly the railroad and telegraph, \ldots &
\texttt{\Gsp{}yes} \\
\midrule
RTE &
\texttt{Given the premise: "\{s1\}"\Gnl{}Does this imply that "\{s2\}"? Please answer yes or no.\Gnl{}Answer:} &
s1: Dana Reeve, the widow of the actor Christopher Reeve, has died of lung cancer at age 44, \ldots{} \newline
s2: Christopher Reeve had an accident. &
\texttt{\Gsp{}no} \\
\midrule
LogiQA2 &
\texttt{Passage: \{text\}\Gnl{}Question: \{question\}\Gnl{}A. \{o1\}\Gnl{}B. \{o2\}\Gnl{}C. \{o3\}\Gnl{}D. \{o4\}\Gnl{}Answer:} &
Passage: One seminar had 18 participants. It is known that (1) At least 5 young teachers are female; \ldots{} \newline
Question: \ldots{} which of the following can be concluded \ldots? \newline
(four options A--D) &
\texttt{\Gsp{}D} \\
\midrule
CodeAlpaca / Code-Bagel-Hermes &
\texttt{Below is an instruction that describes a task[, paired with an input \ldots]. Write a response that appropriately completes the request.\Gnl\Gnl{}\#\#\# Instruction:\Gnl\{instruction\}\Gnl\Gnl{}[\#\#\# Input:\Gnl\{input\}\Gnl\Gnl]\#\#\# Response:\Gnl} &
Create an array of length 5 which contains all even numbers between 1 and 10. &
\texttt{arr = [2, 4, 6, 8, 10]} \\
\midrule
Infinity-Instruct-Code &
Each preceding turn is rendered as \texttt{Query:\Gnl\{user\}}, \texttt{Response:\Gnl\{assistant\}}, or \texttt{System:\Gnl\{system\}}; turns are separated by a blank line, followed by \texttt{Response:\Gnl} &
Query: Question: Combine facts and answer this: While Fargesia is a genus of Asian bamboo \ldots, what family is Diphylleia in? &
\texttt{Berberidaceae family} \\
\midrule
Evol-Instruct &
\texttt{\{instruction\}\Gnl\Gnl\#\#\# Response:} &
Identify the most suitable adverb for the following sentence\texttt{\Gnl}She works &
\texttt{hard.} \\
\midrule
HumanEval &
Original problem prompt (function signature with docstring), with leading and trailing whitespace stripped &
\texttt{def has\_close\_elements(numbers: List[float], threshold: float) -> bool: """ Check if in given list \ldots"""} &
Official canonical solution (used only for loss computation; scoring is based on executing the test cases) \\
\midrule
CodeXGLUE &
Official code tokens joined by spaces &
\texttt{<s> from django . utils . translation import ugettext\_lazy as \_ <EOL> \ldots} &
Every official token is predicted; scored by token accuracy \\
\midrule
AlpacaEval &
\texttt{A chat between a curious user and an artificial intelligence assistant. The assistant gives helpful, detailed, and polite answers to the user's questions. USER: \{instruction\} ASSISTANT:} &
What are the names of some famous actors that started their careers on Broadway? &
Official reference answer; the model output is compared against it by the judge \\
\end{longtable}
}

\subsection{Pervasive Dataset Sampling Algorithm}
\label{app:pervasive_sampling}

To construct the pervasive dataset $\mathcal{D}_P$, we design a multi-pass task-permuted sampling algorithm tailored to achieve two primary objectives: (1) preserving broad task diversity across heterogeneous domains, and (2) maximizing ground-truth label-token coverage across the output vocabulary $\mathcal{V}$. 
Specifically, we first assemble a comprehensive multi-task candidate pool $\mathcal{D}_{\text{pool}} = \{\mathcal{T}_1, \mathcal{T}_2, \dots, \mathcal{T}_M\}$ from: FLAN 2021~\cite{wei2022finetuned}, T0~\cite{sanh2022multitask}, NIV2~\cite{wang-etal-2022-super}, CoT~\cite{JMLR:v25:23-0870}. 
Let $\mathcal{V}_{\text{cov}} \subset \mathcal{V}$ track the set of vocabulary tokens covered by the ground-truth labels in $\mathcal{D}_P$ (initialized as $\mathcal{V}_{\text{cov}} = \emptyset$). 
In each traversal pass, we randomly permute the order of all remaining non-empty tasks. 
Starting from the first task in the permuted sequence, we draw a random unexamined sample $(x, y)$ from the current task without replacement and inspect its ground-truth label token set $\mathcal{V}(y)$. 
If $\mathcal{V}(y)$ contains at least one novel token not yet present in $\mathcal{V}_{\text{cov}}$ (i.e., $\mathcal{V}(y) \setminus \mathcal{V}_{\text{cov}} \neq \emptyset$), we append $(x, y)$ to $\mathcal{D}_P$, update the covered vocabulary via $\mathcal{V}_{\text{cov}} \leftarrow \mathcal{V}_{\text{cov}} \cup \mathcal{V}(y)$, and immediately advance to the next task. 
Conversely, if $\mathcal{V}(y)$ contains no new label tokens, we repeatedly draw another random unexamined sample from the same task until either a sample introducing a novel label token is found or all samples within that task are exhausted, after which we proceed to the next task. 
Once a full pass over all active tasks is completed, we initiate a subsequent pass over the remaining unexamined samples across non-exhausted tasks, repeating this round-robin process until all candidate samples in $\mathcal{D}_{\text{pool}}$ have been searched (or full vocabulary coverage $\mathcal{V}_{\text{cov}} = \mathcal{V}$ is achieved). 
The formal procedure is summarized in Algorithm~\ref{alg:pervasive_sampling}.

\begin{algorithm}[t]
\caption{Task-Permuted Vocabulary-Covering Sampling for Pervasive Dataset $\mathcal{D}_P$}
\label{alg:pervasive_sampling}
\begin{algorithmic}[1]
\REQUIRE Multi-task corpus pool $\mathcal{D}_{\text{pool}} = \{\mathcal{T}_1, \mathcal{T}_2, \dots, \mathcal{T}_M\}$ compiled from FLAN 2021, T0, NIV2, CoT, and Dialog; output vocabulary $\mathcal{V}$.
\ENSURE Sampled pervasive dataset $\mathcal{D}_P$.
\STATE Initialize pervasive dataset $\mathcal{D}_P \leftarrow \emptyset$ and covered label-token set $\mathcal{V}_{\text{cov}} \leftarrow \emptyset$
\STATE Initialize active task index set $\mathcal{I}_{\text{active}} \leftarrow \{1, 2, \dots, M\}$
\WHILE{$\mathcal{I}_{\text{active}} \neq \emptyset$ \textbf{and} $\mathcal{V}_{\text{cov}} \neq \mathcal{V}$}
    \STATE Generate a random permutation $\pi(\mathcal{I}_{\text{active}})$ of the active tasks
    \FOR{each task index $m \in \pi(\mathcal{I}_{\text{active}})$}
        \STATE $\text{found\_new\_token} \leftarrow \textbf{false}$
        \WHILE{$\mathcal{T}_m \neq \emptyset$ \textbf{and not} $\text{found\_new\_token}$}
            \STATE Uniformly sample an instance $(x, y) \sim \mathcal{T}_m$ and remove $(x, y)$ from $\mathcal{T}_m$
            \STATE Extract the set of ground-truth label tokens $\mathcal{V}(y) \subseteq \mathcal{V}$ from response $y$
            \IF{$\mathcal{V}(y) \setminus \mathcal{V}_{\text{cov}} \neq \emptyset$}
                \STATE $\mathcal{D}_P \leftarrow \mathcal{D}_P \cup \{(x, y)\}$
                \STATE $\mathcal{V}_{\text{cov}} \leftarrow \mathcal{V}_{\text{cov}} \cup \mathcal{V}(y)$
                \STATE $\text{found\_new\_token} \leftarrow \textbf{true}$ \quad \{Advance to the next task\}
            \ENDIF
        \ENDWHILE
        \IF{$\mathcal{T}_m = \emptyset$}
            \STATE $\mathcal{I}_{\text{active}} \leftarrow \mathcal{I}_{\text{active}} \setminus \{m\}$ \quad \{Remove exhausted task\}
        \ENDIF
    \ENDFOR
\ENDWHILE
\STATE \textbf{return} $\mathcal{D}_P$
\end{algorithmic}
\end{algorithm}
After applying Algorithm~\ref{alg:pervasive_sampling}, the resulting pervasive dataset $\mathcal{D}_P$ contains 12,110 samples with 551,944 label tokens for Mistral-7B, of which 27,943 are distinct, covering 85.3\% of its 32,768-token vocabulary. Since Qwen2.5-3B and Qwen2.5-7B share the same tokenizer, they share an identical $\mathcal{D}_P$ of 31,931 samples with 1,281,820 label tokens, of which 77,031 are distinct, covering 50.8\% of the 151,665-token tokenizer vocabulary (50.7\% relative to the embedding sizes of 151,936 and 152,064, respectively). On $\mathcal{D}_P$, the base models $M_{\theta_0}$ achieve an average accuracy of 66.4\% for Mistral-7B, 62.2\% for Qwen2.5-3B, and 65.2\% for Qwen2.5-7B (62.9\%, 59.5\%, and 62.4\% when averaged over samples).

\subsection{Baseline Descriptions and Implementation Details}
\label{app:baseline_details}

\noindent\textbf{Baseline Descriptions.}
We compare AEGIS against six representative baselines corresponding to the six PCD-mitigation paradigms established in Section~\ref{subsec:unifying_paradigms}:
\begin{itemize}
    \item \textbf{PSFT}~\cite{ICLR2026_da75d2bb} adapts the clipped probability-ratio surrogate objective to offline supervised fine-tuning to bound the policy divergence $D_{\text{KL}}(\pi_\theta \parallel \pi_{\text{ref}})$ within a proximal trust region, thereby inducing a Fisher metric barrier in activation space that prevents hidden states from breaching LVD decision hyperplanes on pervasive tasks.

    \item \textbf{SMART}~\cite{renduchintala2024smart} formulates instruction data curation as a two-stage cardinality-constrained submodular maximization problem, utilizing Graph Cut for task subset selection and Facility Location for non-redundant instance selection. 
    Within the DMA paradigm, it mitigates PCD by maximizing the convex hull volume spanned by active features in the training mixture, maintaining high global feature sparsity ($S \to 1$) to prevent higher-order geometric phase collapse.

    \item \textbf{KnOTS}~\cite{stoica2025model} applies singular value decomposition (SVD) to jointly transform and align task-specific weight updates ($\Delta W$) from homologous fine-tuned models into a shared, mutually orthogonal singular basis prior to model merging. 
    Within the PSO paradigm, it alleviates PCD by eliminating cross-task singular subspace overlaps in parameter space, severing the linear pathways of inter-task feature interference.

    \item \textbf{DARE}~\cite{pmlr-v235-yu24p} randomly drops a high ratio ($p \in [0.9, 0.99]$) of SFT delta parameters ($\Delta \theta = \theta - \theta^0$) to zero and rescales the remaining parameters by $1/(1-p)$ to approximate the original layer embeddings in expectation without retraining. 
    Within the DPP paradigm, it mitigates PCD by pruning the vast majority of redundant cross-coupling parameter updates responsible for parasitic feature crosstalk along ambient pervasive directions.

    \item \textbf{DiFT}~\cite{yuan2026differential} identifies task-critical parameter rows via Delta-Scale Row (DSR) sensitivity probing on forward activation differences and selectively updates only task-relevant differential parameter subsets while zeroing out gradients on remaining rows. 
    Within the GOO paradigm, it alleviates PCD by restricting gradient updates to task-specific subspaces so that optimization trajectories remain orthogonal to ambient pervasive feature channels and avoid destructive gradient conflicts.

    \item \textbf{CLUE}~\cite{ICLR2026_5576b86e} leverages logical circuit discovery (extracting AND, OR, and ADDER gates) and Conjunctive Normal Form (CNF) Boolean satisfiability solving to disentangle model parameters into task-specific, retain, and conflict nodes. 
    Within the MIL paradigm, it mitigates PCD by freezing task-irrelevant safe and retain sub-networks ($\Delta W_j \equiv 0, \forall j \in \mathcal{N}_P$), directly truncating unconstrained activation drift along ambient pervasive feature channels.
\end{itemize}

\noindent\textbf{Implementation Details.}
All experiments are conducted on a single-node server equipped with NVIDIA A100 GPUs interconnected via NVLink. 
To guarantee numerical precision for geometric margin computation and bisection solving, we enforce full \textbf{FP32} precision across the entire training and evaluation pipeline, encompassing model parameters, gradients, optimizer states, and all forward/backward computations. 
For distributed training, we employ PyTorch Fully Sharded Data Parallel (\textbf{FSDP}) under the full sharding strategy (\texttt{FULL\_SHARD}). 
All models are optimized using the \textbf{AdamW} optimizer with a learning rate of $1 \times 10^{-5}$ and a linear warmup schedule, trained for $3$ epochs with a maximum sequence length of $512$ tokens. 
During training, each mini-batch is constructed in a balanced $1:1$ paired manner containing exactly $1$ target task sample from $\mathcal{D}_T$ and $1$ pervasive task sample from $\mathcal{D}_P$. 
For the hyperparameters in the AEGIS objective, we set the semantic proximity temperature in Eq.~(41) to $\tau = 1.0$, the Huber smoothing threshold in Eq.~(44) to $\kappa = 0.05$, and the geometric regularization balancing weight in Eq.~(47) to $\lambda = 1.0$ (calibrated via 16-step root-mean-square gradient matching), while the exact bisection solver for the shared scalar $\zeta^*$ in Algorithm~1 is executed with an iteration budget of $T_{\max} = 48$ over competing token directions.

\section{Additional Results Across Diverse SFT Scenarios}
\label{app:additional_sft}

\subsection{Mix-up SFT}
\label{app:mixup_sft}

To evaluate the effectiveness of AEGIS when simultaneously adapting to multiple heterogeneous target tasks, we conduct Mix-up SFT experiments on Mistral-7B. 
Specifically, we select one representative dataset from each of the four task categories—natural language understanding (NLU), math reasoning, code generation, and logic \& instruction-following—and uniformly sample a total of 20,000 training instances at random across the four datasets. 
We construct two distinct task-mixture groups and repeat each experiment across five random seeds, with the results summarized in Table~\ref{tab:mixup_sft}.

\begin{table*}[t]
\centering
\caption{Mix-up SFT results on Mistral-7B across three four-task mixture groups. For each dataset, the left column reports the target task metric and the right column reports the pervasive average accuracy (P-Acc.). All experiments are conducted over five random seeds; the variance across seeds is consistently below $0.04$ with no statistically significant impact on the results and is thus omitted for clarity.}
\label{tab:mixup_sft}
\resizebox{\textwidth}{!}{%
\begin{tabular}{lccccccccccccccc}
\toprule
\multirow{2}{*}{\textbf{Method}} & \multicolumn{5}{c}{\textbf{Mix-up Group 1}} & \multicolumn{5}{c}{\textbf{Mix-up Group 2}} &\multicolumn{5}{c}{\textbf{Mix-up Group 3}}\\
\cmidrule(lr){2-6} \cmidrule(lr){7-11} \cmidrule(lr){12-16}
 & \textbf{SST-2} & \textbf{GSM8K} & \textbf{CodeAlpaca} & \textbf{LogiQA2} & P-Acc. & \textbf{RTE} & \textbf{GSM8K} & \textbf{Code-Bagel-Hermes} & \textbf{Evol-Instruct} & P-Acc.&\textbf{CoLA}& \textbf{GSM8K} & \textbf{CodeAlpaca} & \textbf{Evol-Instruct} & P-Acc.\\
\midrule
Vanilla        &0.938  &0.354 &0.115 &0.543 &0.535 &0.795 &0.343 &0.417 &0.429 &0.522 &0.405 &0.349 & 0.114&0.429 &0.539  \\
PSFT           &0.929  &0.361 &0.111 &0.529 &0.539 &0.801 &0.351 &0.419 &0.435 &0.526 &0.394 &0.356 &0.116 &0.436 &0.541  \\
SMART          &0.941  &0.355 &0.118 &0.545 &0.537 &0.798 &0.348 &0.412 &0.436 &0.523 &0.409 &0.348 &0.113 &0.439 &0.535  \\
KnOTS          &0.925  &0.307 &0.094 &0.519 &0.491 &0.764 &0.324 &0.395 &0.405 &0.475 &0.359 &0.314 &0.105 &0.405 &0.501  \\
DARE           &0.918  &0.339 &0.098 &0.506 &0.509 &0.778 &0.334 &0.408 &0.401 &0.488 &0.387 &0.316 &0.101 &0.411 &0.514  \\
DiFT           &0.942  &0.366 &0.108 &0.559 &0.543 &0.802 &0.349 &0.425 &0.435 &0.528 &0.411 &0.349 &0.119 &0.435 &0.538  \\
CLUE           &0.949  &0.384 &0.124 &0.562 &0.547 &0.805 &0.356 &0.421 &0.439 &0.533 &0.416 &0.357 &0.118 &0.436 &0.546  \\
\textbf{AEGIS} &0.969  &0.405 &0.133 &0.568 &0.569 &0.811 &0.374 &0.424 &0.446 &0.546 &0.429 &0.366 &0.124 &0.445 &0.549  \\
\bottomrule
\end{tabular}%
}
\end{table*}

As shown in Table~\ref{tab:mixup_sft}, \textbf{AEGIS} maintains substantial superiority over all baselines in the Mix-up SFT setting. 
This demonstrates that AEGIS's ability to suppress feature interference induced by task adaptation remains robust across diverse target task categories and sample heterogeneity~\cite{yuan2026differential}. 
Similarly, we attribute the performance gains on the individual target tasks to the fact that AEGIS mitigates mutual feature interference among the co-trained target tasks, thereby preserving and enhancing the feature benefit of each respective target task.

\subsection{Parameter-Efficient Fine-Tuning (PEFT)}
\label{app:peft_sft}

\begin{figure*}[t]
    \centering
    \begin{subfigure}[b]{0.32\textwidth}
        \centering
        \includegraphics[width=\textwidth]{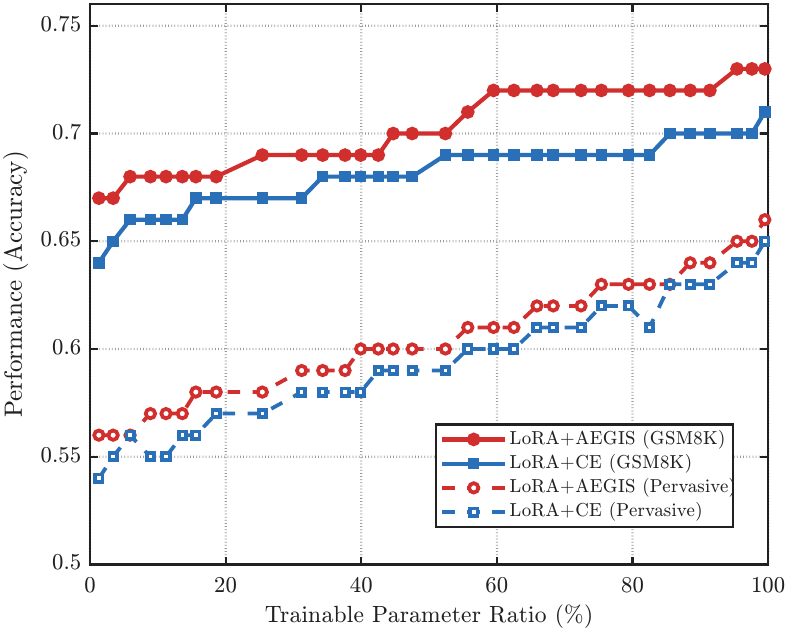}
        \caption{LoRA}
        \label{fig:peft_lora}
    \end{subfigure}
    \hfill
    \begin{subfigure}[b]{0.32\textwidth}
        \centering
        \includegraphics[width=\textwidth]{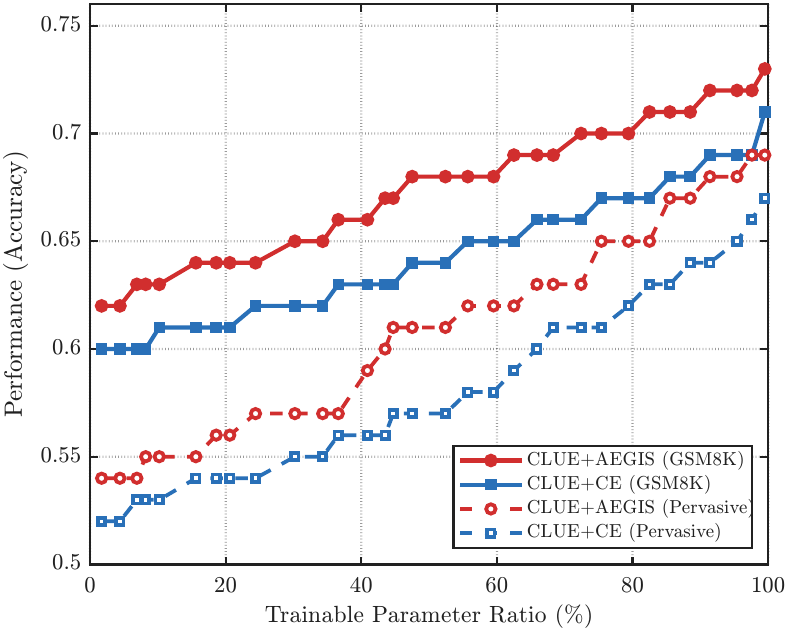}
        \caption{CLUE}
        \label{fig:peft_clue}
    \end{subfigure}
    \hfill
    \begin{subfigure}[b]{0.32\textwidth}
        \centering
        \includegraphics[width=\textwidth]{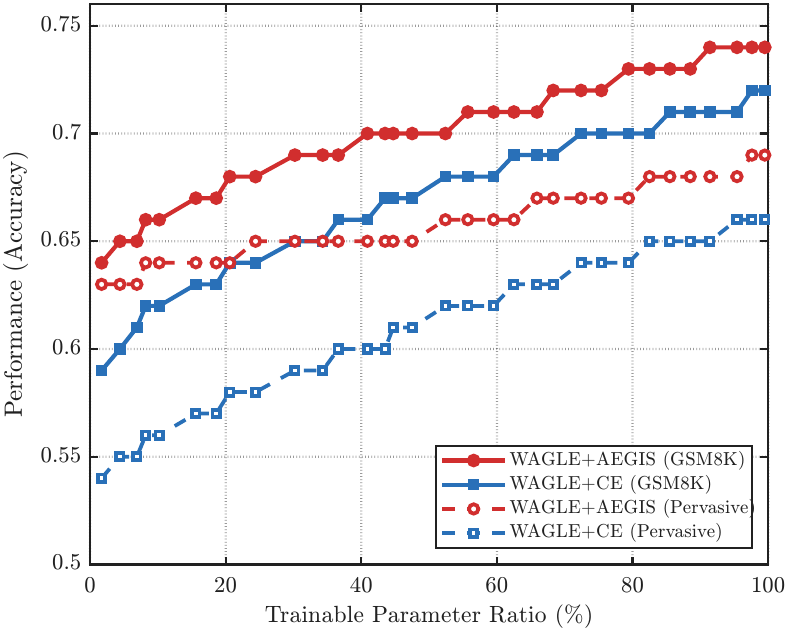}
        \caption{WAGLE}
        \label{fig:peft_wagle}
    \end{subfigure}
    \caption{Performance of AEGIS under different PEFT baselines across varying percentages of trainable parameters.}
    \label{fig:peft_comparison}
\end{figure*}

To evaluate the performance of our method under parameter-efficient fine-tuning (PEFT) regimes, we select three representative baselines that update only a subset of model parameters through distinct mechanisms: \textbf{LoRA}~\cite{hu2022lora}(low rank localization), \textbf{CLUE}~\cite{ICLR2026_5576b86e} (mechanistic interpretability localization), and \textbf{WAGLE}~\cite{jia2024wagle} (gradient-based attribution localization). 
Across these three baselines, we compare the effect of combining AEGIS with standard cross-entropy on the pervasive dataset $\mathcal{D}_P$. 
Specifically, we use ``\textbf{+CE}'' to denote that the loss function on $\mathcal{D}_P$ consists solely of the traditional cross-entropy loss, and ``\textbf{+AEGIS}'' to denote that the loss on $\mathcal{D}_P$ incorporates both the traditional cross-entropy loss and our AEGIS regularization (Eq.~(47)). 
We evaluate Qwen2.5-7B on the GSM8K dataset across varying percentages of trainable parameters, with the results presented in Figure~\ref{fig:peft_comparison}.

As shown in Figure~\ref{fig:peft_comparison}, across both the target task (GSM8K) and the pervasive tasks, our method (\textbf{+AEGIS}) consistently and significantly outperforms the standard PEFT baselines (\textbf{+CE}) at the same trainable parameter ratios. 
We conjecture that this advantage demonstrates the superior parameter efficiency of AEGIS over standard cross-entropy, which stems from its ability to move beyond one-hot categorical supervision and simultaneously supervise activation drift across multiple token directions.

\subsection{Continual SFT}
\label{app:continual_sft}

To evaluate the effectiveness of AEGIS under sequential multi-stage adaptation, we conduct continual SFT experiments across three models: Mistral-7B, Qwen2.5-7B, and Gemma-2-9B. 
We design two five-task sequential training streams spanning diverse task domains: 
\textbf{Group 1} ($\text{GSM8K} \rightarrow \text{SST-2} \rightarrow \text{RTE} \rightarrow \text{Evol-Instruct} \rightarrow \text{Code-Bagel-Hermes}$) and 
\textbf{Group 2} ($\text{CoLA} \rightarrow \text{Infinity-Instruct-Code} \rightarrow \text{LogiQA2} \rightarrow \text{GSM8K} \rightarrow \text{MRPC}$). 
In each subfigure, the horizontal axis denotes the five sequential training task nodes, and the vertical axis reports the relative performance drop ($\text{-}x\%$) compared to single-task SFT. 

\begin{figure*}[t]
    \centering
    \begin{subfigure}[b]{0.32\textwidth}
        \centering
        \includegraphics[width=\textwidth]{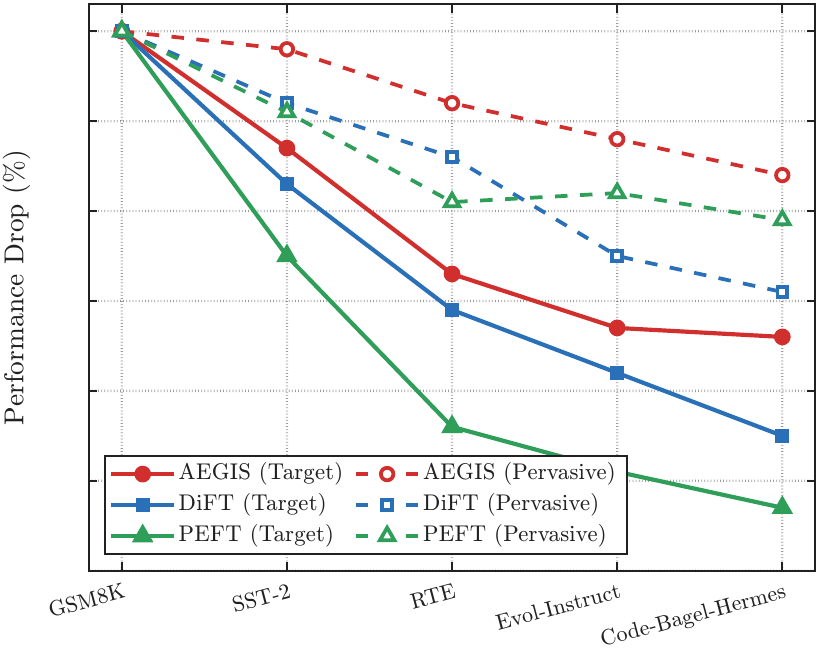}
        \caption{Group 1: Mistral-7B}
        \label{fig:cont_g1_mistral}
    \end{subfigure}
    \hfill
    \begin{subfigure}[b]{0.32\textwidth}
        \centering
        \includegraphics[width=\textwidth]{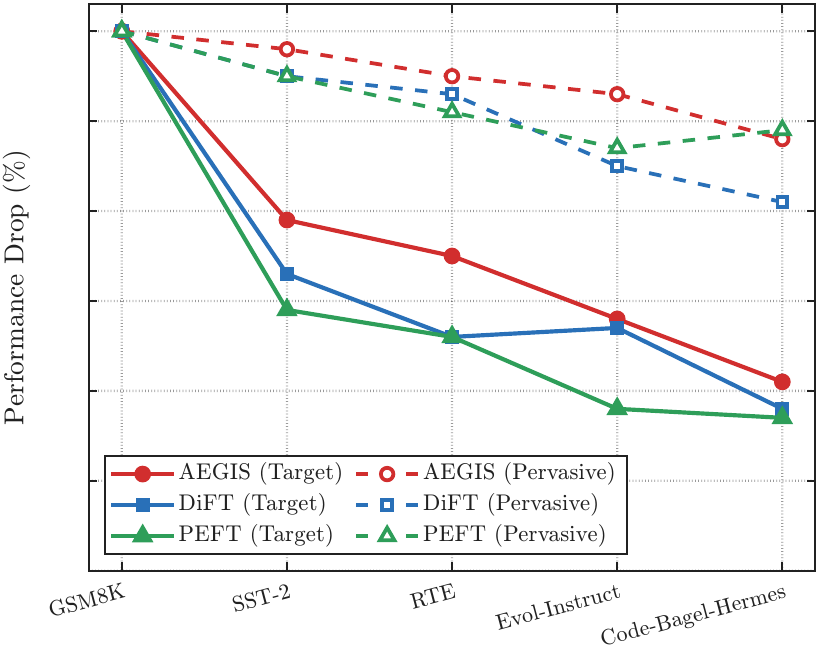}
        \caption{Group 1: Qwen2.5-7B}
        \label{fig:cont_g1_qwen}
    \end{subfigure}
    \hfill
    \begin{subfigure}[b]{0.32\textwidth}
        \centering
        \includegraphics[width=\textwidth]{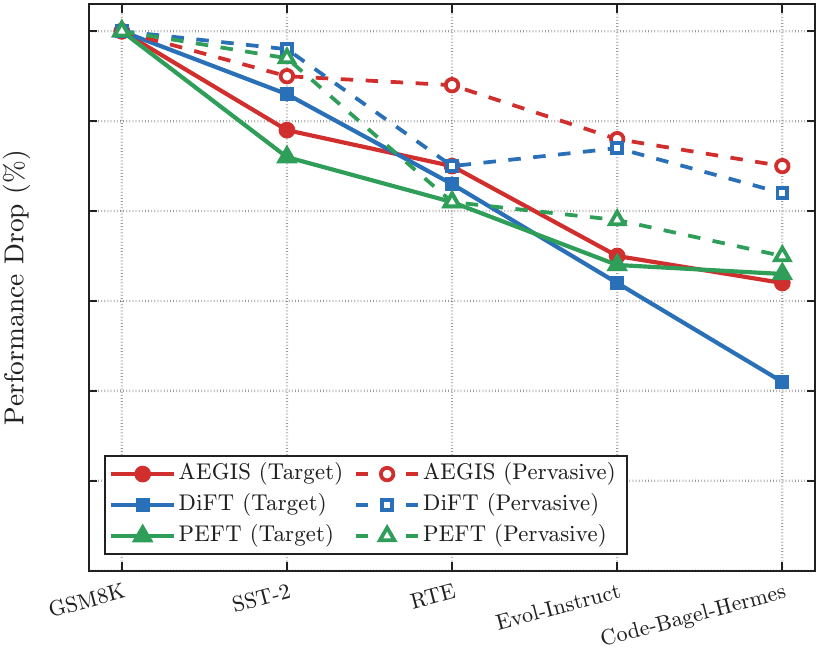}
        \caption{Group 1: Gemma-2-9B}
        \label{fig:cont_g1_gemma}
    \end{subfigure}
    
    \vspace{0.8em}
    
    \begin{subfigure}[b]{0.32\textwidth}
        \centering
        \includegraphics[width=\textwidth]{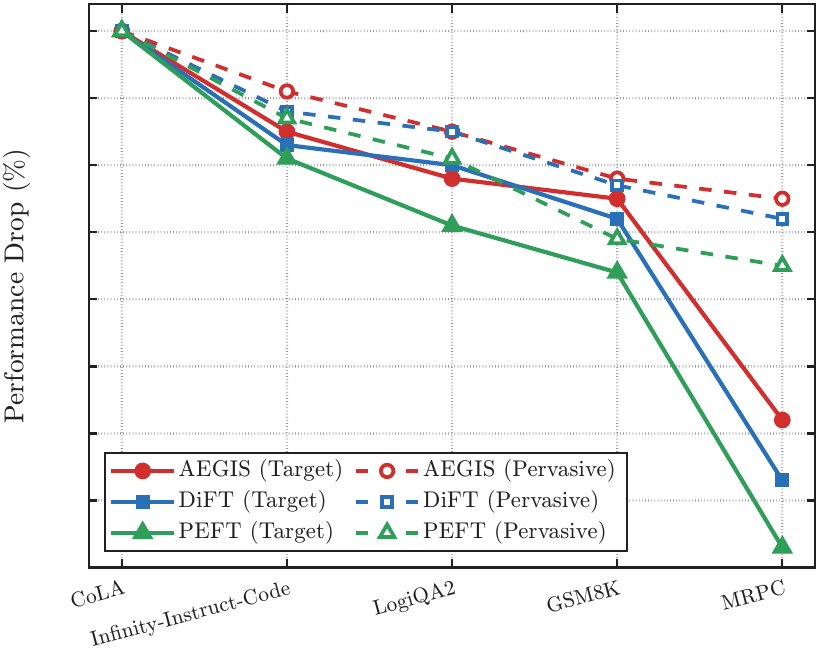}
        \caption{Group 2: Mistral-7B}
        \label{fig:cont_g2_mistral}
    \end{subfigure}
    \hfill
    \begin{subfigure}[b]{0.32\textwidth}
        \centering
        \includegraphics[width=\textwidth]{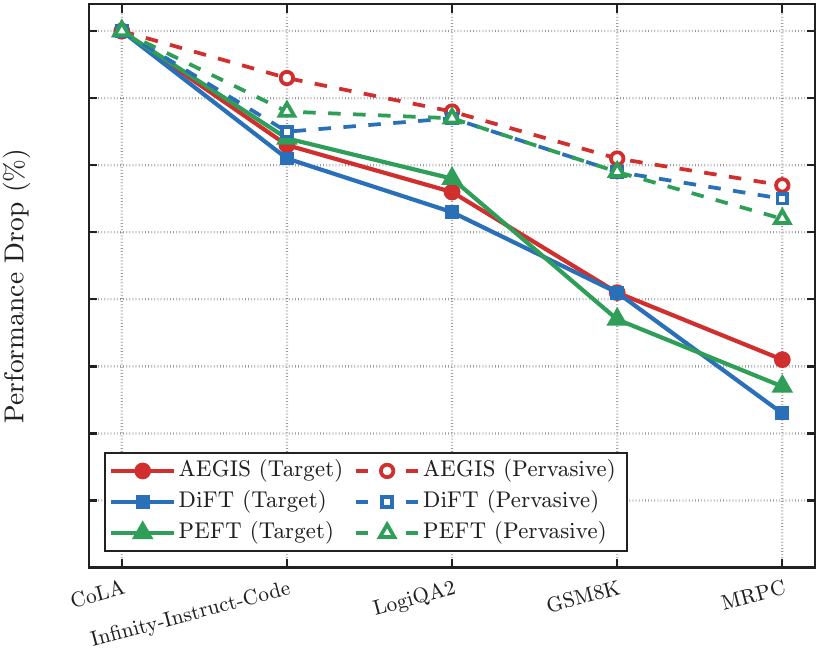}
        \caption{Group 2: Qwen2.5-7B}
        \label{fig:cont_g2_qwen}
    \end{subfigure}
    \hfill
    \begin{subfigure}[b]{0.32\textwidth}
        \centering
        \includegraphics[width=\textwidth]{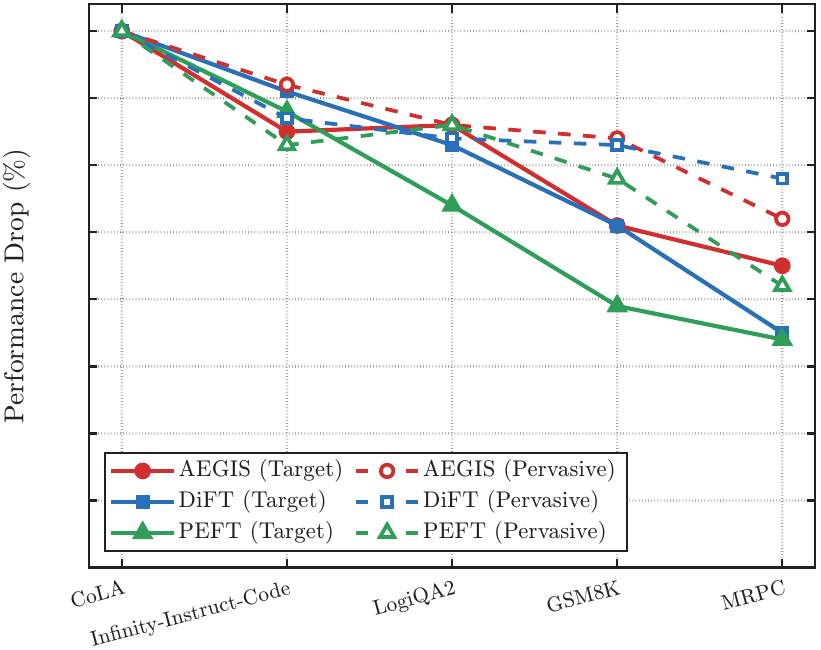}
        \caption{Group 2: Gemma-2-9B}
        \label{fig:cont_g2_gemma}
    \end{subfigure}
    
    \caption{Continual SFT performance across two five-task sequential streams on Mistral-7B, Qwen2.5-7B, and Gemma-2-9B. Solid and dashed lines denote the relative performance drop ($\text{-}x\%$) on target tasks and pervasive tasks compared to single-task training, respectively.}
    \label{fig:continual_sft}
\end{figure*}

We compare \textbf{AEGIS} against \textbf{DiFT}~\cite{yuan2026differential} and \textbf{PEFT}, with the results across all six configurations presented in Figure~\ref{fig:continual_sft}.As shown in Figure~\ref{fig:continual_sft}, \textbf{AEGIS} consistently exhibits strong resilience in continual SFT, incurring minimal negative interference from prior learning stages even after sequentially adapting to multiple tasks. 
In contrast, while \textbf{PEFT} constrains parameter updates to remain close to the distribution of the preceding gradient step and \textbf{DiFT} utilizes a probe model to identify update directions with minimal task interference, both methods progressively suffer from severe catastrophic forgetting as the number of accumulated tasks increases during continual SFT.

\setcounter{table}{0}
\renewcommand{\thetable}{I\arabic{table}}
\makeatletter
\@ifundefined{theHtable}{}{\renewcommand{\theHtable}{I\arabic{table}}}
\makeatother
\providecommand{\Itok}[1]{\texttt{#1}}
\providecommand{\Isp}{\textvisiblespace}
\providecommand{\Ieos}{\textless/s\textgreater}
\providecommand{\Ibyte}[1]{\textless0x#1\textgreater}
\providecommand{\Iguj}{[U+0ABE]}
\iftutex
  \IfFontExistsTF{Noto Sans Gujarati}{%
    \newfontfamily\Igujfont{Noto Sans Gujarati}[Scale=MatchLowercase]%
    \renewcommand{\Iguj}{{\Igujfont\char"0ABE}}}{}%
\fi
\providecommand{\Idrift}[3]{%
  \begin{tabular}[t]{@{}c@{}}\Itok{#1}$\rightarrow$\Itok{#2}\\[-1pt]{\tiny(#3)}\end{tabular}}
\providecommand{\Ilabel}[3]{%
  \begin{tabular}[t]{@{}c@{}}\textbf{\Itok{#1}$\rightarrow$\Itok{#2}}\\[-1pt]{\tiny\textbf{(#3)}}\end{tabular}}

\section{Top-10 Drift Directions}
\label{app:drift_directions}

To further analyze how AEGIS and standard SFT (Vanilla) differ along individual token directions, we group the decision-edge crossings on the pervasive dataset $\mathcal{D}_E$ by their direction. An edge $(t, k)$ crossed along direction $j_t \rightarrow k$ means that the activation at position $t$ has exited $\mathrm{Cell}^{\mathrm{LVD}}_{j_t}$ of the ground-truth token $j_t$ through the hyperplane $\Pi_{j_t,k}$, so that the competing token $k$ outscores $j_t$. Table~\ref{tab:drift_directions} lists, for MRPC, CodeAlpaca, and GSM8K, the ten directions with the most newly crossed edges after SFT, together with the ten directions with the most crossed edges under the base model as a reference.

\begin{table}[h]
\centering
\caption{Top-10 drift directions on the pervasive dataset $\mathcal{D}_E$ from Mistral-7B.  Bold entries point toward a target task label token (\Itok{yes}/\Itok{no} for MRPC, digits for CodeAlpaca and GSM8K). \Itok{\Isp} marks a leading space, \Itok{\Ieos} is the end-of-sequence token, \Itok{\Ibyte{XX}} is a byte-fallback token, and \Iguj{} is the Gujarati vowel sign AA (U+0ABE).}
\label{tab:drift_directions}
\scriptsize
\setlength{\tabcolsep}{2.5pt}
\renewcommand{\arraystretch}{1.15}
\resizebox{\textwidth}{!}{%
\begin{tabular}{c c cc cc cc}
\toprule
\multirow{2}{*}{Rank} & \multirow{2}{*}{Base} & \multicolumn{2}{c}{MRPC} & \multicolumn{2}{c}{CodeAlpaca} & \multicolumn{2}{c}{GSM8K} \\
\cmidrule(lr){3-4} \cmidrule(lr){5-6} \cmidrule(lr){7-8}
 & & Vanilla & AEGIS & Vanilla & AEGIS & Vanilla & AEGIS \\
\midrule
1  & \Idrift{\Ieos}{\Ibyte{0A}}{11784} & \Ilabel{\Ieos}{\Isp yes}{4327} & \Idrift{.}{\Ieos}{1578}            & \Idrift{.}{,}{3330}               & \Idrift{,}{\Isp and}{172}      & \Idrift{\Isp answer}{\Isp final}{4154} & \Idrift{.}{,}{218} \\
2  & \Idrift{\Ieos}{\Isp}{5768}         & \Ilabel{\Ieos}{\Isp no}{3474}  & \Idrift{\Ibyte{0A}}{\Ieos}{1473}   & \Ilabel{1}{2}{3173}               & \Idrift{.}{\Isp and}{156}      & \Idrift{\Isp a}{\Isp the}{2250}        & \Idrift{,}{\Isp and}{125} \\
3  & \Idrift{\Ieos}{.}{5151}            & \Ilabel{\Ieos}{yes}{1927}      & \Idrift{,}{\Ieos}{1445}            & \Ilabel{1}{4}{2150}               & \Idrift{.}{,}{52}              & \Idrift{\Isp The}{\Ieos}{1156}         & \Idrift{\Isp the}{\Isp a}{115} \\
4  & \Idrift{\Ieos}{-}{4044}            & \Idrift{\Isp(}{\Ieos}{1259}    & \Idrift{,}{\Isp and}{1342}         & \Ilabel{3}{4}{2121}               & \Idrift{\Ibyte{0A}}{\Ieos}{42} & \Ilabel{1}{2}{659}                     & \Idrift{.}{\Isp and}{93} \\
5  & \Idrift{\Ieos}{(}{3879}            & \Ilabel{\Ieos}{no}{729}        & \Idrift{\Isp}{\Ieos}{1330}         & \Idrift{\Isp a}{\Isp the}{2119}   & \Idrift{,}{.}{39}              & \Ilabel{4}{2}{237}                     & \Idrift{,}{.}{92} \\
6  & \Idrift{\Ieos}{:}{3873}            & \Idrift{.}{\Ieos}{377}         & \Idrift{\Isp the}{\Isp a}{1329}    & \Ilabel{5}{4}{1108}               & \Idrift{\Isp}{\Ibyte{E0}}{34}  & \Ilabel{5}{2}{224}                     & \Idrift{\Ibyte{E0}}{\Isp}{80} \\
7  & \Idrift{\Ieos}{The}{3794}          & \Idrift{,}{.}{194}             & \Idrift{,}{.}{1235}                & \Ilabel{2}{4}{685}                & \Idrift{\Ibyte{A9}}{\Ibyte{A8}}{33} & \Ilabel{5}{4}{220}                & \Idrift{\Ibyte{E0}}{\Iguj}{77} \\
8  & \Idrift{\Ieos}{...}{3779}          & \Idrift{\Isp}{\Ibyte{E0}}{141} & \Idrift{.}{\Isp and}{1209}         & \Ilabel{5}{6}{277}                & \Ilabel{3}{4}{27}              & \Ilabel{0}{2}{118}                     & \Idrift{\Isp and}{,}{71} \\
9  & \Idrift{\Ieos}{1}{3777}            & \Idrift{\Ibyte{AA}}{\Ibyte{AB}}{121} & \Idrift{\Isp(}{\Ieos}{1179}  & \Ilabel{1}{3}{75}                 & \Idrift{,}{\Isp in}{25}        & \Ilabel{0}{4}{32}                      & \Idrift{\Isp a}{\Isp the}{69} \\
10 & \Idrift{\Ieos}{2}{3664}            & \Idrift{\Isp in}{\Ieos}{113}   & \Idrift{\Isp and}{\Ieos}{1151}     & \Ilabel{0}{4}{67}                 & \Idrift{\Isp}{\Ieos}{23}       & \Ilabel{1}{6}{21}                      & \Idrift{,}{\Isp in}{67} \\
\bottomrule
\end{tabular}}
\end{table}

The drift directions of the base model show that the feature interference inherent in the almost-orthogonal configuration formed during pre-training manifests mainly as crossings toward generic, semantically empty tokens, such as the end-of-sequence token drifting toward a newline, a space, or punctuation. Such crossings barely affect the prediction of the main content. In contrast, cross-entropy SFT on all three target tasks deflects the activations of pervasive tasks toward the label tokens of the target task. On MRPC, whose labels are \Itok{yes} and \Itok{no}, a large number of pervasive activations drift from \Itok{\Ieos} toward \Itok{\Isp yes}, \Itok{\Isp no}, \Itok{yes}, and \Itok{no}, which together account for 10,457 newly crossed edges, and on CodeAlpaca and GSM8K most of the top-10 directions point toward digits. AEGIS clearly suppresses these drifts: none of its top-10 directions on MRPC and GSM8K points toward a target label token, and on CodeAlpaca only \Itok{3}$\rightarrow$\Itok{4} remains, with 27 newly crossed edges compared with 2,121 under Vanilla. The remaining drifts of AEGIS mostly occur between punctuation, conjunctions, articles, the end-of-sequence token, and fragments of multi-byte characters, which resemble the generic crossings of the base model. We regard the shared drift of pervasive activations toward tokens with concrete task semantics as a manifestation of overfitting to the target task. By suppressing this overfitting, AEGIS also improves performance on the target task, which is consistent with the loss and validation accuracy trends in Figure~\ref{fig:loss_acc}.
\end{document}